\documentclass{article}

\usepackage[main, final]{neurips_2026}

\usepackage{algorithm}
\usepackage{algpseudocode}
\usepackage{amssymb}
\usepackage{amsmath}
\usepackage{mathtools}
\usepackage{makecell}
\usepackage{xcolor,colortbl}
\usepackage{xspace}
\usepackage{url}
\usepackage{hyperref}

\usepackage{booktabs}
\usepackage[tableposition=top]{caption} 

\usepackage{multibib}
\newcites{Appendix}{Additional References}

\usepackage{bbm}

\newcommand{\yhdr}[1]{\vspace{-1mm}\noindent{{\bf \underline{#1}}}}
\newcommand{\xhdr}[1]{\vspace{-1mm}\noindent{{\bf #1.}}}
\newcommand{\name}{\textsc{RADT}\xspace}

\usepackage[font=footnotesize,labelfont=bf]{caption} 

\title{Hindsight Relabeling is All You Need \\ for Reach-Avoid Learning}

\author{Kevin Li \\ 
  Massachusetts Institute of Technology\\
  Harvard Medical School\\
  \texttt{likevin@mit.edu} \\
\And
  Marinka Zitnik \\
  Harvard Medical School \\
  Kempner Institute at Harvard University \\
  \texttt{marinka@hms.harvard.edu} \\
}

\begin{document}

\maketitle
\vspace{-4mm}
\begin{abstract}
\vspace{-1mm}
Offline goal-conditioned reinforcement learning methods have shown promise for reach-avoid tasks, where an agent must reach a target state while avoiding undesirable regions of the state space. Existing approaches typically encode avoid-region information into an augmented state space and cost function, which prevents flexible, dynamic specification of novel avoid-region information at evaluation time. They also rely heavily on meticulously designed reward and cost functions, limiting transferability to novel environments and specifications.
We introduce \name, a decision transformer model for offline, reward-free, goal-conditioned, avoid-region-conditioned RL. \name encodes goals and avoid-regions directly as prompt tokens, allowing any number of avoid-regions of arbitrary size to be specified at evaluation time. Using only suboptimal offline trajectories from a random policy, \name can learn reach-avoid behavior in a completely data-driven manner using our novel avoid-region hindsight relabeling approach, without the need for reward/cost supervision.
We benchmark \name against offline goal-conditioned RL models on 17 tasks that span multiple environments and experimental settings. \name generalizes zero-shot to avoid-region sizes and counts outside the training distribution and outperforms baselines that require retraining. In one zero-shot setting, \name reduces normalized cost by 35.7\% relative to the best retrained baseline and achieves a high goal-reaching success rate.
We also apply \name to cell reprogramming, where the model plans perturbation sequences that move cells toward target cell types and away from undesirable intermediate cell states.

\end{abstract}

\vspace{-3mm}
\section{Introduction}\label{introduction}
\vspace{-2mm}
Many sequential decision-making problems~\citep{dsrl, bulletsafetygym, autonomousdrivingrewardreview} are naturally framed as reach-avoid tasks~\citep{livelinessguarantees, mincostreachavoid,cbsreachavoid}, in which an agent must reach a designated goal state while avoiding undesirable regions of the state space. These problems arise in diverse domains, including robotics~\citep{bulletsafetygym, safetygymnasium, cao2024offline} (e.g., robotic arms reaching for targets while avoiding fragile objects), vehicle navigation~\citep{dsrl, autonomousdrivingrewardreview} (e.g., routing to a destination while avoiding slow zones), and biology~\citep{tumorgenesis, stratsfortumorgenesis} (e.g., cell reprogramming strategies that reach a desired gene expression state without traversing tumorigenic intermediates). Despite domain-specific differences, these tasks share a common structure: they require balancing goal achievement with dynamic avoidance of specified hazards.

A diverse array of tasks fall under the reach-avoid framework, each with varying degrees of tolerance to goal-reaching and avoidance success that impose assorted sets of challenges, restrictions, and requirements. In safety-critical tasks where entering undesirable states can have irreversible and unacceptable consequences (e.g., an autonomous vehicle that must avoid pedestrians), online exploration is often precluded, making offline learning necessary~\citep{dsrl}. In tasks where undesirable regions are not hard restrictions and could even be unavoidable (e.g., a vehicle must reach its destination, but should try to avoid construction zones for greater efficiency), methods derived under the framework of stringent constraint satisfaction may not be suitable~\citep{cpo, cmdp, constrained_qlearning}. Furthermore, in practical deployments, the specification of goals and avoid-regions may change based on user preferences or environmental context. For instance, a robot assistant may need to adapt to new furniture layouts, or a therapeutic model may need to avoid different toxic intermediate states based on patient-specific risk factors. These settings require flexible and interpretable models that support zero-shot generalization to unseen goal and avoid specifications without retraining.

\begin{figure}[t]
\begin{centering}
\includegraphics[width=\textwidth]{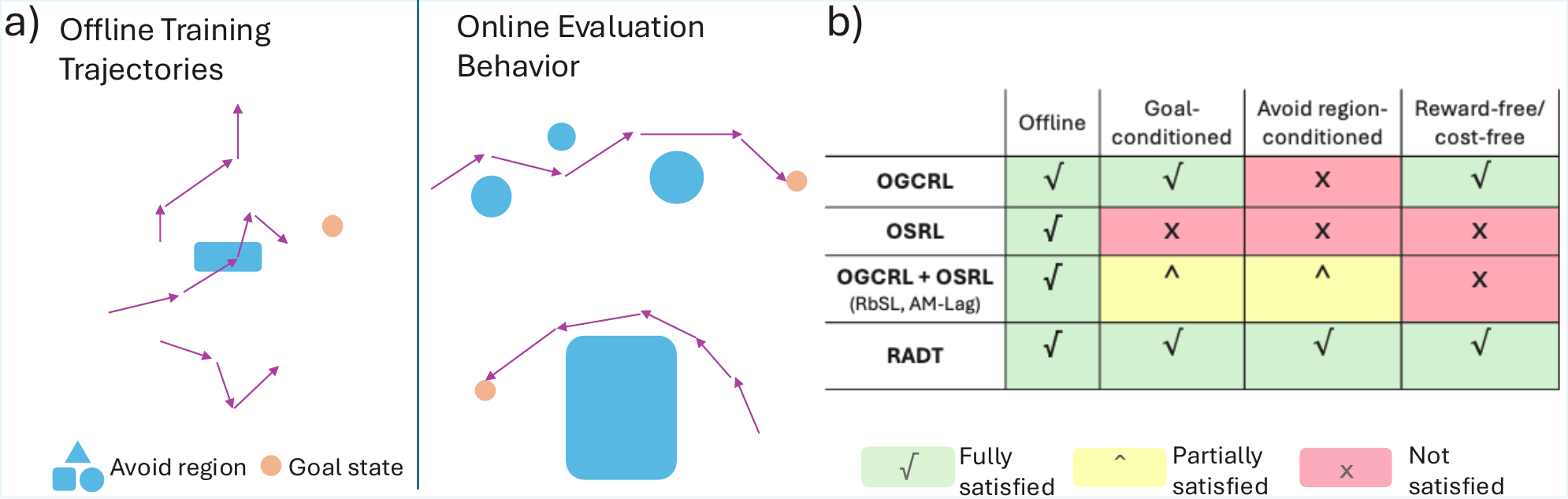}
\caption[]{(a) An ideal reach-avoid model should learn to avoid arbitrarily specified regions of varying number and size at evaluation time, using only suboptimal, random-policy training data. (b) \name is the only model that satisfies all criteria for an ideal reach-avoid learner (Section~\ref{problemformulation}).}\label{fig:crtieria-model}
\vspace{-5mm}
\end{centering}
\end{figure}

Due to these challenges, reach-avoid learning remains difficult. Several lines of work attempt to address parts of the reach-avoid problem (Section~\ref{relatedwork}), but fail to meet one or more key criteria needed for flexible and effective reach-avoid learning, including: offline learning from suboptimal data, dynamic test-time conditioning on arbitrary goals/avoid-regions, and reward-free training (Section~\ref{problemformulation}, Figure~\ref{fig:crtieria-model}). Most existing approaches rely on augmented state representations and carefully shaped cost functions to encode avoid behavior~\citep{cao2024offline, constrained_qlearning, fisor, coptidice, batchlearningcmdp,dsrl}. This tight coupling of avoid-region semantics to model internals prevents flexible deployment and limits generalization. Reward-based formulations often struggle to represent multiple behavioral preferences simultaneously, especially when goal-reaching and avoidance conflict, and rely on constrained optimization setups that imply an inherent prioritization of avoidance over goal-reaching that is not always desired~\citep{autonomousdrivingrewardreview, complexrewardfunctionsandcurriculumlearning, howtoreward, cpo, cmdp, constrained_qlearning}. Reward-free methods avoid these issues but lack a mechanism for dynamically conditioning behavior on new avoid constraints~\citep{gcsl, wgcsl, trajectorytransformer, contrastivelearningrl}. Moreover, many offline approaches rely on expert near-optimal demonstrations to learn strong policies~\citep{ogbench, dsrl, cao2024offline, bcq, bear}, which are often not available in safety-critical or high-dimensional tasks~\citep{limiteddata, shouldidooffline,domainunlabeled}.



\xhdr{Present work}
We introduce \name \footnote{RADT code: \url{https://github.com/mims-harvard/RADT}} (Reach-Avoid Decision Transformer), a reward-free, data-driven, offline RL approach for goal-conditioned and avoid-region-conditioned reach-avoid learning (Figure~\ref{fig:crtieria-model}). \name is a decision transformer that represents goals and avoid-regions as prompt tokens. This formulation decouples the reach-avoid specification from the state representation and enables zero-shot generalization to arbitrary numbers and sizes of avoid-regions. \name learns policies entirely from random-policy trajectories using a novel combination of goal and avoid-region hindsight relabeling, with no need for reward/cost functions.
Our main contributions include:
\textcircled{\small{1}} A prompting framework for reach-avoid learning that encodes goals and avoid-regions as discrete prompt tokens, allowing flexible and interpretable conditioning of behavior at test time.
\textcircled{\small{2}} A novel avoid-region hindsight relabeling strategy that allows the model to learn successful avoid behavior from suboptimal data in a fully data-driven manner, without the need for reward/cost functions.
\textcircled{\small{3}} A decision transformer model trained on random-policy data using this approach, with benchmarking across robotics and biological domains evaluates generalization to out-of-distribution avoid-region sizes and counts.
\textcircled{\small{4}} Strong results showing that \name generalizes zero-shot to 12 unseen reach-avoid configurations, outperforming 3 existing methods retrained directly on those configurations.

\vspace{-2mm}
\section{Desiderata/Desirable Properties of Reach-Avoid RL Models}\label{problemformulation}
\vspace{-1mm}

\xhdr{Notation}
We first establish the notation used throughout this work. A trajectory $\tau$ of length $T$ is a sequence of alternating states and actions collected by an agent making sequential decisions in an environment: 
$$\tau = (\mathbf{s}_1, \mathbf{a}_1, \mathbf{s}_2, \mathbf{a}_2, \dots, \mathbf{s}_T, \mathbf{a}_T)$$
where $\mathbf{s}_t \in \mathcal{S}$ is the state of the agent at timestep $t$ and $\mathbf{a}_t \in \mathcal{A}$ is the action the agent takes from $\mathbf{s}_t$. $\mathcal{S} \subseteq \mathbb{R}^{d_s}$ represents the state space and $\mathcal{A} \subseteq \mathbb{R}^{d_a}$ represents the action space, which are the spaces of all possible states and actions, respectively.
%
All RL models in this work learn a deterministic policy $\pi(\mathbf{a}_t \mid \cdot)$ that selects the most preferred action $\mathbf{\hat{a}}_t$ for the agent to take at timestep $t$ given contextual inputs, typically including at least the current state $\mathbf{s}_t$ (i.e., $\pi(\mathbf{a}_t \mid \mathbf{s}_t)$). At each timestep of a trajectory, a \textit{state transition} occurs, where the agent takes an action $\mathbf{a}$ from their current state $\mathbf{s}$ and ends up in a new state $\mathbf{s}'$. This is denoted by the transition tuple $(\mathbf{s}, \mathbf{a}, \mathbf{s}')$. If used, the reward function $r(\mathbf{s}, \mathbf{a}, \mathbf{s}')$ and cost function $\texttt{cost}(\mathbf{s}, \mathbf{a}, \mathbf{s}')$ return scalar values for each transition tuple in the trajectory. When applicable, these values are included in the trajectory as follows:
$$\tau = (\mathbf{s}_1, \mathbf{a}_1, r_1, c_1, \mathbf{s}_2, \mathbf{a}_2, \dots)$$
In goal-conditioned settings, a goal state $\mathbf{g} \in \mathcal{S}$ is provided as an additional input, yielding conditional functions such as $\pi(\mathbf{a} \mid \cdot, \mathbf{g})$ or $r(\mathbf{s}, \mathbf{a}, \mathbf{s}', \mathbf{g})$. In avoid-region-conditioned settings with $n_{\text{avoid}}$ avoid-regions $\mathbf{b}_j: j \in \{1,2, ..., n_{\text{avoid}}\}$, similar conditioning applies. We refer to the center of avoid-region $\mathbf{b}_j$ as its avoid centroid, denoted $\texttt{centroid}(\mathbf{b}_j)$.

\xhdr{Desiderata/Desirable properties} Reach-avoid problems introduce a dual behavioral objective: the agent must reach a desired target state while avoiding explicitly defined regions of the state space. Reach-avoid models that satisfy this behavioral objective under real-world deployment constraints need to achieve the following key properties:

\xhdr{Property 1 (P1): Pre-collected offline datasets with no online fine-tuning} The model must learn solely from offline datasets $\mathcal{D}$ containing of pre-collected trajectories $\tau^{(i)}$, with no reliance on online fine-tuning. In safety-critical applications, online exploration may be infeasible, especially when entering avoid-regions could cause irreversible harm or failure.
    
\xhdr{Property 2 (P2): Suboptimality-tolerant learning} The model must learn strong policies from offline datasets that contain only suboptimal trajectories, i.e., those that do not reach the goal or that violate the avoid constraint. Specifically, it should support super-demonstration performance by learning from trajectories $\tau^{(i)}$ that: \textbf{(2.1)} fail to reach the target goal state $\mathbf{g}$ at rollout time, and/or \textbf{(2.2)} pass through avoid-regions $\mathbf{b}_j$ rather than successfully avoiding them.

\xhdr{Property 3 (P3): Goal-conditioned generalization} The model must generalize to any arbitrarily specified goal state $\mathbf{g}$ at evaluation time, without additional training. In real-world scenarios, such as autonomous navigation or therapeutic reprogramming, the target goal is often specified dynamically and cannot be hardcoded at training time.

\xhdr{Property 4 (P4): Avoid-region-conditioned generalization} The model must be able to learn a policy that can avoid any dynamically specified avoid-region(s) $\mathbf{b}_j$ of the state space at evaluation time, without additional training/finetuning. This includes supporting changes in: \textbf{(4.1)} the number of avoid-regions $n_{\text{avoid}}$, \textbf{(4.2)} their locations $\texttt{centroid}(\mathbf{b}_j)$, and \textbf{(4.3)} their sizes (i.e., spatial extent of the state space around each $\texttt{centroid}(\mathbf{b}_j)$ to avoid).

\xhdr{Property 5 (P5): Reward-free learning} The model must learn reach-avoid behavior without requiring a manually designed reward or cost function. Reward shaping is often brittle and requires expert domain knowledge~\citep{complexrewardfunctionsandcurriculumlearning, rewardesignreview, howtoreward, autonomousdrivingrewardreview}, especially when preferences over reaching and avoiding are difficult to encode or conflict. Instead, one should be possible to specify goals and avoid-regions directly as inputs to the model. Additionally, reward-based safe RL methods are usually coupled to a constraint satisfaction requirement with respect to a cost function ~\citep{cpo, cmdp, constrained_qlearning, cao2024offline}, implying a priority hierarchy between goal-reaching and avoidance objectives. Such a priority hierarchy may be relevant for certain safety-critical environments, but does not apply to all reach-avoid problems (Section~\ref{introduction}) and can even be actively incompatible in situations where avoid-regions are unavoidable (e.g., Section~\ref{experimentsreprogramming}). Removing reward functions and cost functions eliminates the need to frame the problem as a constrained optimization problem and dissolves this implicit priority hierarchy. This relates to our choice of a Pareto-style evaluation framework discussed in Section~\ref{experiments} under "Evaluation Framework" and Appendix~\ref{sec:related-work-appendix} under "Pareto vs. Constraint Optimization Evaluation Framework."


While many prior approaches address subsets of these properties, none satisfy all five simultaneously (Figure~\ref{fig:crtieria-model}b). We discuss these limitations in detail in Section~\ref{relatedwork}.

\vspace{-2mm}
\section{Related Work}\label{relatedwork}
\vspace{-2mm}

We review four key areas in reach-avoid learning: offline goal-conditioned RL (OGCRL), offline safe RL (OSRL), offline goal-conditioned safe RL (OGCSRL), and decision transformer (DT) models. Figure~\ref{fig:crtieria-model}b summarizes which properties each class of methods satisfies. Additional discussion appears in Appendix~\ref{sec:related-work-appendix}.

\xhdr{Offline Goal-Conditioned RL}
OGCRL methods aim to learn policies that generalize to arbitrary goals specified at evaluation time, typically by conditioning on goal states and applying techniques such as hindsight goal relabeling~\citep{her}. 
\textbf{Reward-based OGCRL} methods~\citep{dqapg, iql-gc, gofar} optimize a policy $\pi(\mathbf{a}|\mathbf{s}, \mathbf{g})$ to maximize a goal-conditioned reward function $r(\mathbf{s}, \mathbf{a}, \mathbf{s}', \mathbf{g})$. While this framework supports goal generalization (P3), it does not satisfy P5, as it requires designing reward functions that capture both goal-reaching and avoid desires, which are difficult to construct in practice~\citep{rewardesignreview, howtoreward, complexrewardfunctionsandcurriculumlearning}. \textbf{Reward-free OGCRL} methods~\citep{contrastivelearningrl, park2023hiql, gcsl, wgcsl, lynch2019play, trajectorytransformer} avoid altogether reward functions by learning directly from hindsight-relabeled trajectories directly using supervised learning. This more modern wave of data-driven methods satisfy P5 but do not inherently support avoid-region conditioning (P4), as they cannot incorporate constraints beyond the goal.

\xhdr{Offline Safe RL}
OSRL methods~\citep{fisor, coptidice, constrained_qlearning, batchlearningcmdp, ieee} are designed for safety-critical tasks, learning policies that satisfy constraints specified via cost functions. A typical approach defines a cost function $\texttt{cost}(\mathbf{s}, \mathbf{a}, \mathbf{s}')$ and learns a policy that maximizes reward return $\sum_t r_t$ subject to a cost return constraint $\sum_t c_t < k$, often via Lagrangian relaxation~\citep{lagrangian}. These methods can enforce avoid behavior, but they are not generally goal-conditioned (P3), and do not support dynamic conditioning on varying avoid-region configurations (P4). In addition, these methods fail P5 due to their reliance on handcrafted reward/cost functions and constrained optimization setups with inflexible inherent priority hierarchies.

\xhdr{Offline Goal-Conditioned Safe RL}
This hybrid category combines elements of OGCRL and OSRL and comes closest to satisfying the full set of reach-avoid properties. Representative methods include Recovery-based Supervised Learning (RbSL) and Actionable Models with Lagrangian Constraints (AM-Lag)~\citep{cao2024offline, actionablemodels, lagrangian}. These models construct an augmented state space $\mathcal{S}^+ \subseteq \mathbb{R}^{d_s + d_s + n_{\text{avoid}} \cdot d_s}$ containing the agent's state $\mathbf{s} \in \mathbb{R}^{d_s}$, goal state $\mathbf{g} \in \mathbb{R}^{d_s}$, and $n_{\text{avoid}}$ avoid centroids $\texttt{centroid}(\mathbf{b}_j) \in \mathbb{R}^{d_s}$, then learn policies with goal-conditioned and avoid-conditioned objectives. This design satisfies P3 and P4.2, enabling generalization to arbitrary $\mathbf{g}$ and avoid-region locations. However, these methods do not satisfy P4.1, as the number of avoid-regions $n_{\text{avoid}}$ is fixed in the state space dimension, requiring retraining to accommodate more regions. They also fail to support P4.3, as the spatial extent of avoid-regions is encoded only in the cost function, requiring redefinition and retraining when region size changes. Furthermore, although these models could in principle satisfy P2.2 (learning from trajectories that violate avoid-regions), the training data in~\cite{cao2024offline} is collected from environments with impassable obstacles, meaning no training trajectories actually pass through avoid-regions (Figure~\ref{fig:reach}b), failing to demonstrate robustness to this type of data suboptimality. See Appendix~\ref{sec:expsetup-gym-appendix} for details.

\xhdr{Decision Transformers}
DT models~\citep{decisiontransformer, trajectorytransformer, onlinedt, wu2023elastic, cdt, cdt2, orlbudget} represent a class of offline RL approaches that frame policy learning as sequence modeling. DTs use causal transformer architectures that take as input a trajectory $\tau$ and autoregressively predict actions, $\pi(\mathbf{a}_t | \tau_{1:t-1}, \mathbf{s}_t)$. Prompting has recently been introduced as a mechanism to extend DTs to goal-conditioned settings~\citep{xu2022prompt, mgpo}. We build our method off of MGPO~\citep{mgpo}, which introduces prompt-based conditioning on arbitrary goals, enabling zero-shot generalization across goal states, but not avoid-regions.

\begin{figure}[!t]
\begin{centering}
\includegraphics[width=\textwidth]{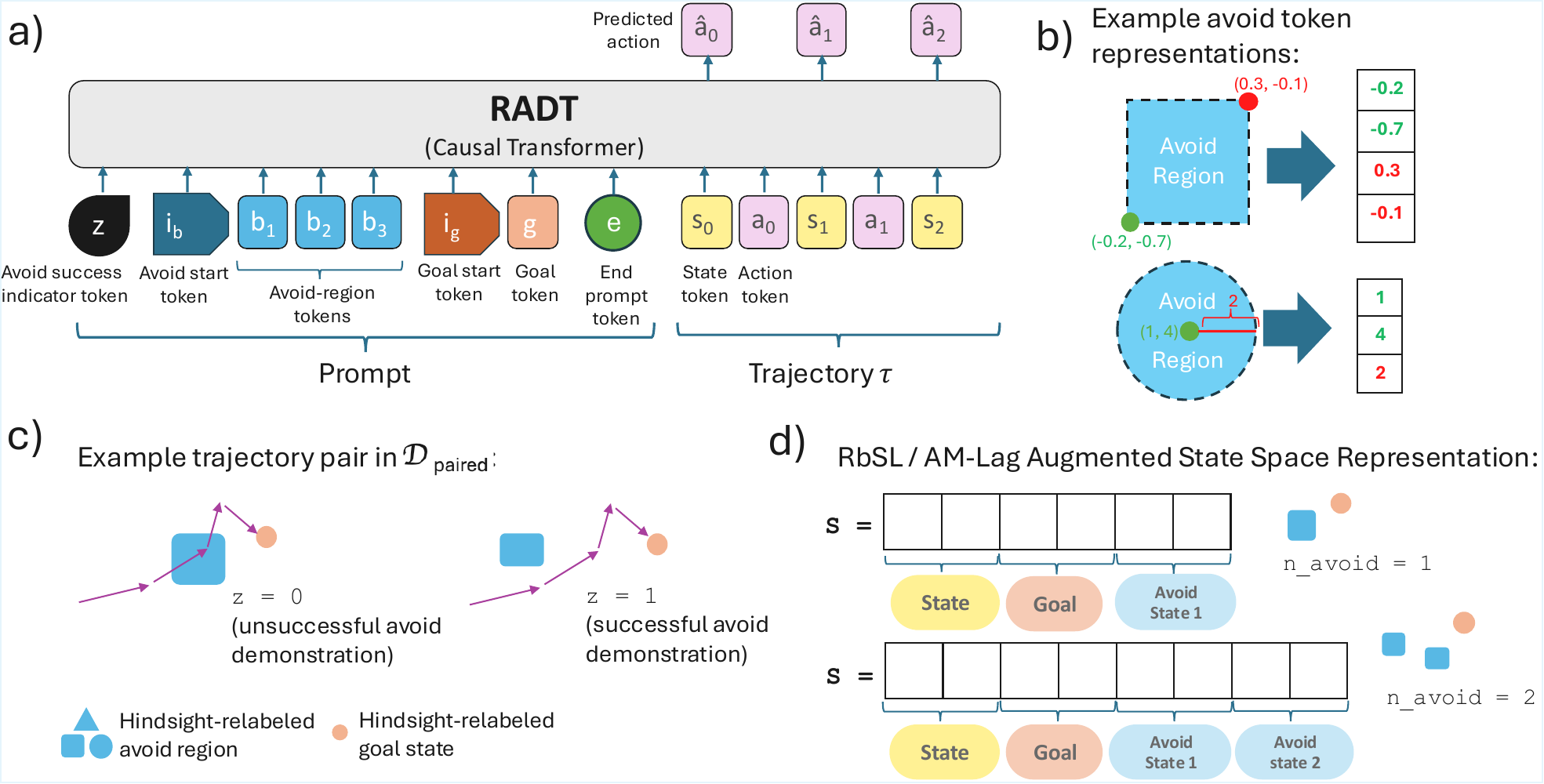}
\caption[]{(a) \name receives goal states and avoid-regions as prompt inputs.
(b) avoid-regions are represented as avoid-region tokens in the prompt, which can be any vector representation that contains position and size information. The box and spherical representations shown here are examples of valid representations.
(c) For each offline trajectory, we generate two versions: one that violates a sampled avoid-region and one that avoids it. Both are labeled with an avoid success token $z$. (d) Prior models encode avoid-regions via augmented state vectors, which grow with the number of avoid-regions, preventing zero-shot generalization to unseen avoid counts.}\label{fig:method}
\end{centering}
\end{figure}

\vspace{-1mm}
\section{Reach-Avoid Decision Transformer (\name)}\label{method}
\vspace{-1mm}

In this section, we describe the main components of our method, \name (Figure~\ref{fig:method}). Similar to MGPO, our method is based on a causal Transformer architecture and utilizes prompts to specify goal states (satisfying P3). However, unlike MGPO, \name additionally allows for the specification of avoid-regions in the prompt (satisfying P4) and does not require reward-driven online prompt optimization (satisfying P1, P2, P5).

\vspace{2mm}
\yhdr{Prompt Tokens and Avoid-Region Representation}\label{methodprompt}
\vspace{1mm}

\name takes in a prompt $p$ to be presented to the Transformer model before a trajectory $\tau$. \name's autoregressive prediction of the next action during rollout is thus additionally conditioned on this prompt: $\pi(\mathbf{a}_{t} | p, \tau_{0:t-1}, \mathbf{s}_t)$. The prompt is structured as follows (Figure~\ref{fig:method}a):
$$p = (z, i_b, \mathbf{b}_1, \mathbf{b}_2, ..., \mathbf{b}_{n_\text{avoid}}, i_g, \mathbf{g}, e)$$
Note that there are six different types of tokens present in the prompt:

\textcircled{\small{1}}~\textbf{The avoid success indicator (SI) token, $z: z \in \{0, 1\}$}, indicates whether the trajectory $\tau$ following the prompt successfully circumvents the avoid state: $z = 1$ if all states in $\tau$ exist outside of all avoid-regions $\mathbf{b}_j: j \in \{1,2,...,n_\text{avoid}\}$ and $z = 0$ otherwise.
This allows us to train the model on trajectories that demonstrate both successful and unsuccessful avoid behaviors; this is important for the model to explicitly learn what \textit{not} to do (see Section \ref{experiments}, Results 4). During evaluation time, we \textit{always} condition on $z = 1$ to achieve optimal avoid behavior. 
\textcircled{\small{2}}~\textbf{The avoid start (AS) token}, $i_b$, explicitly indicates to the model that the upcoming tokens represent undesirable avoid-regions. This is to clearly distinguish the avoid-region tokens from the SI token. 
\textcircled{\small{3}}~\textbf{The avoid-region tokens}, $\mathbf{b}_j \in \mathcal{S} \subseteq \mathbb{R}^{d_a}: j \in \{1, 2, ..., n_\text{avoid}\}$, represents the $n_\text{avoid}$ avoid-regions in the state space we would like the agent to circumvent. Our approach is not constrained to any particular vector representation of avoid-regions, as long as the representation contains information about both the position (the centroid) and the spatial extent of the avoid-region. Detailed overview of the representations we use can be found in Appendix~\ref{sec:data-prep-avoidrepresentation-appendix}. Since prompts can consist of any arbitrary number of avoid-region tokens $\mathbf{b}_j$ and the avoid-region tokens can represent regions of any arbitrary size at evaluation time, this satisfies P4.1, P4.2, and P4.3. 
\textcircled{\small{4}}~\textbf{The goal start (GS) token}, $i_g$, explicitly indicates to the model that the next token to be provided represents a desirable goal token. This is to clearly distinguish the goal token in the prompt from the avoid-region tokens.
\textcircled{\small{5}}~\textbf{The goal token}, $\mathbf{g} \in \mathcal{S} \subseteq \mathbb{R}^{d_s}$, represents the goal state we would like the agent to achieve. This can be set to any state in the state space at evaluation time (satisfying P3).
\textcircled{\small{6}}~\textbf{The prompt end token}, $e$, explicitly marks the end of the prompt, indicating that the next token marks the beginning of the main input sequence, $\tau$.

The initial representations of tokens of different types have different dimensionalities, as described above, but they are all projected into a shared latent embedding space to acquire common dimensionality before being passed through the rest of the transformer architecture. See Appendix~\ref{sec:modeldetails-model-appendix} for details regarding how these different token types are embedded.
Since we use prompts to specify the goal and avoid-region information, we do not need to work with an augmented state space $\mathcal{S}^+$, unlike RbSL and AM-Lag (see Section~\ref{relatedwork}), providing us with greater zero-shot flexibility.

 %

\vspace{2mm}
\yhdr{Avoid-Region Relabeling and Training 
\vspace{1mm}
Data}\label{methodavoidrelabeling}

We specifically consider the scenario in which the training dataset $\mathcal{D}$ contains training trajectories $\tau$ that are generated from a purely random policy (satisfying P1 and P2). For each training trajectory $\tau \in \mathcal{D}$, we relabel the last state $\mathbf{s}_T$ as the goal state $\mathbf{g}$ in hindsight. Additionally, we can also relabel \textit{avoid-regions} in hindsight, a novel strategy that gets rid of the need to use a cost function to learn desirable avoid behavior. The intuition for hindsight avoid-region relabeling (HAR) is similar to goal relabeling: it does not matter whether the policy that collected the training trajectory was actually intending to circumvent the hindsight-relabeled avoid-region; we can still treat it as if the region was "meant" to be avoided, as the trajectory demonstrates how to \textit{not} pass through that region. 

The details of our HAR approach are described in Appendix~\ref{sec:data-prep-relabeling-appendix}. HAR produces a dataset $\mathcal{D}_\text{paired}$ that contains a pair of trajectories $(\tau_\text{orig}, \tau_\text{copy})$ for each trajectory $\tau$ in the original dataset $\mathcal{D}$. One trajectory in each pair contains relabeled avoid regions that the trajectory successfully avoids and receives the label $z = 1$. The other contains avoid regions that the trajectory violates and receives the label $z = 0$. The two trajectories in a pair share the same states and actions, so the label $z$ separates avoidance success and failure from differences between the trajectories (Figure~\ref{fig:method}c). We observe that this pairing enables credit assignment to behaviors that lead to avoidance violations, and the model learns to reduce these behaviors (Section~\ref{experiments}, Results 4). At evaluation/deployment, we condition the model on $z = 1$. We format trajectories in $\mathcal{D}_\text{paired}$ with the prompt structure from the section above and train a causal transformer with a causal language modeling objective. Appendix~\ref{sec:modeldetails-appendix} provides the model and training specifications.

\vspace{-2mm}
\section{Experiments}\label{experiments}
\vspace{-2mm}
We evaluate \name in three sets of reach-avoid environments: \texttt{FetchReachObstacle}, \texttt{MazeObstacle}, and \texttt{Cardiogenesis}. The first two are adapted from Gymnasium Robotics tasks~\citep{gymrobotics}, but with added avoid-regions. In these environments, we compare \name against two OGCSRL baselines (RbSL~\citep{cao2024offline} and AM-Lag~\citep{cao2024offline, actionablemodels}) as well as Weighted Goal-Conditioned Supervised Learning (WGCSL)~\citep{wgcsl}, a strong OGCRL method that does not explicitly account for avoid-regions. To evaluate domain generality, we also test \name in a biological setting: cell reprogramming. Additionally, we perform a series of ablation and stress-testing experiments to validate the design decisions in our approach and evaluate how \name reacts under more extreme out-of-distribution (OOD) avoid specifications, varying mixtures of random vs. expert data, and alternative representations for avoid-regions. Additional task/environment details are provided in Appendix~\ref{sec:expsetup-appendix}.


\xhdr{Evaluation framework}
We treat reach-avoid problem as the Pareto optimization problem $\max_{\theta}\ \big(\, J_{\text{reach}}(\theta), J_{\text{avoid}}(\theta) \,\big)$ with two separate objectives to optimize for that may at times be at conflict with each other: (1) the goal-reaching (“reach”) objective $J_{\text{reach}}(\theta) = \mathbb{E}_{\tau \sim \pi_\theta(\cdot \mid g, B)} \big[\, G(\tau, g) \,\big]$ and (2) the avoidance objective $J_{\text{avoid}}(\theta) = -\mathbb{E}_{\tau \sim \pi_\theta(\cdot \mid g, B)} \big[\, M(\tau, B) \,\big]$, where $g$ is the goal, $B = [b_1, b_2…]$ are the avoid-regions, $G$ is the goal-reaching indicator function with leniency $\delta$ defined as $G(\tau, g) = \mathbbm{1}\left[d(s_L, g) \le \delta \right]$ where $s_L$ is the last state in $\tau$, and $M$ is the MNC function defined as $M(\tau, B) = \frac{1}{|\tau|} \sum_{t=1}^{|\tau|} \mathbbm{1}\left[\, s_t \in \textstyle\bigcup_{j=1}^{n} b_j \,\right]$.
While the two objectives could be combined into a single scalar objective and evaluation metric (e.g., the constrained optimization objectives used in many OSRL works~\citep{fisor, coptidice, constrained_qlearning, batchlearningcmdp, ieee}), that would involve predetermining and implying an optimal tradeoff between the two objectives, which is subjective to different scenarios and user preferences; therefore, we do not inherently define the optimal tradeoff between the two objectives and leave define both objectives defined and evaluated separately. Thus, if one method excels on one objective but underperforms on another objective compared to a second method, there is no objective “better method” in such a case. However, if one method were to dominate on one objective without being significantly worse on the other objective, it is said the method is \textit{Pareto dominant}. Under this Pareto optimization framework, the following results show that \name is Pareto dominant to baselines on the majority of experimental settings, and is incomparable to the other methods on the rest (\textit{but is never Pareto-dominated by a baseline}). Further discussion in included in Appendix~\ref{sec:related-work-appendix} under "Pareto vs. Constrained Optimization Evaluation Framework."

\vspace{2mm}
\yhdr{Results 1: Fetch Reach Environment and Generalization to Varying Avoid-Region Sizes}\label{experimentsfetchreach}
\vspace{1mm}
%

\begin{figure}[!t]
\begin{centering}
\resizebox{\textwidth}{!}{\includegraphics[width=\textwidth]{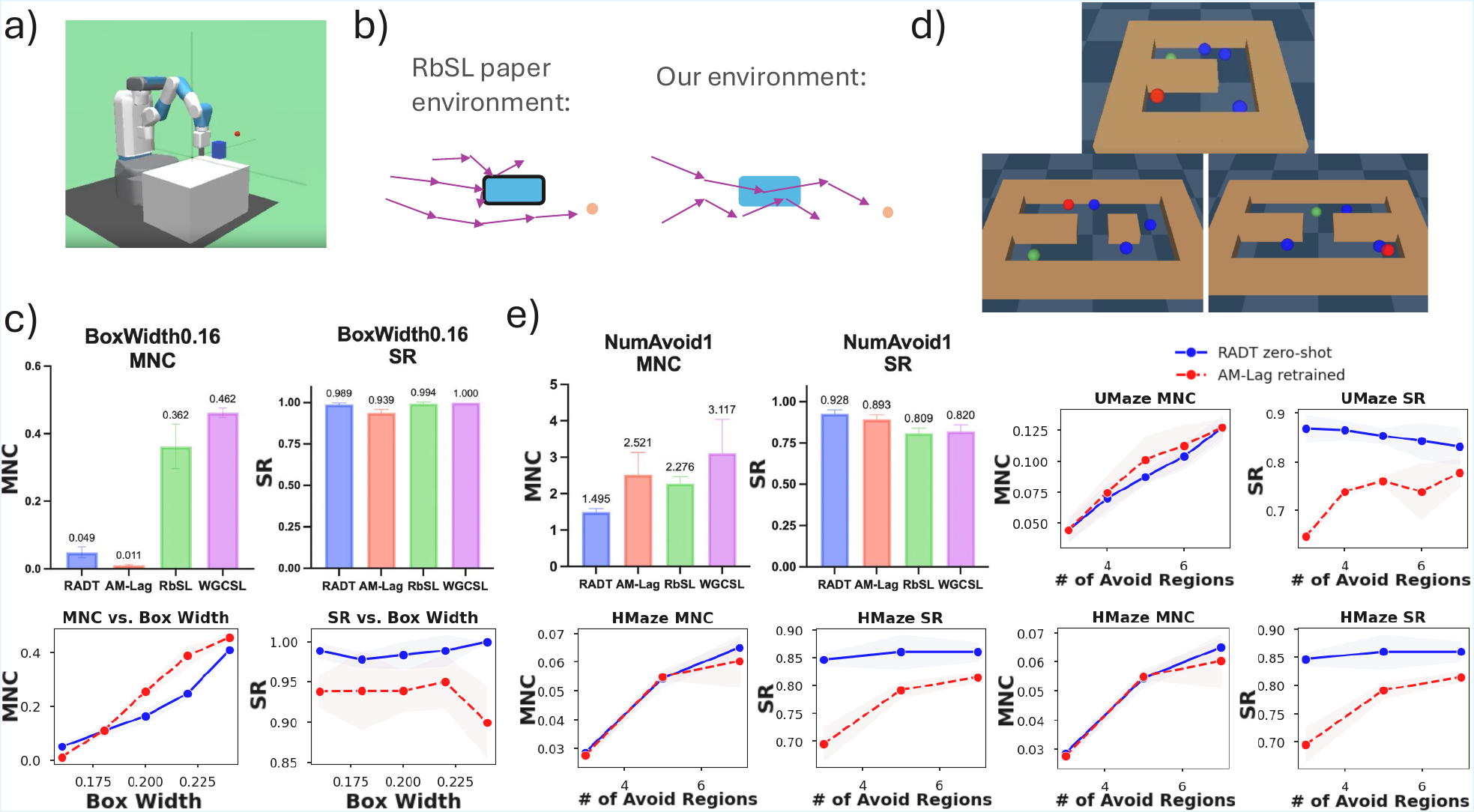}}
\caption[]{(a) Visualization of the \texttt{FetchReachObstacle} environment. The red point is the goal; the blue box is the avoid-region.
(b) Unlike prior setups, the robot arm can pass through avoid-boxes, allowing training data to include violations so that we can test the models' abilities to learn from suboptimal data.
(c) \name and AM-Lag achieve state-of-the-art MNC and SR on in-distribution box sizes and generalizes zero-shot to OOD avoid-box sizes, matching or surpassing the best baseline (AM-Lag), which needs to be \textit{retrained} on every new avoid-box size (i.e., not zero-shot).
(d) Visualization of the \texttt{UMazeObstacle}, \texttt{AMazeObstacle}, and \texttt{HMazeObstacle} environments, with red goal, blue avoid-regions, and green agent.
(e) \name outperforms all baselines on MNC and SR in the in-distribution single-avoid setting for \texttt{UMazeObstacle} and generalizes zero-shot to OOD numbers of avoid-regions on all maze tasks, matching the best \texttt{retrained} baseline (AM-Lag) in MNC and surpassing it in SR. Error bars show ±1 standard deviation.
}
\label{fig:reach} 
\end{centering}
\vspace{-5mm}
\end{figure}

\xhdr{Environment}
The \texttt{FetchReachObstacle} environment requires a robotic arm to reach a goal location while avoiding a randomly positioned box, treated as the avoid-region under our formulation (Figure~\ref{fig:reach}a). The positions of the end-effector, goal, and avoid-box are randomly sampled each episode. Unlike prior work such as~\cite{cao2024offline}, the robot can pass through the avoid-box, allowing us to isolate avoid-region reasoning without hard state space constraints (Figure~\ref{fig:reach}b).

\xhdr{Data}
We collect $2*10^6$ timesteps of random-policy trajectories. Goal relabeling and HAR is then applied, using a contour-based sampling strategy for diverse box placements (Appendix~\ref{sec:data-prep-relabeling-soph-appendix}).

\xhdr{Evaluation metrics}
While \name is trained without a cost function, we define one for evaluation: $\texttt{cost}(s_t, a_t, s_{t+1}) = 1$ if $s_{t+1}$ is inside an avoid-box and $0$ otherwise. We evaluate models using the mean normalized cost return (MNC), computed as the average per-step cost: $\text{MNC}(\tau) = \frac{1}{|\tau|} \sum_{(s,a,s') \in \tau} \texttt{cost}(s,a,s')$. We also report the goal-reaching success rate (SR), defined as the proportion of episodes that reach the goal. Evaluation is performed over 60 episodes.

\xhdr{Setup} We train all models using environments with a fixed avoid-box width (or, in the case of \name, a max hindsight-relabeled box width) of 0.16. RbSL, AM-Lag, and WGCSL are trained using the above cost function. We then evaluate all models in environments in-distribution (ID) with the same 0.16 avoid-box width. To assess generalization, we evaluate \name zero-shot on avoid-boxes with out-of-distribution (OOD) widths ranging from 0.16 to 0.24 (1.5× larger). In contrast, baseline models do not support zero-shot generalization to new avoid-box sizes and must be retrained on separate offline datasets generated for each new size (Sections~\ref{relatedwork} and~\ref{methodprompt}, and Appendix~\ref{sec:expsetup-gym-appendix}). This creates a disadvantageous setting for \name, which is evaluated without retraining, while baselines are retrained for each test condition.

\xhdr{Results}
Results are shown in Table~\ref{tab:table1}. For the ID case (box width 0.16), \name performs comparably to AM-Lag in MNC and better than RbSL and WGCSL. For SR, \name outperforms AM-Lag and matches or exceeds the other baselines (Figure~\ref{fig:reach}c). This confirms that both \name and AM-Lag are competitive for reach-avoid learning, with \name slightly favoring goal-reaching and AM-Lag slightly favoring cost minimization. In the OOD setting, where avoid-box sizes exceed those seen during training, \name continues to perform strongly despite being evaluated zero-shot. Across all OOD widths, \name matches or exceeds the MNC of retrained AM-Lag models and substantially outperforms retrained RbSL and WGCSL models. It also maintains comparable SR to the retrained baselines (Figure~\ref{fig:reach}c). These results show that \name generalizes to unseen avoid-region sizes, even outperforming methods that are \textit{retrained} for each evaluation setting.

\vspace{2mm}
\yhdr{Results 2: Maze Environments and Generalization to Varying Numbers of Avoid-Regions}\label{experimentsmaze}
\vspace{1mm}


\xhdr{Environment}
The \texttt{MazeObstacle} environments require a point agent to navigate through a maze to reach a randomly sampled goal location while avoiding randomly placed circular obstacles. As in \texttt{FetchReachObstacle}, these obstacles are soft constraints that the agent may pass through, and we refer to them as “avoid-regions.” Unlike in \texttt{FetchReachObstacle}, the maze itself imposes hard constraints, introducing impassable regions of the state space. This tests \name's ability to reason over both user-specified avoid-regions and inherent environmental constraints. We evaluate \name on three environments of varying difficulty: \texttt{UMazeObstacle}, \texttt{AMazeObstacle} (2 dead ends), and \texttt{HMazeObstacle} (4 dead ends) (Figure~\ref{fig:reach}d). The Data/Evaluation pipelines are similar to those of \texttt{FetchReachObstacle}, with details in Appendix~\ref{sec:expsetup-gym-appendix}.

\begin{table}[!t]
  \centering
  \caption{Results for \texttt{FetchReachObstacle} with varying box sizes (avg. over 3 seeds).}\label{tab:table1}
  \resizebox{\textwidth}{!}{%
  \begin{tabular}{*9c}
    \toprule
    \multicolumn{1}{c}{\textbf{Box Width}} & \multicolumn{2}{c}{$\text{\textbf{\name}}^\#$}  & \multicolumn{2}{c}{\textbf{AM-Lag}} & \multicolumn{2}{c}{\textbf{RbSL}} & \multicolumn{2}{c}{\textbf{WGCSL}} \\ \cmidrule(r){2-3} \cmidrule(l){4-5} \cmidrule(l){6-7} \cmidrule(l){8-9}
    \omit & \makecell{MNC} & \makecell{SR}  & \makecell{MNC} & \makecell{SR}   & \makecell{MNC} & \makecell{SR}    & \makecell{MNC} & \makecell{SR
} \\
    \hline
    0.16 & $\mathbf{0.049} {\scriptscriptstyle \pm0.016}$ &	$\mathbf{0.989} {\scriptscriptstyle \pm0.01} $ & $\mathbf{0.011} {\scriptscriptstyle \pm0.003}$	& $0.95	{\scriptscriptstyle \pm0.029}$ &	$0.362 {\scriptscriptstyle \pm 0.066}$ & $\mathbf{0.994}{\scriptscriptstyle \pm0.01}$	& $0.462	{\scriptscriptstyle \pm 0.0148}$ & $	\mathbf{1.0}	{\scriptscriptstyle \pm 0.0}$         \\
    0.18 & \cellcolor{blue!15}$\mathbf{0.107}	{\scriptscriptstyle \pm0.018}$ & \cellcolor{blue!15}$0.978	{\scriptscriptstyle \pm0.009}$	& $\mathbf{0.11}	{\scriptscriptstyle \pm0.01}$ & $0.95	{\scriptscriptstyle \pm0.033}$ 	& $0.484	{\scriptscriptstyle \pm0.016}$	& $\mathbf{1.0}	{\scriptscriptstyle \pm0}$	& $0.513	{\scriptscriptstyle \pm0.048}$	&  $\mathbf{0.994}	{\scriptscriptstyle \pm0.01}$ \\
    0.20 & \cellcolor{blue!15}$\mathbf{0.164}	{\scriptscriptstyle \pm0.027}$	& \cellcolor{blue!15}$\mathbf{0.983}	{\scriptscriptstyle \pm0.017}$ &$0.255	{\scriptscriptstyle \pm0.023}$ & $0.955	{\scriptscriptstyle \pm0.025}$	& $0.571	{\scriptscriptstyle \pm0.013}$ &	$\mathbf{1.0}	{\scriptscriptstyle \pm0.0}$ & $0.588	{\scriptscriptstyle \pm0.066}$	& $\mathbf{1.0}	{\scriptscriptstyle \pm0.0}$          \\
    0.22 &\cellcolor{blue!15} $\mathbf{0.247}	{\scriptscriptstyle \pm0.018}$ &\cellcolor{blue!15} $\mathbf{0.989}	{\scriptscriptstyle \pm0.019}$ &	$0.387	{\scriptscriptstyle \pm0.039}$ & $0.944	{\scriptscriptstyle \pm0.02}$	& $0.64	{\scriptscriptstyle \pm0.053}$ &	$\mathbf{0.99}	{\scriptscriptstyle \pm0.01}$ &	$0.699	{\scriptscriptstyle \pm0.019}$	& $\mathbf{0.994}	{\scriptscriptstyle \pm0.01}$          \\
    0.24 & \cellcolor{blue!15}$\mathbf{0.409}	{\scriptscriptstyle \pm0.012}$	&\cellcolor{blue!15} $\mathbf{1.0}	{\scriptscriptstyle \pm0.0}$	& $\mathbf{0.457}	{\scriptscriptstyle \pm0.008}$ &	$0.964	{\scriptscriptstyle \pm0.017}$ &	$0.701	{\scriptscriptstyle \pm0.019}$ & $	\mathbf{1.0}	{\scriptscriptstyle \pm0}$ &	$0.729	{\scriptscriptstyle \pm0.038}$ &	$\mathbf{0.989}	{\scriptscriptstyle \pm0.01}$          \\
    \bottomrule
  \end{tabular}}
  {\footnotesize $^\# = \text{Model capable of zero-shot generalization.}$ 
  \colorbox{blue!15}{Results highlighted in blue are zero-shot results.}}
\vspace{-3mm}
\end{table}
\begin{table}[t]
  \centering
  \caption{Results for \texttt{MazeObstacle} with varying number of avoid states (avg. over 3 seeds).}\label{tab:table2}
  \resizebox{\columnwidth}{!}{%
  \begin{tabular}{*9c}
    \toprule
    \multicolumn{9}{c}{\textbf{UMazeObstacle}} \\
    \cmidrule(r){2-9}
    \multicolumn{1}{c}{\textbf{\# Avoid}} & \multicolumn{2}{c}{$\text{\textbf{\name}}^\#$}  & \multicolumn{2}{c}{\textbf{AM-Lag}} & \multicolumn{2}{c}{\textbf{RbSL}} & \multicolumn{2}{c}{\textbf{WGCSL}} \\ \cmidrule(r){2-3} \cmidrule(l){4-5} \cmidrule(l){6-7} \cmidrule(l){8-9}
    \omit & \makecell{MNC (1e-2)} & \makecell{SR}  & \makecell{MNC (1e-2)} & \makecell{SR}   & \makecell{MNC (1e-2)} & \makecell{SR}    & \makecell{MNC (1e-2)} & \makecell{SR} \\
    \hline
    1 & $\mathbf{1.495} {\scriptscriptstyle \pm0.096}$ &	$\mathbf{.928} {\scriptscriptstyle \pm0.022} $ & $2.521 {\scriptscriptstyle \pm0.613}$	& $0.893	{\scriptscriptstyle \pm0.029}$ &	$2.276 {\scriptscriptstyle \pm 0.193}$ & $0.809{\scriptscriptstyle \pm0.031}$	& $3.117	{\scriptscriptstyle \pm 0.922}$ & $	0.82	{\scriptscriptstyle \pm 0.041}$         \\
    \hline
    3 & $\mathbf{4.455}	{\scriptscriptstyle \pm0.895}$ & $\mathbf{0.868}	{\scriptscriptstyle \pm0.028}$	& $\mathbf{4.47}	{\scriptscriptstyle \pm0.94}$ & $0.645	{\scriptscriptstyle \pm0.01}$ 	& $5.857 {\scriptscriptstyle \pm0.754}$	& $0.175	{\scriptscriptstyle \pm0.054}$	& $6.92	{\scriptscriptstyle \pm0.404}$	&  $\mathbf{0.842}	{\scriptscriptstyle \pm0.033}$ \\
    4 & \cellcolor{blue!15}$\mathbf{7.006}	{\scriptscriptstyle \pm1.156}$	& \cellcolor{blue!15}$\mathbf{0.865}	{\scriptscriptstyle \pm0.02}$ &$7.511	{\scriptscriptstyle \pm1.342}$ & $0.738	{\scriptscriptstyle \pm0.006}$	& $7.648	{\scriptscriptstyle \pm0.357}$ &	$0.768	{\scriptscriptstyle \pm0.043}$ & $9.62	{\scriptscriptstyle \pm1.701}$	& $\mathbf{0.852}	{\scriptscriptstyle \pm0.043}$          \\
    5 &\cellcolor{blue!15} $\mathbf{8.75}	{\scriptscriptstyle \pm0.531}$ &\cellcolor{blue!15} $\mathbf{0.853}	{\scriptscriptstyle \pm0.015}$ &	$10.14	{\scriptscriptstyle \pm1.9}$ & $0.76	{\scriptscriptstyle \pm0.01}$	& $9.622	{\scriptscriptstyle \pm0.531}$ &	$0.053	{\scriptscriptstyle \pm0.012}$ &	$10.28	{\scriptscriptstyle \pm0.503}$	& $\mathbf{0.807}	{\scriptscriptstyle \pm0.033}$          \\
    6 & \cellcolor{blue!15}$\mathbf{10.38}	{\scriptscriptstyle \pm1.015}$	&\cellcolor{blue!15} $\mathbf{0.843}	{\scriptscriptstyle \pm0.038}$	& $11.278	{\scriptscriptstyle \pm1.629}$ &	$0.738	{\scriptscriptstyle \pm0.058}$ &	$11.35	{\scriptscriptstyle \pm1.604}$ & $	0.755	{\scriptscriptstyle \pm0.065}$ &	$12.364	{\scriptscriptstyle \pm1.414}$ &	$\mathbf{0.9}	{\scriptscriptstyle \pm0.017}$          \\
    7 & \cellcolor{blue!15}$\mathbf{12.7}	{\scriptscriptstyle \pm1.0}$	&\cellcolor{blue!15} $\mathbf{0.832}	{\scriptscriptstyle \pm0.038}$	& $\mathbf{12.72}	{\scriptscriptstyle \pm0.702}$ &	$0.777	{\scriptscriptstyle \pm0.028}$ &	$24.17	{\scriptscriptstyle \pm3.65}$ & $	0.002	{\scriptscriptstyle \pm0.003}$ &	$14.48	{\scriptscriptstyle \pm1.439}$ &	$\mathbf{0.825}	{\scriptscriptstyle \pm0.018}$          \\
    \hline \\
    \multicolumn{1}{c}{} & \multicolumn{4}{c}{$\text{\textbf{AMazeObstacle}}$}  & \multicolumn{4}{c}{\textbf{HMazeObstacle}} \\ 
    \cmidrule(r){2-5}  \cmidrule(l){6-9}
    \multicolumn{1}{c}{\textbf{\# Avoid}} & \multicolumn{2}{c}{$\text{\textbf{\name}}^\#$}  & \multicolumn{2}{c}{\textbf{AM-Lag}} & \multicolumn{2}{c}{$\text{\textbf{\name}}^\#$} & \multicolumn{2}{c}{\textbf{AM-Lag}} \\ 
    \cmidrule(r){2-3} \cmidrule(l){4-5} \cmidrule(l){6-7} \cmidrule(l){8-9}
    \omit & \makecell{MNC (1e-2)} & \makecell{SR}  & \makecell{MNC (1e-2)} & \makecell{SR}   & \makecell{MNC (1e-2)} & \makecell{SR}    & \makecell{MNC (1e-2)} & \makecell{SR} \\
    \cmidrule(r){2-5}  \cmidrule(l){6-9}
    3 & $\mathbf{3.1}	{\scriptscriptstyle \pm0.361}$ & $\mathbf{0.926}	{\scriptscriptstyle \pm0.006}$	& $4.467	{\scriptscriptstyle \pm0.252}$ & $0.805	{\scriptscriptstyle \pm0.02}$ 	& $\mathbf{2.867} {\scriptscriptstyle \pm0.153}$	& $\mathbf{0.847}	{\scriptscriptstyle \pm0.012}$	& $\mathbf{2.767}	{\scriptscriptstyle \pm0.208}$	&  $0.695	{\scriptscriptstyle \pm0.03}$ \\
    5 & \cellcolor{blue!15}$\mathbf{4.933}	{\scriptscriptstyle \pm0.252}$ & \cellcolor{blue!15}$\mathbf{0.908}	{\scriptscriptstyle \pm0.024}$	& $5.473	{\scriptscriptstyle \pm1.0}$ & $0.808	{\scriptscriptstyle \pm0.056}$ 	& \cellcolor{blue!15}$\mathbf{5.433} {\scriptscriptstyle \pm0.306}$	& \cellcolor{blue!15}$\mathbf{0.86}	{\scriptscriptstyle \pm0.03}$	& $\mathbf{5.5}	{\scriptscriptstyle \pm0.1}$	&  $0.792	{\scriptscriptstyle \pm0.014}$ \\
    7 & \cellcolor{blue!15}$\mathbf{7.0}	{\scriptscriptstyle \pm0.127}$ & \cellcolor{blue!15}$\mathbf{0.922}	{\scriptscriptstyle \pm0.013}$	& $\mathbf{6.933}	{\scriptscriptstyle \pm1.415}$ & $0.86	{\scriptscriptstyle \pm0.022}$ 	& \cellcolor{blue!15}$6.5 {\scriptscriptstyle \pm0.265}$	& \cellcolor{blue!15}$\mathbf{0.86}{\scriptscriptstyle \pm0.017}$	& $\mathbf{6.03} {\scriptscriptstyle \pm0.907}$	&  $0.815	{\scriptscriptstyle \pm0.009}$ \\
    
    \bottomrule
  \end{tabular}}
\vspace{-4.5mm}
\end{table}

\xhdr{Setup}
We begin by training all models using data generated in environments with a single avoid-region, and evaluate in matching single-avoid-region settings (ID). To evaluate generalization, we then train and evaluate all models in environments with three avoid-regions, followed by testing in environments with 4-7 avoid-regions to assess zero-shot OOD performance. For each new number of avoid-regions, baseline models (RbSL, AM-Lag, WGCSL) are retrained with an appropriately expanded state space (Figure~\ref{fig:method}d; see Sections~\ref{relatedwork} and~\ref{methodprompt}). In contrast, \name is evaluated zero-shot without any retraining. This design intentionally favors the baselines, as they are tailored to each test setting, whereas \name is held fixed across all configurations. Only the best-performing baseline in \texttt{UMazeObstacle} (AM-Lag) is evaluated on \texttt{AMazeObstacle} and \texttt{HMazeObstacle}.

\xhdr{Results}
Results are shown in Table~\ref{tab:table2}. Across all settings of \texttt{UMazeObstacle} (both ID and OOD), \name achieves the lowest MNC, consistently outperforming all baselines. It also outperforms AM-Lag and RbSL on SR, and performs comparably or better than WGCSL. Importantly, for all evaluations with more than three avoid-regions, \name is deployed zero-shot, while the baselines are retrained. Despite this disadvantage, \name matches or exceeds their performance, demonstrating its ability to generalize to OOD numbers of avoid-regions without retraining.

In \texttt{AMazeObstacle} and \texttt{HMazeObstacle}, \name outperforms AM-Lag in goal-reaching SR for every number of avoid regions. In MNC, \name outperforms or matches AM-Lag, and its advantage decreases at 5 and 7 avoid regions, which are OOD for \name and ID for AM-Lag.


\vspace{2mm}
\yhdr{Results 3: Applications in Biology: Zero-Shot Avoidance in Stochastic Cell Reprogramming}\label{experimentsreprogramming}
\vspace{1mm}

\begin{figure}[!t]
\begin{centering}
\resizebox{\columnwidth}{!}{\includegraphics[width=\textwidth]{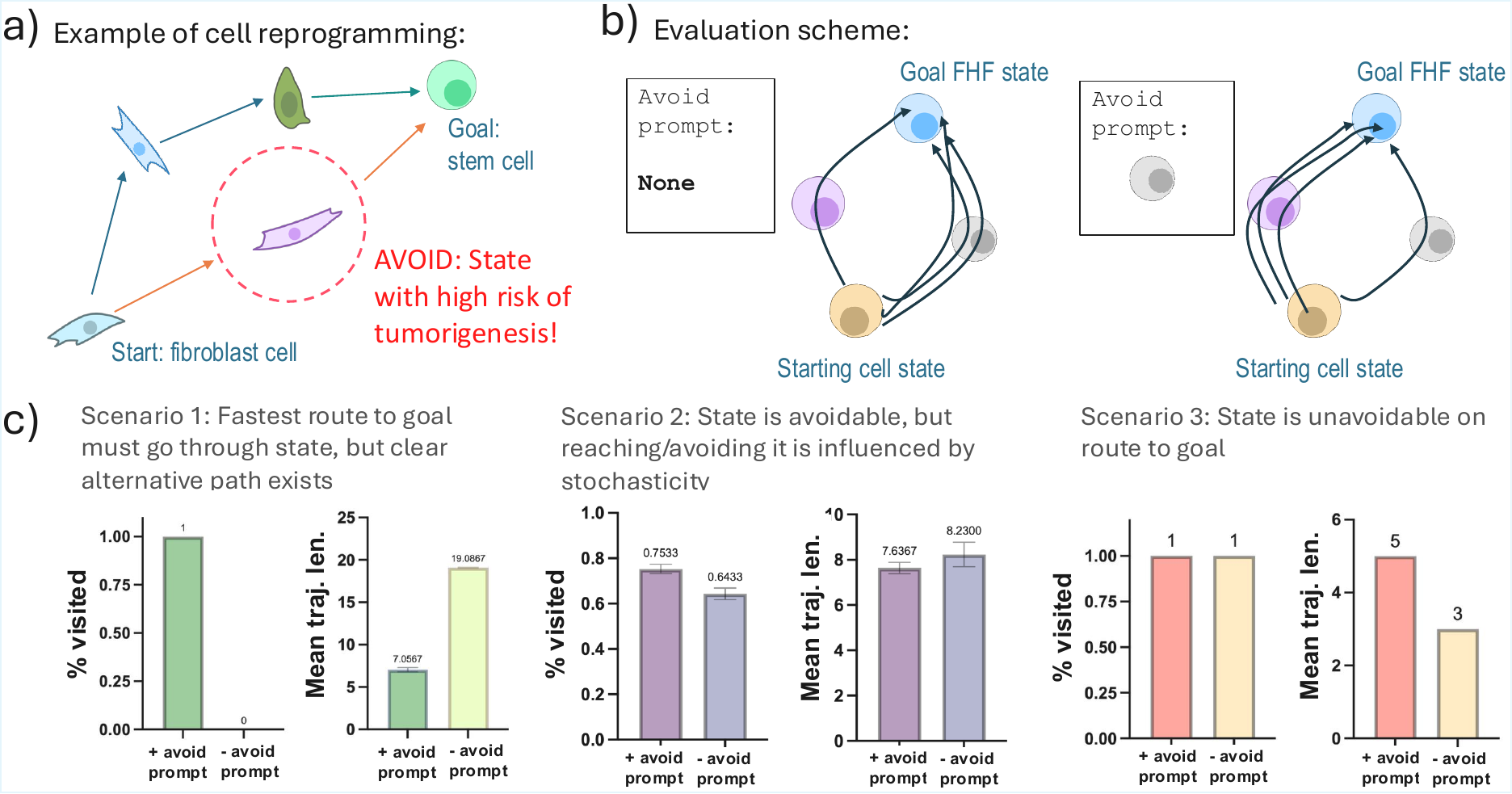}}
\caption[]{{(a) Cell reprogramming involves sequential perturbations to reach a target gene expression state while avoiding undesirable intermediate states. (b) Evaluation pipeline: \name is first run without an avoid-region token. The most frequently visited intermediate state is then added as an avoid-region token, and \name is re-evaluated. (c) \name reduces visitation rate to specified avoid states and minimizes time spent in unavoidable avoid states. Error bars show ±1 standard deviation. Some illustrations adapted from NIAID NIH BIOART (Appendix~\ref{sec:references-appendix}).}}\label{fig:cardiogenesis}
\vspace{-5mm}
\end{centering}
\end{figure}

We apply \name to a biomedical problem: cell reprogramming. The goal is to transition a cell from one gene expression state to another using sequences of genetic perturbations. This technique underpins regenerative medicine~\citep{Wan2023-uk,Vasan2021-oq}, stem cell therapy~\citep{yamanaka, fastchem}, and anti-aging strategies~\citep{Paine2024-fn, Pereira2024-kf}. However, intermediate gene expression states encountered during reprogramming may carry risks, e.g. tumorigenesis~\citep{tumorgenesis,stratsfortumorgenesis}. Safe reprogramming fits the reach-avoid formulation: reach a target state while avoiding undesirable intermediate states (Figure~\ref{fig:cardiogenesis}a).

\xhdr{Environment}
We introduce \texttt{Cardiogenesis}, an environment based on a well-established Boolean network model of gene expression dynamics during mouse cardiogenesis~\citep{booleannetworkcardiogenesis, cardiogenesisapp}. This model is described in detail in Appendix~\ref{sec:expsetup-cardio-appendix}. The key takeaway is that, unlike the Gymnasium Robotics environments, this domain features: (1) a fully discrete and combinatorial state-action space, (2) high-dimensional interdependencies between state variables due to Boolean logic, and (3) stochastic transitions. While \name can be trivially adapted to this very different state-action structure, our primary baseline methods cannot, as we show in Appendix~\ref{sec:expsetup-cardio-appendix} under "Baselines." See Appendix~\ref{sec:expsetup-cardio-appendix} for full environment specifications.

\xhdr{Data}
We generate $6*10^4$ random-policy timesteps of training data. Our HAR approach (Section~\ref{methodavoidrelabeling}) is directly applicable to this discrete setting with minimal modification (Appendix~\ref{sec:expsetup-cardio-appendix}).

\xhdr{Setup}
We conduct a case study where the reprogramming goal is to reach the first heart field (FHF) state, a critical attractor in cardiac development~\citep{booleannetworkcardiogenesis, cardiogenesisapp}. We train \name on the offline data, then evaluate its ability to reach the FHF state starting from a distinct attractor state (Appendix~\ref{sec:expsetup-cardio-appendix}). At first pass, we run 200 evaluation episodes with no avoid-region specified. From these, we identify the most frequently visited intermediate state on the way to the goal. We then run a second set of 200 episodes, this time providing the most visited intermediate state as an \textit{avoid-region token} in the prompt. We evaluate whether \name is able to discover alternative reprogramming paths that avoid this state and compare both visitation rates and trajectory lengths. This setup is illustrated in Figure~\ref{fig:cardiogenesis}b. See Appendix~\ref{sec:expsetup-cardio-appendix} for details. 

\xhdr{Results}
All evaluation trajectories successfully reach the FHF goal state, so we report only intermediate state visitation and trajectory lengths. Figure~\ref{fig:cardiogenesis}c summarizes results. We observe three distinct behavioral patterns, depending on the initial state. 
\textbf{Scenario 1.} The most visited intermediate (state A) is encountered in every trajectory when no avoid-region token is used. When A is included as an avoid-region token, \name avoids it entirely and instead follows alternative, longer trajectories that always bypass A. 
\textbf{Scenario 2.} Without an avoid-region token, the most visited intermediate state (state B) appears in a large fraction of trajectories, and the remaining trajectories reach the goal state through alternative paths. State B is a frequent transition under the stochastic dynamics of the environment, and reaching the goal state does not depend on it. When we condition \name on an avoid-region token for state B, the model reduces the visitation rate of state B in this noisy setting.
\textbf{Scenario 3.} The most visited state (state C) is present in all trajectories regardless of prompt. When C is added as an avoid-region token, \name cannot avoid passing through it, indicating that it is structurally unavoidable from the given initial state. However, \name reduces the number of steps spent in C, shortening the portion of the trajectory that includes it. This suggests that even when avoidance is infeasible, the model learns to minimize time spent in unsafe states.
These results highlight that \name supports reach-avoid planning in discrete, structured, and stochastic domains, and also exhibits flexible avoidance strategies, including temporal minimization of contact with undesirable states when full avoidance is not possible. Full results are in Appendix~\ref{sec:expsetup-cardio-appendix}.

%
\vspace{2mm}
\yhdr{Results 4: Additional Stress-Testing and Ablation Experiments}\label{experimentsablation}
\vspace{1mm}

We conduct ablation studies to highlight the importance of the SI, AS, and GS prompt tokens and the usage of attention boosting to the prompt. We also perform additional stress-testing experiments to examine the extremities of \name's OOD generalization, the empirical performance ceiling induced by the usage of random-policy data, and the \name's robustness to changing avoid-region token representations. We summarize the results here, and leave the detailed experimental setup, results, and discussion in Appendix~\ref{sec:newexp-appendix-ablation},\ref{sec:newexp-appendix-oodlimits},\ref{sec:newexp-appendix-expertmix},\ref{sec:newexp-appendix-alternativerep} and Tables~\ref{tab:table5},\ref{tab:table6},\ref{tab:table7},\ref{tab:table8},\ref{tab:table9},\ref{tab:table10} in Appendix~\ref{sec:figures-appendix}.

\xhdr{Ablation experiments} We perform ablation experiments of the SI, AS, GS, and avoid-region tokens, avoid-region relabeling, and attention boosting. Performance drops with each of these ablations the importance of these elements to our methodological design. (Fig.~\ref{fig:adelta}, Tables~\ref{tab:table5},\ref{tab:table6},\ref{tab:table10}).

\xhdr{Extremities of OOD generalization} To test the extremities \name's zero-shot generalization, we test zero-shot performance up to 20 avoid-regions. We find that zero-shot \name is still able to maintain comparable or superior MNC to the best \textit{retrained} baseline at 10 avoid-regions ($>$300\% of the maximum seen during training), demonstrating strong OOD generalization (Table~\ref{tab:table7}). Notably, zero-shot \name maintains a superior SR to retrained AM-Lag all the way up to 20 avoid-regions. 

\xhdr{Effects of expert-policy data} To explore the performance ceiling induced by using purely random-policy data, we examine the effects of introducing varying amounts of expert-policy trajectories into the training data. We observe that the introduction of increasing amounts of expert-policy data does not significantly impact performance on ID box sizes, but does improve avoid ability (MNC) while \textit{impairing} goal-reaching SR on OOD box sizes (Table~\ref{tab:table8}). This suggests the particular expert trajectories used biases \name to be more conservative in OOD avoid configurations.

\xhdr{Effects of changing avoid-region representation}
We test \name's flexibility with regards to handling various avoid-region token representations. We demonstrate that using an alternative representation for avoid-region tokens does not prevent \name from learning strong reach-avoid policies, establishing that our approach is agnostic to avoid-region token representation (Table~\ref{tab:table9}). 

\xhdr{Simultaneous varying of multiple avoid-region axes}
We show that \name is capable of responding to variation in multiple axes of avoid-region specification simultaneously using a navigation environment that generates a random number (ranging from 2 to 4) of avoid-regions, each with random location, aspect ratio (from 1:1 to 1:2.5), and size (side lengths from 0.24 to 0.6). As shown in Table~\ref{tab:table10.1}, \name outperforms baselines in MNC and SR on this task.

\xhdr{Image-based observation spaces and avoid-region representations}
To explore \name's potential in handling 1) high-dimensional observation spaces and 2) unstructured, flexible avoid-region token representations that can capture abstract avoidance desires without knowledge of a pre-defined vectorization, we conduct a preliminary proof-of-concept demonstration of \name applied to an image-based version of \texttt{UMazeObstacle} that we call \texttt{UMazeObstacleImage} (Fig.~\ref{fig:fig12}). In this environment, we present states, avoid-region tokens, and goal tokens as images encoded using a ResNet-18 model \citep{resnetcvpr,torchvision2016}. \name achieves superior MNC and comparable or superior goal-reaching SR to baselines in this environment, demonstrating its potential in being adapted to both high-dimensional tasks and flexible avoid-region representations that do not require precise knowledge of a strict, vectorized structure to specify avoid-regions (Table~\ref{tab:table11}). Experiment setup is described in Appendix~\ref{sec:newexp-appendix-image}.


\vspace{-2mm}
\section{Conclusion}\label{conclusion}
\vspace{-2mm}

We introduce \name, a prompting-based offline RL approach for reach-avoid tasks that satisfies all key desiderata for flexible reach-avoid learning. \name is trained entirely on suboptimal data using hindsight relabeling and generalizes zero-shot to unseen avoid-region configurations, achieving competitive or superior performance to SOTA methods retrained for each evaluation setting. \name is also domain-agnostic, demonstrating strong performance in both robotics and cell biology domains. We attribute \name's versatility and adaptability to its interpretable prompt-based design and fully data-driven, reward-free learning procedure. These properties enable \name to serve as a general-purpose framework for sequential decision-making under dynamically shifting avoidance desires. Limitations of our model are described in Appendix~\ref{sec:limitations-appendix}.

\clearpage

{\small \begin{ack}
We gratefully acknowledge support, in part, by NSF CAREER 2339524, Harvard Data Science Initiative, Amazon Faculty Research, Google Research Scholar Program, AstraZeneca Research, Roche Alliance with Distinguished Scientists (ROADS) Program, Sanofi iDEA-iTECH Award, GlaxoSmithKline Award, Boehringer Ingelheim Award, Merck Award, Optum AI Research Collaboration Award, Pfizer Research, Gates Foundation (INV-079038), Chan Zuckerberg Initiative, John and Virginia Kaneb Fellowship at Harvard Medical School, Biswas Computational Biology Initiative in partnership with the Milken Institute, Harvard Medical School Dean’s Innovation Fund for the Use of Artificial Intelligence, and the Kempner Institute for the Study of Natural and Artificial Intelligence at Harvard University. This work was delivered as part of the Aurora project supported by the Cancer Grand Challenges partnership funded by Cancer Research UK (CGCAI1-Mar26/100001).
Any opinions, findings, conclusions or recommendations expressed in this material are those of the authors and do not necessarily reflect the views of the funders.
All authors declare no competing interests.
\end{ack}}




\bibliography{references}

@inproceedings{
	mincostreachavoid,
	title={Solving Minimum-Cost Reach Avoid using Reinforcement Learning},
	author={So, Oswin and Ge, Cheng and Fan, Chuchu},
	booktitle={The Thirty-eighth Annual Conference on Neural Information Processing Systems},
	year={2024},
	url={https://openreview.net/forum?id=jzngdJQ2lY}
}

@INPROCEEDINGS{livelinessguarantees,
    AUTHOR    = {Kai-Chieh Hsu$^*$ and Vicenç Rubies-Royo$^*$ and Claire J. Tomlin and Jaime F. Fisac},
    TITLE     = {Safety and Liveness Guarantees through Reach-Avoid Reinforcement Learning},
    BOOKTITLE = {Proceedings of Robotics: Science and Systems},
    YEAR      = {2021},
    ADDRESS   = {Virtual},
    MONTH     = {July},
    DOI       = {10.15607/RSS.2021.XVII.077}
}

@misc{cbsreachavoid,
        title={Safe Multi-Agent Navigation guided by Goal-Conditioned Safe Reinforcement Learning}, 
        author={Meng Feng and Viraj Parimi and Brian Williams},
        year={2025},
        eprint={2502.17813},
        archivePrefix={arXiv},
        primaryClass={cs.RO},
        url={https://arxiv.org/abs/2502.17813}, 
  }

@article{cao2024offline,
  title={Offline Goal-Conditioned Reinforcement Learning for Safety-Critical Tasks with Recovery Policy},
  author={Cao, Chenyang and Yan, Zichen and Lu, Renhao and Tan, Junbo and Wang, Xueqian},
  journal={arXiv preprint arXiv:2403.01734},
  year={2024}
}

@article{actionablemodels,
    title={Actionable Models: Unsupervised Offline Reinforcement Learning of
           Robotic Skills},
    author={Yevgen Chebotar and Karol Hausman and Yao Lu and 
            Ted Xiao and Dmitry Kalashnikov and Jake Varley and
            Alex Irpan and Benjamin Eysenbach and Ryan Julian and 
            Chelsea Finn and Sergey Levine},
    journal={arXiv preprint arXiv:2104.07749},
    year={2021}
}

@ARTICLE{domainunlabeled,
  title        = "Offline reinforcement learning with domain-unlabeled data",
  author       = "Nishimori, Soichiro and Cai, Xin-Qiang and Ackermann,
                  Johannes and Sugiyama, Masashi",
  year         =  2024,
  primaryClass = "cs.LG",
  eprint       = "2404.07465"
}

@ARTICLE{limiteddata,
  title        = "Integrating domain knowledge for handling limited data in
                  offline {RL}",
  author       = "Gangopadhyay, Briti and Wang, Zhao and Yeh, Jia-Fong and
                  Takamatsu, Shingo",
  year         =  2024,
  primaryClass = "cs.LG",
  eprint       = "2406.07041"
}

@inproceedings{shouldidooffline,
title={Should I Run Offline Reinforcement Learning or Behavioral Cloning?},
author={Aviral Kumar and Joey Hong and Anikait Singh and Sergey Levine},
booktitle={International Conference on Learning Representations},
year={2022},
url={https://openreview.net/forum?id=AP1MKT37rJ}
}

@inproceedings{bcq,
  title={Off-Policy Deep Reinforcement Learning without Exploration},
  author={Fujimoto, Scott and Meger, David and Precup, Doina},
  booktitle={International Conference on Machine Learning},
  pages={2052--2062},
  year={2019}
}

@article{bear,
  author       = {Aviral Kumar and Justin Fu and George Tucker and Sergey Levine},
  title        = {Stabilizing Off-Policy Q-Learning via Bootstrapping Error Reduction},
  conference   = {NeurIPS 2019},
  year = 2019,
  url          = {http://arxiv.org/abs/1906.00949},
}

@inproceedings{ogbench,
  title={OGBench: Benchmarking Offline Goal-Conditioned RL},
  author={Park, Seohong and Frans, Kevin and Eysenbach, Benjamin and Levine, Sergey},
  booktitle={International Conference on Learning Representations (ICLR)},
  year={2025},
}

@ARTICLE{her,
  title        = "Hindsight Experience Replay",
  author       = "Andrychowicz, Marcin and Wolski, Filip and Ray, Alex and
                  Schneider, Jonas and Fong, Rachel and Welinder, Peter and
                  McGrew, Bob and Tobin, Josh and Abbeel, Pieter and Zaremba,
                  Wojciech",
  year         =  2017,
  primaryClass = "cs.LG",
  eprint       = "1707.01495"
}

@article{contrastivelearningrl,
  title={Contrastive Learning as Goal-Conditioned Reinforcement Learning},
  author={Eysenbach, Benjamin and Zhang, Tianjun and Salakhutdinov, Ruslan and Levine, Sergey},
  journal={arXiv preprint arXiv:2206.07568},
  year={2022}
}

@inproceedings{
park2023hiql,
title={{HIQL}: Offline Goal-Conditioned {RL} with Latent States as Actions},
author={Seohong Park and Dibya Ghosh and Benjamin Eysenbach and Sergey Levine},
booktitle={Thirty-seventh Conference on Neural Information Processing Systems},
year={2023},
url={https://openreview.net/forum?id=cLQCCtVDuW}
}

@article{gcsl,
title={Learning To Reach Goals Without Reinforcement Learning},
author={Dibya Ghosh and Abhishek Gupta and Justin Fu and Ashwin Reddy and Coline Devin and Benjamin Eysenbach and Sergey Levine},
journal={ArXiv},
year={2019},
volume={abs/1912.06088}
}

@inproceedings{
wgcsl,
title={Rethinking Goal-Conditioned Supervised Learning and Its Connection to Offline {RL}},
author={Rui Yang and Yiming Lu and Wenzhe Li and Hao Sun and Meng Fang and Yali Du and Xiu Li and Lei Han and Chongjie Zhang},
booktitle={International Conference on Learning Representations},
year={2022},
url={https://openreview.net/forum?id=KJztlfGPdwW}
}

@article{lynch2019play,
  title   = {Learning Latent Plans from Play},
  author  = {Lynch, Corey and Khansari, Mohi and Xiao, Ted and Kumar, Vikash
             and Tompson, Jonathan and Levine, Sergey and Sermanet, Pierre},
  journal = {Conference on Robot Learning (CoRL)},
  year    = {2019},
  url     = {https://arxiv.org/abs/1903.01973},
}

@inproceedings{trajectorytransformer,
  title = {Offline Reinforcement Learning as One Big Sequence Modeling Problem},
  author = {Michael Janner and Qiyang Li and Sergey Levine},
  booktitle = {Advances in Neural Information Processing Systems},
  year = {2021},
}

@ARTICLE{dqapg,
  title        = "Swapped goal-conditioned offline reinforcement learning",
  author       = "Yang, Wenyan and Wang, Huiling and Cai, Dingding and
                  Pajarinen, Joni and K{\"a}m{\"a}r{\"a}inen, Joni-Kristen",
  year         =  2023,
  primaryClass = "cs.LG",
  eprint       = "2302.08865"
}

@ARTICLE{iql-gc,
  title        = "Offline reinforcement learning with implicit Q-learning",
  author       = "Kostrikov, Ilya and Nair, Ashvin and Levine, Sergey",
  year         =  2021,
  primaryClass = "cs.LG",
  eprint       = "2110.06169"
}

@inproceedings{
gofar,
title={Offline Goal-Conditioned Reinforcement Learning via \$f\$-Advantage Regression},
author={Yecheng Jason Ma and Jason Yan and Dinesh Jayaraman and Osbert Bastani},
booktitle={Advances in Neural Information Processing Systems},
editor={Alice H. Oh and Alekh Agarwal and Danielle Belgrave and Kyunghyun Cho},
year={2022},
url={https://openreview.net/forum?id=_h29VprPHD}
}

@inproceedings{
mgpo,
title={Pre-Trained Multi-Goal Transformers with Prompt Optimization for Efficient Online Adaptation},
author={Haoqi Yuan and Yuhui Fu and Feiyang Xie and Zongqing Lu},
booktitle={The Thirty-eighth Annual Conference on Neural Information Processing Systems},
year={2024},
url={https://openreview.net/forum?id=DHucngOEe3}
}

@article{
  dsrl,
  title={Datasets and Benchmarks for Offline Safe Reinforcement Learning},
  author={Zuxin Liu and Zijian Guo and Haohong Lin and Yihang Yao and Jiacheng Zhu and Zhepeng Cen and Hanjiang Hu and Wenhao Yu and Tingnan Zhang and Jie Tan and Ding Zhao},
  journal={Journal of Data-centric Machine Learning Research},
  year={2024}
}

@techreport{bulletsafetygym,
	author = {Gronauer, Sven},
	institution = {mediaTUM},
	title = {Bullet-Safety-Gym: A Framework for Constrained Reinforcement Learning},
	year = {2022},
	doi = {10.14459/2022md1639974}}

@article{safetygymnasium,
    author = {Ray, Alex and Achiam, Joshua and Amodei, Dario},
    title = {Benchmarking Safe Exploration in Deep Reinforcement Learning},
    year = {2019}
}

@ARTICLE{lagrangian,
  title        = "Responsive safety in reinforcement learning by {PID}
                  Lagrangian methods",
  author       = "Stooke, Adam and Achiam, Joshua and Abbeel, Pieter",
  year         =  2020,
  primaryClass = "math.OC",
  eprint       = "2007.03964"
}

@inproceedings{
fisor,
title={Safe Offline Reinforcement Learning with Feasibility-Guided Diffusion Model},
author={Yinan Zheng and Jianxiong Li and Dongjie Yu and Yujie Yang and Shengbo Eben Li and Xianyuan Zhan and Jingjing Liu},
booktitle={The Twelfth International Conference on Learning Representations},
year={2024},
url={https://openreview.net/forum?id=j5JvZCaDM0}
}

@ARTICLE{constrained_qlearning,
  title     = "Constraints Penalized Q-learning for safe offline reinforcement
               learning",
  author    = "Xu, Haoran and Zhan, Xianyuan and Zhu, Xiangyu",
  journal   = "Proc. Conf. AAAI Artif. Intell.",
  publisher = "Association for the Advancement of Artificial Intelligence
               (AAAI)",
  volume    =  36,
  number    =  8,
  pages     = "8753--8760",
  month     =  jun,
  year      =  2022
}

@inproceedings{coptidice,
title={{CO}pti{DICE}: Offline Constrained Reinforcement Learning via Stationary Distribution Correction Estimation},
author={Jongmin Lee and Cosmin Paduraru and Daniel J Mankowitz and Nicolas Heess and Doina Precup and Kee-Eung Kim and Arthur Guez},
booktitle={International Conference on Learning Representations},
year={2022},
url={https://openreview.net/forum?id=FLA55mBee6Q}
}

@InProceedings{batchlearningcmdp,
  title = 	 {Batch Policy Learning under Constraints},
  author =       {Le, Hoang and Voloshin, Cameron and Yue, Yisong},
  booktitle = 	 {Proceedings of the 36th International Conference on Machine Learning},
  pages = 	 {3703--3712},
  year = 	 {2019},
  editor = 	 {Chaudhuri, Kamalika and Salakhutdinov, Ruslan},
  volume = 	 {97},
  series = 	 {Proceedings of Machine Learning Research},
  month = 	 {09--15 Jun},
  publisher =    {PMLR},
  url = 	 {https://proceedings.mlr.press/v97/le19a.html}
}

@inproceedings{
cdt,
title={Safe Decision Transformer with Learning-based Constraints},
author={Ruhan Wang and Dongruo Zhou},
booktitle={Neurips Safe Generative AI Workshop 2024},
year={2024},
url={https://openreview.net/forum?id=FPRHvvEAZf}
}

@article{decisiontransformer,
  title={Decision Transformer: Reinforcement Learning via Sequence Modeling},
  author={Lili Chen and Kevin Lu and Aravind Rajeswaran and Kimin Lee and Aditya Grover and Michael Laskin and Pieter Abbeel and Aravind Srinivas and Igor Mordatch},
  journal={arXiv preprint arXiv:2106.01345},
  year={2021}
}

@article{wu2023elastic,
  author={Wu, Yueh-Hua and Wang, Xiaolong and Hamaya, Masashi},
  title={Elastic Decision Transformer},
  archivePrefix = {arXiv},
  eprint = {2307.02484},
  primaryClass = {cs.LG},
  year={2023},
}

@article{onlinedt,
  publtype={informal},
  author={Qinqing Zheng and Amy Zhang and Aditya Grover},
  title={Online Decision Transformer},
  year={2022},
  cdate={1640995200000},
  journal={CoRR},
  volume={abs/2202.05607},
  url={https://arxiv.org/abs/2202.05607}
}

@inproceedings{xu2022prompt,
    title={Prompting Decision Transformer for Few-Shot Policy Generalization},
    author={Xu, Mengdi and Shen, Yikang and Zhang, Shun and Lu, Yuchen 
    and Zhao, Ding and Tenenbaum, B. Joshua and Gan, Chuang},
    booktitle={Thirty-ninth International Conference on Machine Learning},
    year={2022}
}

@INPROCEEDINGS{autonomousdrivingrewardreview,
  title           = "A review of reward functions for reinforcement learning in
                     the context of autonomous driving",
  booktitle       = "2024 {IEEE} Intelligent Vehicles Symposium ({IV})",
  author          = "Abouelazm, Ahmed and Michel, Jonas and Z{\"o}llner, J
                     Marius",
  publisher       = "IEEE",
  month           =  jun,
  year            =  2024,
  conference      = "2024 IEEE Intelligent Vehicle Symposium (IV)",
  location        = "Jeju Island, Korea, Republic of"
}

@ARTICLE{complexrewardfunctionsandcurriculumlearning,
  title        = "Curriculum reinforcement learning for complex reward
                  functions",
  author       = "Freitag, Kilian and Ceder, Kristian and Laezza, Rita and
                  {\AA}kesson, Knut and Chehreghani, Morteza Haghir",
  year         =  2024,
  primaryClass = "cs.LG",
  eprint       = "2410.16790"
}

@inproceedings{howtoreward,
title={How to Specify Reinforcement Learning Objectives},
author={W. Bradley Knox and James MacGlashan},
booktitle={Finding the Frame: An RLC Workshop for Examining Conceptual Frameworks},
year={2024},
url={https://openreview.net/forum?id=2MGEQNrmdN}
}

@article{rewardesignreview,
    author = {W. Bradley Knox and Alessandro Allievi and Holger Banzhaf and Felix Schmitt and Peter Stone},
    doi = {https://doi.org/10.1016/j.artint.2022.103829},
    issn = {0004-3702},
    journal = {Artificial Intelligence},
    pages = {103829},
    title = {Reward (Mis)design for autonomous driving},
    url = {https://www.sciencedirect.com/science/article/pii/S0004370222001692},
    volume = {316},
    year = {2023}}

@software{gymrobotics,
  author = {Rodrigo de Lazcano and Kallinteris Andreas and Jun Jet Tai and Seungjae Ryan Lee and Jordan Terry},
  title = {Gymnasium Robotics},
  url = {http://github.com/Farama-Foundation/Gymnasium-Robotics},
  version = {1.3.1},
  year = {2024},
}

@UNPUBLISHED{cardiogenesisapp,
  title    = "Optimizing sequential gene expression modulation for cellular
              reprogramming - Coupled Boolean modeling and Reinforcement
              Learning based method",
  author   = "Singh, Vivek",
  journal  = "bioRxiv",
  month    =  mar,
  year     =  2024
}

@ARTICLE{booleannetworkcardiogenesis,
  title     = "A boolean model of the cardiac gene regulatory network
               determining first and second heart field identity",
  author    = "Herrmann, Franziska and Gro{\ss}, Alexander and Zhou, Dao and
               Kestler, Hans A and K{\"u}hl, Michael",
  journal   = "PLoS One",
  publisher = "Public Library of Science (PLoS)",
  volume    =  7,
  number    =  10,
  pages     = "e46798",
  month     =  oct,
  year      =  2012,
  language  = "en"
}

@ARTICLE{fastchem,
  title     = "Chemical reprogramming of human somatic cells to pluripotent
               stem cells",
  author    = "Guan, Jingyang and Wang, Guan and Wang, Jinlin and Zhang,
               Zhengyuan and Fu, Yao and Cheng, Lin and Meng, Gaofan and Lyu,
               Yulin and Zhu, Jialiang and Li, Yanqin and Wang, Yanglu and
               Liuyang, Shijia and Liu, Bei and Yang, Zirun and He, Huanjing
               and Zhong, Xinxing and Chen, Qijing and Zhang, Xu and Sun,
               Shicheng and Lai, Weifeng and Shi, Yan and Liu, Lulu and Wang,
               Lipeng and Li, Cheng and Lu, Shichun and Deng, Hongkui",
  journal   = "Nature",
  publisher = "Springer Science and Business Media LLC",
  volume    =  605,
  number    =  7909,
  pages     = "325--331",
  month     =  may,
  year      =  2022,
  language  = "en"
}

@ARTICLE{yamanaka,
  title     = "Induction of pluripotent stem cells from mouse embryonic and
               adult fibroblast cultures by defined factors",
  author    = "Takahashi, Kazutoshi and Yamanaka, Shinya",
  journal   = "Cell",
  publisher = "Elsevier BV",
  volume    =  126,
  number    =  4,
  pages     = "663--676",
  month     =  aug,
  year      =  2006,
  copyright = "http://www.elsevier.com/open-access/userlicense/1.0/",
  language  = "en"
}

@ARTICLE{tumorgenesis,
  title     = "Prevention of tumor risk associated with the reprogramming of
               human pluripotent stem cells",
  author    = "Wuputra, Kenly and Ku, Chia-Chen and Wu, Deng-Chyang and Lin,
               Ying-Chu and Saito, Shigeo and Yokoyama, Kazunari K",
  journal   = "J. Exp. Clin. Cancer Res.",
  publisher = "Springer Science and Business Media LLC",
  volume    =  39,
  number    =  1,
  pages     = "100",
  month     =  jun,
  year      =  2020,
  copyright = "https://creativecommons.org/licenses/by/4.0",
  language  = "en"
}

@ARTICLE{stratsfortumorgenesis,
  title     = "Possible strategies to reduce the tumorigenic risk of
               reprogrammed normal and cancer cells",
  author    = "Lin, Ying-Chu and Ku, Cha-Chien and Wuputra, Kenly and Liu,
               Chung-Jung and Wu, Deng-Chyang and Satou, Maki and Mitsui, Yukio
               and Saito, Shigeo and Yokoyama, Kazunari K",
  journal   = "Int. J. Mol. Sci.",
  publisher = "MDPI AG",
  volume    =  25,
  number    =  10,
  pages     = "5177",
  month     =  may,
  year      =  2024,
  copyright = "https://creativecommons.org/licenses/by/4.0/",
  language  = "en"
}

@ARTICLE{Paine2024-fn,
  title     = "Partial cellular reprogramming: A deep dive into an emerging
               rejuvenation technology",
  author    = "Paine, Patrick T and Nguyen, Ada and Ocampo, Alejandro",
  journal   = "Aging Cell",
  publisher = "Wiley",
  volume    =  23,
  number    =  2,
  pages     = "e14039",
  month     =  feb,
  year      =  2024,
  copyright = "http://creativecommons.org/licenses/by/4.0/",
  language  = "en"
}

@ARTICLE{Pereira2024-kf,
  title     = "Epigenetic reprogramming as a key to reverse ageing and increase
               longevity",
  author    = "Pereira, Beatriz and Correia, Francisca P and Alves, In{\^e}s A
               and Costa, Margarida and Gameiro, Mariana and Martins, Ana P and
               Saraiva, Jorge A",
  journal   = "Ageing Res. Rev.",
  publisher = "Elsevier BV",
  volume    =  95,
  number    =  102204,
  pages     = "102204",
  month     =  mar,
  year      =  2024,
  copyright = "http://creativecommons.org/licenses/by/4.0/",
  language  = "en"
}

@ARTICLE{Wan2023-uk,
  title     = "Strategies and mechanisms of neuronal reprogramming",
  author    = "Wan, Yue and Ding, Yan",
  journal   = "Brain Res. Bull.",
  publisher = "Elsevier BV",
  volume    =  199,
  number    =  110661,
  pages     = "110661",
  month     =  jul,
  year      =  2023,
  copyright = "http://creativecommons.org/licenses/by/4.0/",
  language  = "en"
}

@ARTICLE{Vasan2021-oq,
  title     = "Direct neuronal reprogramming: Bridging the gap between basic
               science and clinical application",
  author    = "Vasan, Lakshmy and Park, Eunjee and David, Luke Ajay and
               Fleming, Taylor and Schuurmans, Carol",
  journal   = "Front. Cell Dev. Biol.",
  publisher = "Frontiers Media SA",
  volume    =  9,
  pages     = "681087",
  month     =  jul,
  year      =  2021,
  copyright = "https://creativecommons.org/licenses/by/4.0/",
  language  = "en"
}

@ARTICLE{ieee,
  author={Lin, Haohong and Ding, Wenhao and Liu, Zuxin and Niu, Yaru and Zhu, Jiacheng and Niu, Yuming and Zhao, Ding},
  journal={IEEE Robotics and Automation Letters}, 
  title={Safety-Aware Causal Representation for Trustworthy Offline Reinforcement Learning in Autonomous Driving}, 
  year={2024},
  volume={9},
  number={5},
  pages={4639-4646},
  doi={10.1109/LRA.2024.3379805}}

@inproceedings{cdt2, author = {Liu, Zuxin and Guo, Zijian and Yao, Yihang and Cen, Zhepeng and Yu, Wenhao and Zhang, Tingnan and Zhao, Ding}, title = {Constrained decision transformer for offline safe reinforcement learning}, year = {2023}, publisher = {JMLR.org}, booktitle = {Proceedings of the 40th International Conference on Machine Learning}, articleno = {893}, numpages = {20}, location = {Honolulu, Hawaii, USA}, series = {ICML'23} }

@inproceedings{orlbudget,
author = {Lin, Qian and Tang, Bo and Wu, Zifan and Yu, Chao and Mao, Shangqin and Xie, Qianlong and Wang, Xingxing and Wang, Dong},
title = {Safe offline reinforcement learning with real-time budget constraints},
year = {2023},
publisher = {JMLR.org},
booktitle = {Proceedings of the 40th International Conference on Machine Learning},
articleno = {871},
numpages = {26},
location = {Honolulu, Hawaii, USA},
series = {ICML'23}
}

@inproceedings{torchvision2016, author = {Marcel, S\'{e}bastien and Rodriguez, Yann}, title = {Torchvision the machine-vision package of torch}, year = {2010}, isbn = {9781605589336}, publisher = {Association for Computing Machinery}, address = {New York, NY, USA}, url = {https://doi.org/10.1145/1873951.1874254}, doi = {10.1145/1873951.1874254}, booktitle = {Proceedings of the 18th ACM International Conference on Multimedia}, pages = {1485–1488}, numpages = {4}, location = {Firenze, Italy}, series = {MM '10} }

@InProceedings{resnetcvpr,
author = {He, Kaiming and Zhang, Xiangyu and Ren, Shaoqing and Sun, Jian},
title = {Deep Residual Learning for Image Recognition},
booktitle = {Proceedings of the IEEE Conference on Computer Vision and Pattern Recognition (CVPR)},
month = {June},
year = {2016}
}

@inproceedings{
silva2024pay,
title={Pay Attention to What Matters},
author={Pedro Luiz Silva and Fadhel Ayed and Antonio De Domenico and Ali Maatouk},
booktitle={MINT: Foundation Model Interventions},
year={2024},
url={https://openreview.net/forum?id=F6o58A0OrX}
}

@article{radford2019language,
  title={Language Models are Unsupervised Multitask Learners},
  author={Radford, Alec and Wu, Jeff and Child, Rewon and Luan, David and Amodei, Dario and Sutskever, Ilya},
  year={2019}
}

@article{liaw2018tune,
    title={Tune: A Research Platform for Distributed Model Selection and Training},
    author={Liaw, Richard and Liang, Eric and Nishihara, Robert
            and Moritz, Philipp and Gonzalez, Joseph E and Stoica, Ion},
    journal={arXiv preprint arXiv:1807.05118},
    year={2018}
}

@inproceedings{wolf-etal-2020-transformers,
    title = "Transformers: State-of-the-Art Natural Language Processing",
    author = "Thomas Wolf and Lysandre Debut and Victor Sanh and Julien Chaumond and Clement Delangue and Anthony Moi and Pierric Cistac and Tim Rault and Rémi Louf and Morgan Funtowicz and Joe Davison and Sam Shleifer and Patrick von Platen and Clara Ma and Yacine Jernite and Julien Plu and Canwen Xu and Teven Le Scao and Sylvain Gugger and Mariama Drame and Quentin Lhoest and Alexander M. Rush",
    booktitle = "Proceedings of the 2020 Conference on Empirical Methods in Natural Language Processing: System Demonstrations",
    month = oct,
    year = "2020",
    address = "Online",
    publisher = "Association for Computational Linguistics",
    url = "https://www.aclweb.org/anthology/2020.emnlp-demos.6",
    pages = "38--45"
}

@misc{Katsura2021,
  author = {Katsura, Naoki and Baldassarre, Federico },
  title = {Cosine Annealing with Warmup for PyTorch},
  year = {2021},
  publisher = {GitHub},
  journal = {GitHub repository},
  howpublished = {\url{https://github.com/katsura-jp/pytorch-cosine-annealing-with-warmup.git}},
}

@book{convexconcave,
author = {Vinh, Phan and Dung, Nguyen},
year = {2023},
month = {04},
pages = {},
title = {Context-Aware Systems and Applications. 11th EAI International Conference, ICCASA 2022 Vinh Long, Vietnam, October 27–28, 2022 Proceedings.},
isbn = {978-3-031-28815-9}
}

@article{park2013,
author = {Park, J.-S and Oh, S.-J},
year = {2013},
month = {03},
pages = {379-392},
title = {A New Concave Hull Algorithm and Concaveness Measure for n-dimensional Datasets},
volume = {29},
journal = {Journal of Information Science and Engineering}
}

@misc{Tang2022,
  author = {Tang, Zhixiong},
  title = {concave\_hull},
  year = {2022},
  publisher = {GitHub},
  journal = {GitHub repository},
  howpublished = {\url{https://github.com/cubao/concave_hull.git}},
}

@ARTICLE{Graham1972-hm,
  title     = "An efficient algorith for determining the convex hull of a
               finite planar set",
  author    = "Graham, R L",
  journal   = "Inf. Process. Lett.",
  publisher = "Elsevier BV",
  volume    =  1,
  number    =  4,
  pages     = "132--133",
  month     =  jun,
  year      =  1972,
  language  = "en"
}

@ARTICLE{cmdp,
  title     = "Constrained Markov decision processes with total cost criteria:
               Occupation measures and primal {LP}",
  author    = "Altman, Eitan",
  journal   = "Math. Methods Oper. Res. (Heidelb.)",
  publisher = "Springer Science and Business Media LLC",
  volume    =  43,
  number    =  1,
  pages     = "45--72",
  month     =  feb,
  year      =  1996,
  language  = "en"
}

@misc{sparserewards,
      title={Learning Successor States and Goal-Dependent Values: A Mathematical Viewpoint}, 
      author={Léonard Blier and Corentin Tallec and Yann Ollivier},
      year={2021},
      eprint={2101.07123},
      archivePrefix={arXiv},
      primaryClass={cs.LG},
      url={https://arxiv.org/abs/2101.07123}, 
}

@inproceedings{
dinowm,
title={{DINO}-{WM}: World Models on Pre-trained Visual Features enable Zero-shot Planning},
author={Gaoyue Zhou and Hengkai Pan and Yann LeCun and Lerrel Pinto},
booktitle={Forty-second International Conference on Machine Learning},
year={2025},
url={https://openreview.net/forum?id=D5RNACOZEI}
}

@misc{pldm,
      title={Learning from Reward-Free Offline Data: A Case for Planning with Latent Dynamics Models}, 
      author={Vlad Sobal and Wancong Zhang and Kyunghyun Cho and Randall Balestriero and Tim G. J. Rudner and Yann LeCun},
      year={2025},
      eprint={2502.14819},
      archivePrefix={arXiv},
      primaryClass={cs.LG},
      url={https://arxiv.org/abs/2502.14819}, 
}

@inproceedings{cpo,
author = {Achiam, Joshua and Held, David and Tamar, Aviv and Abbeel, Pieter},
title = {Constrained policy optimization},
year = {2017},
publisher = {JMLR.org},
booktitle = {Proceedings of the 34th International Conference on Machine Learning - Volume 70},
pages = {22–31},
numpages = {10},
location = {Sydney, NSW, Australia},
series = {ICML'17}
}
\bibliographystyle{plain}

\clearpage
\appendix
\section*{Appendix}
\section{Data Preparation}\label{sec:data-prep-appendix} 
\subsection{Datasets}\label{sec:data-prep-dataset-appendix} 
The download link for the preprocessed datasets, as well as the setup and preprocessing code used to generate these datasets, will be released with the code repository for the project. See Appendix~\ref{sec:code-appendix}.

\subsection{Avoid Representation Details}\label{sec:data-prep-avoidrepresentation-appendix}
avoid-regions are represented in our experiments as \textit{box coordinates} (Fig 2b) by default, though any vector representation could work. The box coordinate vector of an avoid-region $\mathbf{b}_j \in \mathbb{R}^{2 * d_s}$ represents a "box" in the state space to be avoided. It is defined such that the first $d_s$ entries represent the lower bounds of each of the state space dimensions (the "lower left corner" of the box in a 3D analogy) and the second $d_s$ entries represent the upper bounds of each of the state space dimensions (the "upper right corner"):
$$\mathbf{b}_j = \underbrace{[l_1, l_2, ..., l_{d_s}}_\text{lower bounds }, \overbrace{u_{1},u_{2} ...,u_{d_s}}^\text{upper bounds}] $$
The policy should avoid guiding the agent into the region of the state space bounded by this box. To see how we adapt this box representation for non-rectangular avoid-regions in the \texttt{MazeObstacle} (circular avoid-regions) and \texttt{Cardiogenesis} environments (discrete avoid states), refer to the "Avoid-Region Relabeling" sections of Appendix~\ref{sec:expsetup-gym-appendix} and Appendix~\ref{sec:expsetup-cardio-appendix}, respectively.

In addition to the box representation, we also conduct experiments using the following representation for spherical avoid-regions to demonstrate \name's flexibility in handling different avoid-region token representations:
    $$\mathbf{b}_j = [b_1, b_2, ..., b_{d_s}, r]$$
where $(b_1, b_2, ..., b_{d_s})$ represents the location of $\texttt{centroid}(\mathbf{b}_j$) in the state space and $r$ represents the radius of the avoid-region (Figure~\ref{fig:method}b). Results for the experiment are found in Appendix~\ref{sec:newexp-appendix-alternativerep}. 

\subsection{Avoid Relabeling Details}\label{sec:data-prep-relabeling-appendix} 
\subsubsection{Avoid Relabeling Two-Pass Algorithm}
The two-pass hindsight avoid-region relabeling method introduced in Section~\ref{methodavoidrelabeling} is described in more detail here and visualized in Algorithm \ref{fig:algo1} (first pass) and Algorithm \ref{fig:algo2} (second pass).

For each $\tau^{(i)} \in \mathcal{D}$, we carry out HAR in two passes. In the first pass, we randomly sample avoid-boxes $\mathbf{b}_j: j \in \{1,2, ..., n_{\text{avoid}}\}$ of random sizes in the state space $\mathcal{S}$ and check whether any $s_t \in \tau^{(i)}$ violate any $\mathbf{b}_j: j \in \{1,2, ..., n_{\text{avoid}}\}$. If there are no violations, then the SI for $\tau^{(i)}$ is set to $z^{(i)} = 1$, otherwise it is set to $z^{(i)} = 0$. 
In the second pass, we create a copy of the dataset, $\mathcal{D_\text{copy}}$, and go through the same process above with the trajectories ${\tau_\text{copy}}^{(i)} \in \mathcal{D}_\text{copy}$. This time, however, for a trajectory ${\tau_\text{copy}}^{(i)}$, we keep re-sampling avoid-boxes until the avoid success token ${z_\text{copy}}^{(i)}$ is the \textit{opposite} of ${z}^{(i)}$ for the corresponding $\tau^{(i)}$ in the original $\mathcal{D}$.

Combining the datasets $\mathcal{D}$ and $\mathcal{D_\text{copy}}$ into $\mathcal{D_\text{paired}}$, we now have a \textit{pair} of trajectories $(\tau_\text{orig}^{(i)}, \tau_\text{copy}^{(i)})$ for each $\tau^{(i)}$ in the original dataset, where one of $(\tau_\text{orig}^{(i)}, \tau_\text{copy}^{(i)})$ demonstrates successful avoid behavior and the other demonstrates unsuccessful avoid behavior.

\begin{algorithm}[!b]
\caption{Hindsight Avoid-Region Relabeling: Initial Pass}\label{fig:algo1}
\begin{algorithmic}[1]
\Require state space $\mathcal{S}$, offline training dataset $\mathcal{D}$, maximum avoid-box width $w_{\text{max}}$, number of avoid states $n_{\text{avoid}}$
\For{training trajectory $\tau^{(i)}$ in $\mathcal{D}$}
\State Initialize avoid-regions list $\text{b\_list}^{(i)} \gets [ ]$
\For{$j$ in 1, 2, ..., $n_{\text{avoid}}$}
\State Randomly initialize avoid centroid $\mathbf{x}_j \in \mathcal{S}$
\State Randomly choose avoid-box width $w \in [0, w_{\text{max}}]$
\State avoid-box $\mathbf{b}_j \gets $ concatenate($\mathbf{x}_j - \frac{w}{2}$, $\mathbf{x}_j + \frac{w}{2}$ ) \Comment{box of width $w$ centered around $\mathbf{x}_j$}
\State Append $\mathbf{b}_j$ to $\text{b\_list}^{(i)}$
\EndFor
\If{All none of the states in $\tau^{(i)}$ are in any of the avoid-boxes in  b\_list$^{(i)}$}
\State Avoid success $z^{(i)} \gets 1$
\Else 
\State Avoid success $z^{(i)} \gets 0$
\EndIf
\State Add $z^{(i)}$ and $\text{b\_list}^{(i)}$ to the corresponding training prompt $p^{(i)}$ for $\tau^{(i)}$
\EndFor
\end{algorithmic}
\end{algorithm}

\begin{algorithm}[!t]
\caption{Hindsight Avoid-Region Relabeling: Second Pass}\label{fig:algo2}
\begin{algorithmic}[1]
\Require state space $\mathcal{S}$, copy of the original offline training dataset  $\mathcal{D_\text{copy}}$, maximum avoid-box width $w_{\text{max}}$, number of avoid states $n_{\text{avoid}}$
\For{training trajectory ${\tau_\text{copy}}^{(i)}$ in $\mathcal{D_\text{copy}}$}
\State ${z_\text{copy}}^{(i)} \gets z^{(i)}$
\While{${z_\text{copy}}^{(i)} = z^{(i)}$}
\State Initialize avoid-regions list $\text{b\_list}^{(i)} \gets [ ]$
\For{$j$ in 1, 2, ..., $n_{\text{avoid}}$}
\State Randomly initialize avoid centroid $\mathbf{x}_j \in \mathcal{S}$
\State Randomly choose avoid-box width $w \in [0, w_{\text{max}}]$
\State avoid-box $\mathbf{b}_j \gets $ concatenate($\mathbf{x}_j - \frac{w}{2}$, $\mathbf{x}_j + \frac{w}{2}$ ) \Comment{box of width $w$ centered around $\mathbf{x}_j$}
\State Append $\mathbf{b}_j$ to $\text{b\_list}^{(i)}$
\EndFor
\If{All none of the states in $\tau^{(i)}$ are in any of the avoid-boxes in  b\_list$^{(i)}$}
\State Avoid success ${z_\text{copy}}^{(i)} \gets 1$
\Else 
\State Avoid success ${z_\text{copy}}^{(i)} \gets 0$
\EndIf
\EndWhile
\State Add ${z_\text{copy}}^{(i)}$ and $\text{b\_list}^{(i)}$ to the corresponding training prompt ${p_\text{copy}}^{(i)}$ for ${\tau_{\text{copy}}}^{(i)}$
\EndFor
\end{algorithmic}
\end{algorithm}

\subsubsection{More Sophisticated Avoid Centroid Sampling Strategies}\label{sec:data-prep-relabeling-soph-appendix}

The sampling of avoid centroids in Line 4 of Algorithm~\ref{fig:algo1} can be done using naive uniform sampling over the state space (the default) or more sophisticated methods.

\xhdr{Contour-Based Sampling} One such sophisticated method we use samples avoid centroids that fit into the nooks of the broad contour of a training trajectory $\tau$ (Figure~\ref{fig:hull}a). This is the sampling method used for the \texttt{FetchReachObstacle} environment (a very open environment with no inherent restricted areas of the state space). We can acquire an outline of the general \textit{contour} of a $\tau$ by calculating a concave hull of the data points $\mathbf{s}_t \in \tau$ in the state space $\mathcal{S}$ using the algorithm presented in \cite{park2013} as implemented in the \texttt{concave\_hull} library~\cite{Tang2022}. We denote the set of states in $\tau$ that belong to the convex hull as $\mathcal{S_\text{convex}}$. We can then find "nooks" (concave portions) of the trajectory by calculating the convex hull using the algorithm presented in \cite{Graham1972-hm} as implemented the \texttt{concave\_hull} library, and then find the points that are part of the concave hull but not the convex hull; these points outline the concave nooks of the contour and we denote the set of these states as $\mathcal{S_\text{nook}}$. For each nook in trajectory $\tau$, we then find the two convex hull points in $\mathcal{S_\text{convex}}$ bordering the set $\mathcal{S_\text{nook}}$ and sample a point between them as the avoid centroid. As a result, the trajectory of states \textit{appears} to be "attempting" to avoid the hindsight-relabeled avoid centroid by wrapping around it (Figure~\ref{fig:hull}a). Although the training trajectory was not trying to avoid this state, this does not matter because the resulting hindsight-relabeled trajectory does demonstrate how to circumvent an avoid state by wrapping around it. This sampling method is found to improve performance on the \texttt{FetchReachObstacle} environment (Figure~\ref{fig:hull}b).

\xhdr{Sampling From a Limited Portion of the State Space} In certain state spaces, there may be regions where sampling an avoid centroid there would not provide helpful information. Thus, instead of sampling from the entire state space, we can just sample from just the portion of the state space $\mathcal{S}_\text{limited} \subset \mathcal{S}$ instead. This is done in the \texttt{MazeObstacle} and \texttt{Cariogenesis} environments, see Appendices~\ref{sec:expsetup-gym-appendix} and \ref{sec:expsetup-cardio-appendix} for details.

\section{Model/Training Details}\label{sec:modeldetails-appendix} 
\subsection{Model Architectural Details and Hyperparameters}\label{sec:modeldetails-model-appendix} 
\xhdr{Representations and Embeddings for Prompt Tokens} All prompt tokens, regardless of type, are embedded into the same latent space as the action ($\mathbf{a}_t$) and state ($\mathbf{s}_t$) tokens in the main trajectory sequence: $\mathbb{R}^{d_h}$, where $d_h$ is the hidden dimension/embedding dimension (corresponding to hyperparameter \texttt{embed\_dim} below). \textbf{The avoid success indicator token} $z$ is initially presented to the model as a one-hot vector: $\left[\begin{smallmatrix}1 \\ 0\end{smallmatrix}\right]$ if $z = 1$ and $\left[\begin{smallmatrix}0 \\ 1\end{smallmatrix}\right]$ if $z = 0$. This one-hot representation is then embedded into $\mathbb{R}^{d_h}$ via the learnable embedding matrix $E_z \in \mathbb{R}^{2 \times d_h}$. \textbf{The avoid start token}, $i_b$, and \textbf{the goal start token}, $i_g$, are initially presented to the model as the one-hot vectors $\left[\begin{smallmatrix}1 \\ 0\end{smallmatrix}\right]$ and $\left[\begin{smallmatrix}0 \\ 1\end{smallmatrix}\right]$, respectively. They are then embedded into $\mathbb{R}^{d_h}$ via the learnable embedding matrix $E_i \in \mathbb{R}^{2 \times d_h}$. The \textbf{prompt end start token}, e, is embedded into $\mathbb{R}^{d_h}$ as learnable vector $\mathbf{e} \in \mathbb{R}^{d_h}$. \textbf{Avoid-region tokens} $\mathbf{b}_j \in \mathbb{R}^{2d_s}$ and the \textbf{goal token} $\mathbf{g} \in \mathbb{R}^{d_s}$ are embedded into $\mathbb{R}^{d_h}$ via learnable embedding matrix $E_b \in \mathbb{R}^{2d_s \times d_h}$ and learnable embedding matrix $E_g \in \mathbb{R}^{d_s \times d_h}$, respectively.

\xhdr{Attention Boosting}
We find that adding a bias \texttt{adelta} (a hyperparameter) to the attention logits to all prompt tokens, using the strategy described in \cite{silva2024pay}, noticeably improves the instruction-following ability of \name to the prompt—i.e., improves goal-reaching behavior without negatively impacting avoid behavior (Figure~\ref{fig:adelta}).

\xhdr{Model Hyperparameters} 
The base causal transformer architecture we use is based on the HuggingFace implementation~\cite{wolf-etal-2020-transformers} of the GPT-2 architecture~\cite{radford2019language}; all model details that are not explicitly specified here are set to the default values for the HuggingFace implementation of GPT-2. We perform hyperparameter optimization (HPO) with a simple random search algorithm, using the RayTune framework~\cite{liaw2018tune}. The tunable model hyperparameters and their respective set of possible values in the search space are described in Table~\ref{tab:table3}. Training hyperparameters are described in the next section.

\begin{table}[!h]
  \centering
  \caption{Model hyperparameters and search space}\label{tab:table3}
  \resizebox{\textwidth}{!}{%
  \begin{tabular}{*3c}
    \toprule
    \textbf{Hyperparameter} & \textbf{Description} & \textbf{Search space} \\
    \hline
    \texttt{n\_head} & \text{Number of attention heads} & \texttt{tune.choice([1, 3, 4, 6, 12])} \\
    \texttt{n\_layer} & \text{Number of self-attention layers} & \texttt{tune.choice([3, 6, 12])} \\
    \texttt{embed\_dim} & \text{Size of latent embedding space (hidden dimension)} & Fixed to be \texttt{64 * n\_head} \\
    \texttt{adelta} & Attention boosting bias to the prompt & \texttt{tune.choice([0, 1, 2, 3])} \\
    \bottomrule
  \end{tabular}}
\end{table}

The model hyperparameter configurations used are \texttt{\{n\_head=4, n\_layer=4, embed\_dim=256, adelta=2\}} for the \texttt{FetchReachObstacle} environment, \texttt{\{n\_head=6, n\_layer=6, embed\_dim=384, adelta=1\}} for the \texttt{MazeObstacle} environment, and  \texttt{\{n\_head=6, n\_layer=6, embed\_dim=384, adelta=1\}} for the \texttt{Cardiogenesis} environment.

\subsection{Training Details}\label{sec:modeldetails-training-appendix} 
\xhdr{Loss Function} 
For a batch of size $B$ of trajectories of length $T$, we use the following two-component loss function during training. The first component is the typical loss used in decision transformer models, the action loss $\mathcal{L}_\text{action}$. This is the mean squared error (MSE) between the predictions of next actions $\hat{\mathbf{a_t}}$ based on the last-layer embeddings of $(p, \tau_{1:t-1}, \mathbf{s}_t)$ and the actual next actions $\mathbf{a_t}$:
$$\mathcal{L}_\text{action} = \frac{1}{B} \sum_{i=1,2,...,B} \frac{1}{T} \sum_{t=1,2,...,T} (\widehat{\mathbf{a}^{(i)}_t} - \mathbf{a}_t^{(i)})^2 $$

The second component is used to encourage \name to learn how to be aware, at each timestep, whether or not the current state $\mathbf{s_t}$ violates any avoid-box {$\mathbf{b}_j: j  \in \{1, 2, ..., n_\text{avoid}\}$} in prompt $p$. We define the indicator $k_t$ to indicate whether or not $\mathbf{s_t}$ violates any avoid-boxes  $\mathbf{b}_j$ in the prompt, with $k_t = 1$ indicating that $\mathbf{s_t}$ does not violate any avoid-boxes and $k_t = 0$ indicating $\mathbf{s_t}$ violates at least one avoid-box. During training \textit{only}, in addition to predicting the next action $\mathbf{a_t}$, we also make \name predict ${k_t}$ using the last-layer embeddings of $(p, \tau_{1:t-1}, \mathbf{s}_t)$. We define the box awareness loss $\mathcal{L}_\text{avoid\_awareness}$ to be the binary cross entropy (BCE) loss between predicted ${\hat{k_t}}$ and actual ${k_t}$. 
$$\mathcal{L}_\text{avoid\_awareness} = \frac{1}{B} \sum_{i=1,2,...,B} \frac{1}{T} \sum_{t=1,2,...,T} k_t^{(i)} \log(\widehat{k^{(i)}_t}) + (1 - k_t^{(i)}) \log(1-\widehat{k_t^{(i)}}))$$

The combined loss function is then:
$$\mathcal{L} = \mathcal{L}_\text{action} + \alpha \mathcal{L}_\text{avoid\_awareness}$$

$\alpha$ is a tunable hyperparameter to balance the different components of the loss function. During hyperparameter optimization, we try values of $\alpha$ ranging from $0.1$ to $10$ (Fig.~\ref{fig:initialhpo}). Very large values of $\alpha$ (e.g., 5 and 10) de-prioritize $\mathcal{L}_\text{action}$ too much and significantly impairs learning a policy with strong goal-reaching (SR) and avoidance (MNC) behavior. While very small values of $\alpha$ (e.g., 0.1) results in the fastest convergence to a policy with strong SR and MNC performance, intermediate $\alpha$ values ($\alpha = 1$) allows the model to eventually converge on a policy with the lowest MNC, with equally good SR. Thus, we set it to 1.0 in all our experiments.

\xhdr{Stopping Criteria} We let all \name models train for 10,000 training steps (no visual improvement in SR or MNC beyond that), checkpointing every 100 steps. As we use a similar training scheme as the one used in the original Prompt DT~\citep{xu2022prompt} and MGPO~\citep{mgpo} papers, where we sample batches of trajectories with replacement, we cannot use "epoch" as a measurement of training progress. At every checkpointing iteration, we run an evaluation on the model in the same way we did in our experiments (described in Section~\ref{experiments}). We choose the checkpoint with the best (lowest) MNC whose SR is within $0.05$ from the checkpoint with the highest SR (to ensure we have a model that is not achieving seemingly good avoid ability by significantly sacrificing goal-reaching ability, e.g., not moving). We do the same process for all baselines, except we train the baseline for 500 epochs as is done in the RbSL paper~\citep{cao2024offline} (no visual improvement in SR or MNC beyond that).

\xhdr{Training Hyperparameters} 
Training hyperparameters are listed with their respective search spaces in Table~\ref{tab:table4}. The scheduler value \texttt{\lq lambdalr'} corresponds to the PyTorch scheduler \texttt{torch.optim.lr\_scheduler.LambdaLR} and the scheduler value  \texttt{\lq cosinewarmrestarts' } corresponds to the scheduler \texttt{CosineAnnealingWarmupRestarts} from the \texttt{cosine\_annealing\_warmup} library~\cite{Katsura2021} library is used to tune training hyperparameters. Any hyperparameters not explicitly listed default to the values used in MGPO~\citep{mgpo}.

The chosen training hyperparameter configurations are \texttt{\{batch\_size=128, learning\_rate=1e-4, scheduler=\lq cosinewarmrestarts\rq, T\_0=1000, warmup\_steps=500, alpha1=1\}} for the \texttt{FetchReachObstacle} environment, \texttt{\{batch\_size=32, learning\_rate=1e-4, scheduler=\lq lamdalr', warmup\_steps=1000, alpha1=1\}} for the \texttt{MazeObstacle} environment, and \texttt{\{batch\_size=128, learning\_rate=1e-4, scheduler=\lq lamdalr', warmup\_steps=1000, alpha1=1\}} for the \texttt{Cardiogenesis} environment.

\begin{table}[!ht]
  \centering
  \caption{Training hyperparameters and search space}\label{tab:table4}
  \resizebox{\textwidth}{!}{%
  \begin{tabular}{*3c}
    \toprule
    \textbf{Hyperparameter} & \textbf{Description} & \textbf{Search space} \\
    \hline
    \texttt{batch\_size} & \text{Batch size} & \texttt{tune.choice([32, 64, 128, 256])} \\
    \texttt{learning\_rate} & \text{Maximum learning rate} & \texttt{tune.choice([1e-4, 1e-5])} \\
    \texttt{scheduler} & \text{Learning rate scheduler} & \texttt{tune.choice([\lq cosinewarmrestarts\rq, \lq lambdalr\rq])} \\
    \texttt{warmup\_steps} & \text{Number of warmup steps} & \texttt{tune.choice([500, 1000])} \\
    \texttt{T\_0} & \texttt{T\_0} parameter to \texttt{CosineAnnealingWarmupRestarts} & Fixed to be \texttt{1000} \\
    \texttt{T\_mult} & \texttt{T\_mult} parameter to \texttt{CosineAnnealingWarmupRestarts} & Fixed to be \texttt{1} \\
    \texttt{weight\_decay} & $\lambda$ constant used in weight decay & Fixed to be \texttt{1e-4} \\
    \texttt{dropout} & Dropout ratio during training & Fixed to be \texttt{0.1} \\
    \texttt{alpha1} & The $\alpha$ weight balancing the two-component loss function & \texttt{tune.loguniform([0.1, 10])} \\
    \bottomrule
  \end{tabular}}
\end{table}
\xhdr{Compute Resources} 
All training sessions of \name models were done using a single H100 GPU. On average, \name takes around 10 hours (i.e., 10 GPU hours) to converge on a strong policy on Gymnasium Robotics tasks (depending on the task) in terms of pure \textit{training} time. The \textit{total} amount of time depends on how often we perform validation/evaluation (e.g., performing 100 interactive episodes with the environment during each evaluation session can take quite a bit of time). The maximum memory utilization during training hit 40 GB. We acknowledge this is a higher computation overhead compared to baselines, which can converge on a strong policy using 1 GPU within 2 hours of pure training time, with the maximum memory utilization during training hitting 20-30 GB. However, given that all training was done with a single GPU and a strong policy can be achieved within a day, we do not think \name is prohibitively expensive to train. Additionally, \name has the additional benefit of being able to generalize zero-shot to novel avoid-region configurations dynamically specified at test time while other methods do not, removing the need for compute-heavy retraining in many scenarios where baselines would have to be retrained.

\section{Experiment Details}\label{sec:expsetup-appendix}
\subsection{GymnasiumRobotics Environments}\label{sec:expsetup-gym-appendix}
\xhdr{Choice of Environments} 
While there are more complicated environments that are part of the Fetch and Maze suites in Gymnasium robotics, we do not evaluate on those, since the focus of those environments is testing for proficiency in long-range planning, skills learning, and hierarchical learning, which are not the focus of the current iteration of \name. Additionally, we are focusing only on the domain where the training data is 100\% generated from a random policy (Criterion 1.2), and it is difficult for RL approaches across the board to just learn good goal-reaching performance on these more complex environments under this data regime \citep{ogbench}, let alone good avoid behavior. With low goal-reaching success rates, our evaluation of avoid behavior will not be very meaningful. As an extreme example, a policy that makes the agent stay in place will achieve an MNC of 0, but its SR will also be 0; this would not be considered good avoid behavior despite the low MNC.

\xhdr{Passable and Impassable Avoid-Regions} In our custom environments, we make the avoid-regions passable (i.e., the agent can pass through these avoid states). This is in contrast to physical boxes that provide a hard constraint on the state space (i.e., the agent cannot cross the boundaries of the avoid-box) as is used in \citep{cao2024offline}. Hard constraints on the state space present an issue: no training trajectories generated in this environment truly violate any $\mathbf{b}_j$, because no training trajectories actually pass \textit{through} any $\mathbf{b}_j$ (Figure~\ref{fig:reach}b). Thus, all trajectories are somewhat optimal, in the sense that they never demonstrate "unsuccessful" avoid behavior. This setup thus fails to demonstrate Property (2.2) in Section~\ref{problemformulation}. We argue that this is an issue because it couples successful goal-reaching behavior with successful avoiding behavior. If training trajectories cannot go through avoid-boxes, all training trajectories that have succeeded in reaching the goal must have done so by circumventing avoid-boxes. A training trajectory where the agent is adamant on attempting to go through an avoid-box will get stuck at the box boundary and fail to reach the goal. Therefore, with a hard constraint setup, goal-reaching ability is tightly coupled with avoid ability, while we wish to \textit{isolate} these two aspects and see if a model can learn to acquire both goal-reaching and avoid abilities when it is possible to only acquire one but not the other. While this may not be important in the specific context of these Gymnasium Robotics tasks, in general, there may be application scenarios in which we would like to avoid a region of the state space that the agent is technically \textit{able} to pass through, but we would prefer it not. For example, there may be a road in which a self-driving vehicle \textit{can} pass through or \textit{has} previously passed through, but we would rather it not on some particular day due to construction traffic.

\xhdr{Environment Specifications} For the baseline models (AM-Lag, RbSL, WGCSL), the \texttt{FetchReachObstacle} environment state space consists of $19$ dimensions. The first $10$ dimensions correspond to the state space of the original \texttt{FetchReach} environment in Gymnasium Robotics~\citep{gymrobotics} and describe information about the location/orientation of the robot arm and the environment in general. The next $9$ dimensions describe information about the single obstacle/avoid-region present in the environment, with the same setup as used in~\citep{cao2024offline}. For \name, the \texttt{FetchReachObstacle} environment just has the first $10$ dimensions (i.e., the same state space as the original \texttt{FetchReach} environment), as \name does not utilize an avoid-region-augmented state space. For the baseline models, the \texttt{MazeObstacle} environment state space consists of $4 + 2 * n_\text{avoid}$ dimensions. The first $4$ dimensions correspond to the state space of the original \texttt{PointMaze} from Gymnasium Robotics and describe the location and velocity of the agent. Then, for each avoid-region in the environment, there are an additional $2$ dimensions added to the state space representing the $xy$-location of the centroid. Thus, the state space dimensionality gets larger the more avoid-regions the environment is specified to have. For \name, however, the state space of \texttt{MazeObstacle} is just the state space of \texttt{PointMaze} and contains $4$ dimensions. The avoid-regions in the \texttt{MazeObstacle} environment are circles of radius 0.2 around the avoid centroids.

Each environment has a parameter \texttt{max\_episode\_steps}, which dictates the maximum number of timesteps in an episode of interaction with that environment. An episode ends either when the \texttt{max\_episode\_steps} number of timesteps is reached or if the agent successfully reaches the goal. \texttt{max\_episode\_steps} is set to be \texttt{50} for \texttt{FetchReachObstacle}, \texttt{300} for \texttt{UMazeObstacle}, and \texttt{800} for \texttt{AMazeObstacle} and \texttt{HMazeObstacle}, which correspond to the original default values of \texttt{max\_episode\_steps} for \texttt{FetchReach} and \texttt{PointMaze\_UMaze} from Gymnasium Robotics.

\xhdr{Length-Normalized Cost Return} 
We observe in the visualizations of our preliminary experiments that it is possible to "hack" absolute cost return by attempting to reach the goal state in as few timesteps as possible and rushing directly through the avoid-region (Figure~\ref{fig:cost}). Such a policy, which we shall refer to as the "rushed policy," can achieve low absolute cost by minimizing the number of timesteps in the trajectory in total, and thus also minimizing the absolute number of timesteps spent in the avoid-region. However, for reach-avoid applications where safety can be critical, the rushed policy is not preferred to a slower, more cautious policy (which we shall refer to as the "cautious policy") that may take more timesteps to reach the goal but demonstrates a stronger attempt at circumventing the avoid-region. The cautious policy may result in a trajectory that accumulates a similar absolute cost return as the rushed policy trajectory (because it may have skimmed the avoid-region for a similar absolute number of timesteps as the rushed policy spent in the avoid-region), but because the overall cautious policy trajectory makes a better attempt at circumventing the avoid-region, it achieves a much lower length-normalized cost return compared to the rushed policy trajectory. That is, a smaller \textit{proportion} of the timesteps in the cautious policy trajectory violate the avoid-region compared to the rushed policy trajectory, indicating higher quality avoid behavior compared to the rushed policy trajectory. In safety-critical contexts, as long as the agent can reach the desired goal in a reasonable time (i.e., the specified \texttt{max\_episode\_steps}), the \textit{quality} of avoid behavior is more important than the speed of reaching the goal. Figure~\ref{fig:cost} demonstrates this with two visual examples.

\xhdr{Avoid-Region Relabeling} For training data from the \texttt{FetchReachObstacle} environment, hindsight-relabeled avoid centroids are sampled according to the contour-based sampling method described in~\ref{sec:data-prep-relabeling-appendix}. For training data from the \texttt{MazeObstacle} environments, hindsight-relabeled avoid centroids are sampled using the naive uniform strategy, but only from the parts of the state space that are accessible by the agent; i.e., avoid centroids cannot be sampled inside the walls of the mazes. This is to prevent the model from conflating dynamically specified avoid-regions in the prompt with hard constraints on the state space that are inherent to the environment. Since the avoid-regions in \texttt{MazeObstacle} environments are circles of radius 0.2, we choose the circumscribing box of width 0.4 around the avoid centroid for our avoid-box representation in \name, a conservative choice.

\xhdr{Baselines Cannot Generalize Zero-shot}
Here, we clarify in more concrete examples why zero-shot generalization to avoid-box sizes and numbers is not feasible with AM-Lag and RbSL as they are set up in~\citep{cao2024offline}. 

For the \texttt{FetchReachObstacle} tasks, assume at training time the avoid-boxes $b$ in the environment have width $w$. The augmented state space only includes information about the center of the avoid-box: $\texttt{centroid}(b)$. The \textit{size} of the avoid-box is encoded by the cost function. The cost function that AM-Lag and RbSL are trained on outputs a cost of $1$ if the agent's state is within the box of width $w$ surrounding $\texttt{centroid}(b)$, otherwise, it outputs $0$. Now, say we increase the avoid-box sizes to $2w$ at evaluation time. The only information the agents get about the avoid-region at evaluation time is in the centroid information $\texttt{centroid}(b)$ present in the state space, which carries no information about the box size. The agents are still trained to stray away from the box of width $w$ surrounding $\texttt{centroid}(b)$, as this is the cost function they are trained to minimize the value of. Thus, to generalize to this new box size, we must define a new cost function that outputs a cost of $1$ if the agent's state is within the box of width $2w$ surrounding $\texttt{centroid}(b)$, then \textit{retrain} the agents on this new cost function.

For the \texttt{MazeObstacle} tasks, assume at training time that there are 3 avoid-regions $b_1, b_2, b_3$ in the environment. The augmented state space for AM-Lag and RbSL at training time thus has dimensionality $4 + 2 + 2 + 2 = 10$, with 2 additional dimensions for each of $\texttt{centroid}(b_1)$$,\texttt{centroid}(b_1)$$,\texttt{centroid}(b_1)$ (see Environment Specifications above). Say, at evaluation time, we increase the number of avoid-regions to 5: $b_1, b_2, b_3, b_4, b_5$. Now the state space dimensionality is $4 + 2 + 2 + 2 + 2 + 2 = 14$. We will have to retrain the agents on this new state space.

\subsection{Cardiogenesis Environment}\label{sec:expsetup-cardio-appendix}
\xhdr{Cardiogenesis Boolean Network Details} The boolean network model comprises 15 genes, each represented by a binary variable indicating expression (1) or non-expression (0), yielding a discrete state space of size $2^{15}$. The model is depicted by Figure~\ref{fig:cardiomodel}a and described mathematically in \citep{cardiogenesisapp, booleannetworkcardiogenesis}. Actions correspond to genetic perturbations that flip the expression value of a single gene. After perturbation, the Boolean network to a new gene expression state based on its internal logic. Updates are done asynchronously, as is done in~\citep{cardiogenesisapp}: a random gene node is chosen, and its value is set (either changed or remained in place) such that it satisfies all boolean rules . A \textit{stable attractor state} is defined as a state in which all boolean rules in the boolean network are satisfied, such that the boolean network is self-consistent and further asynchronous updates will not change the node values (assuming absence of an external perturbation).

In our \texttt{Cardiogenesis} environment simulations, a state-action-state transition sequence is obtained as follows: 1) we start with the boolean network representing the current gene expression state $\mathbf{s_t}$, 2) an action is chosen (i.e., a chosen gene node is perturbed and has its value flipped from \texttt{0} to \texttt{1} or vice versa) to create a post-perturbation \textit{transient state}, 3) \texttt{k} asynchronous updates to the boolean network are performed to the network, and 4) the resulting state after \texttt{k} asynchronous updates is defined to be the next state in the trajectory, $\mathbf{s_{t+1}}$. This is depicted in Figure~\ref{fig:cardiomodel}b. Since the boolean network updates are asynchronous, there is some stochasticity involved in the transitions due to the random selection of genes to be updated at each asynchronous update step. The value of \texttt{k} affects how stochastic our transitions are; the higher the value of \texttt{k}, the more likely the boolean network model will hit a \textit{stable attractor state} in the process, and the less noisy the transitions of our \texttt{Cardiogenesis} environment will be. We choose \texttt{k=10}. Note that this value of \texttt{k} does not guarantee that all states $\mathbf{s_t}$ will be stable attractor states; this is intentional, since in reality, a cell can be perturbed while it is unstable and still in the process of reaching a stable attractor state.

\xhdr{Avoid-Region Relabeling}
At training time, the offline dataset of 60,000 random policy steps is split into trajectories of length 30. Hindsight goal relabeling is done as usual. During hindsight avoid-region relabeling, we sample avoid-regions from the top 8 most represented states (attractors and non-attractors) in the offline dataset. This choice results in approximately half of all training trajectories being successful demonstrations of avoid behavior on the first pass of hindsight avoid-region relabeling, removing the need to do a second pass. Since we are working with a discrete state space, avoid states that are discrete states rather than regions of a continuous state space, and the corresponding avoid-region token representation $\mathbf{b} \in \mathbb{R}^{15}$ is simply the avoid state itself.


\xhdr{Start State/Goal State Sampling} At evaluation time, start states are randomly chosen from stable attractor states. The goal state is fixed as the FHF state (\texttt{100010010101100}) for our experiment.

\xhdr{Example Trajectories from Experiments} Here, we show some example trajectories depicting the three scenarios described in Section~\ref{experimentsreprogramming}. Note that, unless otherwise specified, we depict trajectories where repeated states are collapsed, which we call the "collapsed trajectory." For example, the trajectory \texttt{(a, a, b, c, d, d, d)} would be collapsed into the collapsed trajectory \texttt{(a, b, c, d)}. It is important to note that it is possible for an action to lead to staying in the same state due to the dynamics of the Boolean network model. These trajectory examples are also organized in Table~\ref{tab:table4.5}.

\begin{table}[!ht]
  \centering
  \caption{Example trajectories from the \texttt{Cardiogenesis} environment}\label{tab:table4.5}
  \resizebox{\textwidth}{!}{%
  \begin{tabular}{*3c}
    \toprule
    \textbf{Initial state} & \textbf{Most visited intermediate} & \textbf{Examples}  \\
    \hline
    \texttt{000000000000000} & 
    \texttt{100000000001100} & 
    \shortstack{ 
        \textbf{- avoid prompt}: \\
        \texttt{(000000000000000, 100000000001100, FHF)} \\
        \textbf{+ avoid prompt}: \\
        \texttt{(000000000000000, 000011010100000, 000010010100000, FHF)}} \\
    \hline

    \texttt{010111111010011} & 
    \texttt{001111011011111} & 
    \shortstack{\textbf{- avoid prompt}: \\
                \texttt{(010111111010011, 001111011011111, FHF)} \\
                \texttt{(10111111010011, 010111111010011, 001111011011111,} \\ \texttt{...000111010011111, 001111011011111, 100000000001100, FHF)} \\
                \texttt{(010111111010011, 001111011011111, 000111010011111, 001111011011111, FHF)} \\
                \textbf{+ avoid prompt}: \\
                \texttt{(010111111010011, 000000000000011, 010010010100000,} \\ \texttt{...010111111010011, 000010010100000, FHF)} \\
                \texttt{(010111111010011, 000010010100000, 010111111010011, 000010010000000, FHF)}} \\
    \hline
                
    \texttt{000010010100000} & 
    \texttt{000010010100000} &
    \shortstack{ 
        \textbf{- avoid prompt}: \\
        \texttt{(000010010100000, 000010010100000, 000010010100000, } \\ \texttt{...000010010100000, 000010010100000, FHF)} \\
        \textbf{+ avoid prompt}: \\
        \texttt{(000010010100000, 000010010100000, 000010010100000, FHF)}} \\ 
  \end{tabular}}
\end{table}

As an example of Scenario 1, when the starting state is \texttt{000000000000000}, \name's policy, when given no avoid prompt, always leads to a collapsed trajectory:

\texttt{(000000000000000, 100000000001100, FHF)}

However, upon adding \texttt{100000000001100} as an avoid prompt, the resulting policy always leads to a collapsed trajectory:

\texttt{(000000000000000, 000011010100000, 000010010100000, FHF)}

which is a longer, alternative path that deterministically circumvents \texttt{100000000001100}.

In Scenario 2, the stochastic environment plays a much more prominent role. For example, when the starting state is \texttt{010111111010011}, and \name is given no avoid prompt, the resulting policy leads to a variety of trajectories. A few examples are: 

\texttt{(010111111010011, 001111011011111, FHF)}

\texttt{(10111111010011, 010111111010011, 001111011011111, 000111010011111, 001111011011111, 100000000001100, FHF)}

\texttt{(010111111010011, 001111011011111, 000111010011111, 001111011011111, FHF)}

Upon adding \texttt{001111011011111} as an avoid prompt, we still get some similar trajectories involving \texttt{001111011011111} as an intermediate state, but we get a higher frequency of trajectories that manage to avoid it, such as:

\texttt{(010111111010011, 000000000000011, 010010010100000, 010111111010011, 000010010100000, FHF)}

\texttt{(010111111010011, 000010010100000, 010111111010011, 000010010000000, FHF)}

The takeaway here is the diversity of trajectories that are possible given the same policy due to the stochastic nature of the environment.

As an example of Scenario 3, when the starting state is \texttt{000010010100000}, all trajectories have collapsed form:

\texttt{(000010010100000, FHF)} 

when there is no avoid prompt, meaning the cell can directly reach the goal state from the start state without passing through a third intermediate state. However, the uncollapsed form of these trajectories has length 5 and is:

\texttt{(000010010100000, 000010010100000, 000010010100000, 000010010100000, 000010010100000, FHF)}

indicating that \name makes multiple attempts to get out of \texttt{000010010100000} to land on \texttt{FHF} and only manages to get out on the 5th attempt, spending a total of 5 timesteps stuck in the initial state \texttt{000010010100000}. However, when we add the initial state \texttt{000010010100000} as an avoid prompt (which is unavoidable since it is the starting state), the uncollapsed form of all resulting trajectories has length 3 instead and is:

\texttt{(000010010100000, 000010010100000, 000010010100000, FHF)}

Here, \name cannot avoid the avoid state \texttt{000010010100000} entirely, but it is encouraged and finds a way to spend less time at the \texttt{000010010100000} state, getting out of the state more quickly (40\% fewer attempts).

\xhdr{Baselines} Additionally, while our primary baselines (AM-Lag and RbSL) are DDPG-based actor-critic models that are designed to work specifically with continuous action/state spaces, we create naively discretized versions of the baselines (i.e., adding a softmax layer to the actor model for outputting discrete actions) and run them on `Cardiogenesis`. We recognize that these discretized versions are not representative of the original methodologies and that more sophisticated method of translating them to discretized spaces exist, though they are not trivial and thus highlight the lack of adapability of these approaches to different state/action space structures. As we suspected, the results are quite poor as shown in Table~\ref{tab:table12}. Whilst \name was able to achieve 100\% goal-reaching SR at all times (with or without avoid prompts), AM-Lag and RbSL SRs are quite low, making it difficult to properly compare avoid behavior with \name. Even when RbSL manages to have a 100\% SR in Scenarios 1 and 3, as soon as we add the avoid prompt of the most visited intermediate, SR drops to 0\%. MNC drops to 0\% as well, but this is not meaningful, as the agent could just be sitting still or wandering aimlessly in the state space. It is difficult to tell whether the poor performance is attributed to the baselines’ inherent capabilities, or to the poor translation of a continuous model into a discrete model. Regardkess, this does further emphasize \name’s advantage in being much more adaptable to different state/action space structures compared to baselines.

\subsection{Other Notes}\label{sec:expsetup-other-appendix}

\xhdr{Lidar-based Observation Spaces and Goal-Conditioning} Regarding the discussion about OSRL approaches in Section~\ref{relatedwork}, it is worth noting that the standard "reach avoid" benchmarking task for OSRL algorithms, the \texttt{Safe Navigation} environment from Safety Gymnasium~\citep{safetygymnasium, bulletsafetygym}, may seem like it is testing the goal-conditioning and avoid-region-conditioning abilities of OSRL models, but it does not require true goal-conditioning. This is because it utilizes a relative-perspective, lidar-based observation space $\mathcal{S}_\text{lidar}$ that allows the agent to only have to learn to reach \textit{one} state in $\mathcal{S}_\text{lidar}$ in order to "generalize" to any target location in physical space: the $\mathbf{0}$ vector (indicating that the agent is 0 distance away from the goal). Therefore, models do not need to generalize to any arbitrary goal in the observation space $\mathcal{S}_\text{lidar}$, and thus OSRL models do not need to be truly goal-conditioned in $\mathcal{S}_\text{lidar}$ (most are not). The same logic can be applied to argue that the tasks do not check for true avoid-region-conditioning in the observation space. While relative-perspective observation spaces are advantageous in this regard, they cannot be conceived or used in all environments, e.g., cell reprogramming.


\section{Additional Experiments}\label{sec:newexp-appendix}

\subsection{Ablation Experiments}\label{sec:newexp-appendix-ablation}
All ablation studies are performed in the \texttt{FetchReachObstacle-BoxWidth0.16} environment, and the AS/GS tokens ablation study was additionally done in the \texttt{UMazeObstacle} environment with 3, 5, and 7 avoid states.

\xhdr{Ablation of the SI token} As shown in Table~\ref{tab:table5} in the Appendix~\ref{sec:figures-appendix}, \name with the SI token (+SI) does drastically better in terms of both MNC and SR compared to \name with the SI token removed and trained on the full dataset of both successful and unsuccessful avoid examples (-SI, full data). This is as expected, since without the SI token, the model has no signal to differentiate what is considered successful and unsuccessful avoid behavior in the training examples. Additionally, +SI also drastically outperforms \name with the SI token removed and trained on a dataset \textit{only} consisting of successful avoid examples (-SI, successes only). This clearly demonstrates the importance of including both failure examples and the SI token to teach the model what is \textit{undesirable}.

\xhdr{Ablation of the AS/GS tokens} As shown in Table~\ref{tab:table6} in Appendix~\ref{sec:figures-appendix}, \name with the AS and GS tokens (+AS,GS) performs notably better in terms of both MNC and SR compared to \name without the AS and GS tokens (-AS,GS) in \texttt{FetchReachObstacle}. In the \texttt{UMazeObstacle} environments, we observe that MNC is worse in the -AS,GS experiments than in +AS,GS setup (approaching MNC values closer to those seen with the WGCSL baseline), but there is no significant difference in SR. The performance drops in terms of MNC are not as drastic compared to those seen with the -SI ablation experiments, as -AS,GS model may potentially be able to learn that only the last token in the prompt is considered a goal state regardless of how many tokens precede it. These results show that the AS and GS tokens are not as critical as the SI token, but do provide noticeable benefits.

\xhdr{Ablation of avoid-region tokens and size component} We also run experiments in the \texttt{FetchReachObstacle} environment where we 1) ablate the avoid-region (AR) tokens while leaving the indicator tokens and 2) ablate only the \textit{size component} of the AR token vectors (i.e., only including centroid location information, no box or radius representation) to isolate the importance of including size/extent information in our avoid-region token representation.  As shown in Table~\ref{tab:table6.1} in Appendix~\ref{sec:figures-appendix}, both ablations drastically worsen MNC across all box sizes.

\xhdr{Ablation of attention boosting} We find that adding a bias \texttt{adelta} (a hyperparameter) to the attention logits to all prompt tokens, using the strategy described in \cite{silva2024pay}, noticeably improves the instruction-following ability of \name to the prompt (Figure~\ref{fig:adelta} in Appendix~\ref{sec:figures-appendix}).

\xhdr{Ablation of avoid-region relabeling} As shown in Table~\ref{tab:table10} in Appendix~\ref{sec:figures-appendix}, the complete ablation of avoid-region relabeling (i.e., using the avoid-regions that were originally generated by the environment during the collection of training data) significantly impairs \name's MNC performance across both ID and OOD box sizes, demonstrating the critical nature of avoid-region relabeling in \texttt{FetchReachObstacle}. Note that goal relabeling and the avoidance success indication (SI token) are left intact in these experiments; therefore, the goal-reaching SR is not significantly impacted.

\subsection{Extremities of OOD Generalization}\label{sec:newexp-appendix-oodlimits}
To test the limits of \name's zero-shot OOD generalization, we train \name on \texttt{UMazeObstacle} trajectories with maximum 3 avoid-regions and evaluated on \texttt{UMazeObstacle} environments with 10, 15, and 20 avoid-regions. We also evaluate the strongest baseline, AM-Lag, on these environments, but like with the main experiments, AM-Lag is retrained on each new environment (i.e., ID, not zero-shot) which gives it an advantage over \name, and thus it represents the “ideal” performance. The results included the results in Table~\ref{tab:table7}.

At 10 avoid-regions, \name’s zero-shot OOD performance is still on par with the retrained AM-Lag model in terms of MNC and better than the retrained AM-Lag model in terms of SR. At 15 and 20 avoid-regions, however, we see that \name’s zero-shot avoid behavior starts to lag behind the retrained AM-Lag’s according to MNC, indicating that \name’s OOD performance does indeed have a limit (MNC suffers when the number of avoid-regions is over 300\% of the maximum number of avoid-regions seen during training). However, interestingly, the zero-shot \name policy results in a SR that is still higher than the retrained AM-Lag policies. This is a strong result because retraining AM-Lag ensures that scenarios with a larger number of avoid-regions become in-distribution settings for AM-Lag, whereas they remain out-of-distribution for zero-shot \name.

\xhdr{Effects on attention mechanism} Additionally, we qualitatively analyze the effects of increasing prompt length (due to increasing the number of avoid-regions) on the quality of the attention mechanism. Specifically, we check to make sure that long prompt lengths do not result in critical attention defects such as consistent attention sinks or uniform/zero attention.

For each episode during evaluation, we calculate the attention weights, averaged across all attention heads, that the model gives to the first 30 tokens of the input sequence (including both the prompt and the early state/action tokens in the trajectory history) when generating the \textit{last} action token of the trajectory. We plot these values for 1, 5, 15, and 20 avoid-regions (Fig.~\ref{fig:fig11}). We choose to analyze the attention mechanism at the last action token as this is the most challenging token in terms of long-range dependence on the prompt. We notice no major attention sinks or debilitating attention uniformity across the prompt tokens that result from long prompt lengths, as there is still healthy diversity in attention distributions across different trajectories even when there is a very large, OOD number of avoid-region tokens in the prompt (e.g. 15-20).

However, we do note a few interesting observations. First, \name pays very little to no attention to the state/action tokens in the early portion of the trajectory history (occurring right after the prompt tokens) when generating the last action in the trajectory. This is likely because the model has learned that it is sufficient to make action predictions with a Markovian-like policy, taking into account only the more recent states while ignoring historical states from earlier parts of the trajectory history. Second, while there are no major attention sinks and uniformity observed, we do notice a drop in attention to individual avoid-region tokens on \textit{average} with very OOD prompt lengths (e.g. 15 and 20 avoid-regions). Intuitively, this is to be expected. It is more challenging for a model to keep track of a high number of different avoid-regions, so it must be selective in distributing its attention only to the most relevant ones at any particular action; it cannot afford to constantly pay attention to every single avoid-region like it can when there are only a couple. Additionally, this would be consistent with our discussion in Appendix G.4 “More complex avoid-region shapes.” As avoid-regions increase in the \texttt{MazeObstacle} environments, avoid-regions increasingly start to overlap with one another to compose effectively larger avoid-regions, so it would make sense that the significance of individual avoid-regions would decrease given the large amounts of overlap. While we acknowledge the decrease in average attention to individual avoid-region tokens with longer prompts can be a limitation, it is much more likely that the difficulty of the task itself would limit the number of avoid-regions before issues related to attention across long contexts arise. For example, 20 avoid regions in the maze environment already makes the environment way too crowded.  Third, \name always puts a lot of attention weight on the \textit{goal token} at the end of the prompt, regardless of how many avoid-region tokens precede it or how long the prompt is as a whole. This is not surprising, as the goal token is likely the most relevant token at the last action of the trajectory, and there is ever only one goal token. However, it does show that \name is robust to increasing prompt length, as it is still able to identify the critical nature of the goal token despite the many more potentially "distracting" tokens in longer prompts.

\subsection{Effects of Expert-policy Data}\label{sec:newexp-appendix-expertmix} 
To examine the performance ceiling induced by only using random-policy data, we run an experiment in the \texttt{FetchReachObstacle} environment where we train \name on a mix of random-policy and expert-policy data provided by the RbSL study~\citep{cao2024offline} at 4 levels (0\%, 10\%, 20\%, and 30\% expert-policy data). The MNC and goal-reaching SR results are included in Table~\ref{tab:table8}.

Looking at the MNC values, we see that introducing expert-policy data seems to have little effect on avoid performance for ID box sizes (box width 0.16) and slightly OOD box sizes (box width 0.18). This suggests that, for ID box sizes, the performance ceiling of \name’s avoidance behavior when using only hindsight-relabeled, random policy training data is close to what \name can ideally achieve by learning from expert demonstrations (i.e., using only hindsight-relabeled, random-policy data provides no clear limitation). 

However, when we get to larger, more OOD box sizes (box width 0.20, 0.22, and 0.24), we see that introducing expert-policy data increasingly starts to make a difference and improve avoid performance. This suggests that using only hindsight-relabeled, random policy training data can limit performance ceiling on \name for \textit{larger, OOD} box sizes (i.e., \name can generalize better in a zero-shot manner to larger, more OOD box sizes if provided with expert data). A possible explanation for this is that the expert-policy demonstrations provided by the RbSL study are more conservative trajectories (i.e., widely avoiding the box with a much larger spatial buffer) than hindsight-relabeled, random-policy trajectories. When provided with these demonstrations involving large, wide, conservative arcs to avoid the box, \name can better generalize to larger box sizes even when the box sizes themselves are technically OOD. 

Interestingly, we also observe that the goal-reaching SR values are \textit{adversely} impacted by introducing expert-policy data across all box sizes, and more prominently with larger, more OOD box sizes (box width 0.2, 0.22, 0.24). In fact, with only random-policy data, \name maintains a high SR across all box sizes, but when expert-policy data is introduced, SR deteriorates with increasing box size. A possible explanation for this is that the cost of the improved avoid performance for larger box sizes provided by the inclusion of expert-policy data is a deterioration in goal-reaching ability. This is expected as larger box sizes require longer trajectories to successfully circumvent, so given that the allotted time to reach the goal is fixed across all experiments, it is more likely that the agent does not successfully reach the goal in time when it is successfully circumventing a very large box. 

We also provide baseline results for this expert-mixture experiment. The results are shown in the Table~\ref{tab:table8}; as usual, it is important to note that the baselines are retrained for every box size while \name was applied zero-shot. As with the main results in Table!\ref{tab:table1}, \name beats both baselines in MNC across all settings except for the box width 0.16 settings, where \name loses to AM-Lag on MNC but has superior SR. In general, we notice the same trend as we saw with \name, where increasing the proportion of expert data generally improves MNC while sacrificing SR, implying that the bias that expert demonstration brings is to be more conservative. However, the drop in SR across expert \% for box sizes 0.2-0.24 is much less drastic for the baselines than for \name, which we attribute to \name being applied zero-shot and the baselines being re-trained on every setting. It thus seems that the data bias introduced by expert data is exacerbated the more OOD the zero-shot setting gets.

\subsection{Effects of Changing Avoid-Region Token Representation}\label{sec:newexp-appendix-alternativerep} 

To examine how flexible \name is with regards to the vector representation used for avoid-region tokens $\mathbf{b}_j \in \mathbb{R}^{d_a}: j \in \{1, 2, ..., n_\text{avoid}\}$, we train and evaluate \name on the \texttt{UMazeObstacle} environment using the alternative representation for spherical avoid-regions, as described in Appendix~\ref{sec:data-prep-avoidrepresentation-appendix}. This representation should allow the model to learn a tighter bound around the avoid-regions, compared to the more conservative circumscribing box representation described in Appendix~\ref{sec:expsetup-gym-appendix}. 

As the results in Table~\ref{tab:table9} show, changing the avoid-region token representation from the box representation to the spherical representation does not impact \name's to learn strong reach-avoid policies, with performance that is comparable to the original \texttt{UMazeObstacle} experiments. One difference is that using the performance using the spherical representation seems to lead to \textit{slightly} worse MNC and a higher SR; this is consistent with intuition, as the original circumscribing box representation would provide a more conservative boundary of the avoid-regions in \texttt{UMazeObstacle} and thus result in more conservative trajectories that have better MNC at the cost of goal-reaching SR.

\xhdr{More complex avoid-region shapes} In addition to being flexible to different avoid-region token representations, \name can be prompted to avoid more complex avoid-regions by representing them as a composition of multiple smaller avoid-regions. This is implicitly demonstrated in the \texttt{MazeObstacle} experiments with a high number of avoid-regions (Section~\ref{experimentsmaze} Results 2, Section~\ref{experimentsablation} Results 4 "Extremities of OOD generalization", Appendix~\ref{sec:newexp-appendix-oodlimits}). In these settings, the high number of randomly generated avoid-regions will result in groups of partially overlapping avoid-regions that compose together to create effectively larger avoid-regions of complex shapes (Fig.~\ref{fig:complex}).

\subsection{Simultaneous Varying of Multiple Avoid-Region Axes}\label{sec:newexp-appendix-motionplanning} 

To demonstrate \name's ability to respond to variation in multiple axes of avoid-region specification simultaneously, we implement a navigation task called \texttt{MotionPlanning}, where the agent must reach a goal in an open rectangular environment while avoiding a random number (ranging from 2 to 4) of smaller rectangular avoid-regions, each generated with random location, aspect ratio (ranging from 1:1 to 1:2.5), and size (with side lengths ranging from 0.24 to 0.6). Unlike previous tasks, this environment simultaneously involves variation in location, shape, size, and number of avoid-regions. \name and baselines were trained in-distribution on this task, and the results are in Table~\ref{tab:table10.1}. As shown, \name achieves the best MNC on this task and is on par with AM-Lag for SR, with RbSL lagging far behind.

{\subsection{Image-Based Observation Spaces and Avoid-Region Representations}\label{sec:newexp-appendix-image} 
To explore \name's potential in handling 1) high-dimensional observation spaces and 2) unstructured, flexible avoid-region token representations that can capture abstract avoidance desires without knowledge of a pre-defined vectorization, we conduct a preliminary proof-of-concept demonstration of \name applied to an image-based version of \texttt{UMazeObstcale} that we call \texttt{UMazeObstacleImage}. Here, we replace the observation space for \texttt{UMazeObstacle} with an image-based observation space. To obtain state $\mathbf{s}_t$, we take an 224x224 image of the maze with the agent's current position visualized as a green sphere, then pass the image through Torchvision's pretrained ResNet-18 model \citep{resnetcvpr, torchvision2016}. We then take the representation outputted by the fully-connected layer at the end of the ResNet-18 architecture as our state representation, giving us a 1000-dimensional state vector $\mathbf{s}_t \in \mathbb{R}^{1000}$ (Fig.~\ref{fig:fig12}). Similarly, we acquire avoid-region tokens and goal tokens via the same process, except we replace the green sphere representing the agent's position in the original image with blue objects representing the avoid-regions and red spheres representing goals, respectively. Therefore, the avoid-region tokens $\mathbf{b}_j \in \mathbb{R}^{1000}$ and the goal token $\mathbf{g} \in \mathbb{R}^{1000}$ have the same dimensionality as the state vectors.

We then use the same hindsight-relabeled training trajectories from our \texttt{UMazeObstacle-NumAvoid1} dataset for training, but replace all the state, avoid-region, and goal tokens with these new representations. We benchmark against AM-Lag, RbSL, and WGCSL as baselines using the same evaluation process described for the \texttt{UMazeObstacle} environments. As shown by the results in Table~\ref{tab:table11}, \name achieves superior MNC and comparable or superior goal-reaching SR to all baselines. The SR values here are lower than the SR results from the original \texttt{UMazeObstacle-NumAvoid1} task, but this is to be expected given the much more difficult nature of a high-dimensional state space. The SR values are also within the range of typical SR values that SOTA GCRL approaches achieve on image-based, goal-reaching maze tasks (that do not include an avoidance objective), according to the results from OGBench (which do not involve an avoidance component) \citep{ogbench}. While there is room for additional improvement through further optimization of hyperparameters, this preliminary study demonstrates \name's strong potential in being adapted to both high-dimensional tasks and flexible avoid-region representations that do not require precise knowledge of a strict, vectorized structure to specify avoid-regions.

\section{Extended Related Work}\label{sec:related-work-appendix}

Below are some additional discussions regarding the categories of related work described in Section~\ref{relatedwork} and their limitations with regard to satisfying the ideal properties described in Section~\ref{problemformulation}

\xhdr{Offline Goal-Conditioned RL}
While reward-based OGCRL can technically satisfy Property (4) by designing a reward function that takes into account both goal information and avoid-region information (e.g. positive reward for approaching the goal, negative reward for approaching an avoid-region), they \textit{effectively} fails to satisfy Property (4) in practice. While it is theoretically possible to express information about both desirable and undesirable states in a multi-component reward function, such functions are practically difficult to design such that both these sometimes conflicting desires are balanced properly~\citep{rewardesignreview, howtoreward, complexrewardfunctionsandcurriculumlearning}. Also, the reason why reward-free OGCRL algorithms strictly do not satisfy Property (4) is because, without a reward, there is no way to specify any additional desires outside of the goal $\mathbf{g}$, such as avoid-regions. Thus, reward-free OGCRL is usually a good fit in situations where the path to the goal does not matter or the demonstrations used for training are expert/optimal demonstrations that take an ideal path to the goal.

An argument can be made that reward design for reward-based methods are trivial in simpler environments and goal-reaching tasks where we can just define a sparse reward function for reaching the goal. Regarding this, reward-free GCRL has one other major advantage: sample-efficiency. The referenced work \cite{sparserewards} formally explores why reward-free, self-supervised methods (e.g., contrastive learning methods, DT-based methods like ours, etc.) are more sample-efficient than TD-based methods (e.g., Q-learning variants) with a sparse reward function. The intuition, as explained in \cite{sparserewards}, is that TD-based methods in sparse reward settings do not get any learning signal until a reward is observed, and when a reward is observed, credit assignment among the various state transitions in the trajectories is difficult to disentangle without large amounts of data. Self-supervised approaches, on the other hand, can learn from every individual state transition (by learning how to reach every successor state in every state transition). That is, they work by implicitly learning state succession dynamics with every state transition data point; therefore, self-supervised approaches can be considered in between model-based (explicitly learning environment dynamics) and model-free RL, and are therefore more sample-efficient for the same reasons why model-based RL is often more sample-efficient.

\xhdr{Offline Safe RL}
OSRL approaches circumvent the challenge of having to design multi-component reward functions by using a separate cost function to capture information about avoid behavior and solving a constrained Markov decision process~\cite{cmdp}, isolating the handling of goal-reaching and avoiding desires. We claim in Section~\ref{relatedwork} that OSRL algorithms are not truly goal-conditioned in general; however, this may seem surprising since a very common benchmark for OSRL algorithms is a reach-avoid environment called \texttt{Safe Navigation} from Safety Gymnasium~\citep{safetygymnasium}. However, we explain in Appendix~\ref{sec:expsetup-other-appendix} why the tasks in this environment are not true multi-goal RL tasks.

\xhdr{\textit{Online} Goal-Conditioned Safe RL}
While the RbSL paper~\citep{cao2024offline} (and the associated algorithms RbSL and AM-Lag) is one of the only pieces of literature we have found that explicitly attempts to create an approach that is \textit{offline}, goal-conditioned, and avoids region-conditioned, we acknowledge that there exist other goal-conditioned, avoid-region-conditioned algorithms designed for reach-avoid tasks that are $\textit{online}$ algorithms \citep{cbsreachavoid, mincostreachavoid, livelinessguarantees}. Online algorithms may work in domains like robotics, where we can reasonably create safe testing environments or have good simulators, but we would like to create an approach for domains in which online training is not feasible.

\xhdr{Decision Transformers}
We build our model off of MGPO, which is truly goal-conditioned and thus satisfies Property (3). During the offline pretraining phase, MGPO is purely data-driven and does not use a reward function, fully relying on hindsight goal relabeling. However, MGPO does use an online prompt optimization phase in addition to offline pretraining (failing to satisfy Properties (1) and (2)), utilizes reward functions during online finetuning (ultimately failing to satisfy Properties (5)), and does not explicitly take into account avoid-region information, unless it is baked into the reward function design (effectively failing to satisfy Properties (4)). Our work builds upon MGPO such that it is more ideal for reach-avoid tasks, aiming to satisfy these remaining properties.

\xhdr{JEPA-based models}
There have been recent JEPA (Joint-Embedding Predictive Architecture)-based planning methods that can also tackle sequential decision making problems in a reward-free, suboptimality-tolerant, and offline manner \cite{pldm, dinowm}. However, the primary difference between such planning methods (which are more aligned with a model-based RL paradigm) and model-free RL methods like \name is test-time/deployment-time compute. Planning methods require exploring, searching through, and rolling out many possible plans at test time, and avoidance checking would thus happen at test time (i.e., used to select appropriate plans in the search), requiring much higher test-time compute. Model-free methods like \name, on the other hand, front-loads avoidance checking to training (i.e., to calculate the values for the training examples); at deployment time, a single rollout is made (no searching) and no avoidance checks need to be performed. Additionally, while it is possible to add an avoid-region violation penalty to the planning cost to make JEPA-based planning models have avoidance capabilities, there are a couple of fundamental challenges here that \name does not face. First, since the primary innovation of JEPA-based models is that planning is done in the latent space, this avoidance check would have to be performed at test time in the latent space. Checking violation of the exact avoid-region boundaries in the state space would be difficult to do in the latent space, which does not necessarily preserve state space geometry, and decoding back into the state space to perform this check would be computationally expensive and undermine the point of JEPA-based approaches. Second, introducing an avoid-violation penalty term in the planning cost function would once again re-introduce the problem of having to manually tune the balance between the possibly conflicting objectives of reach and avoid (Property (5)). Third, models like Dino-WM are designed to work strictly in a continuous state space with no clear way to cleanly translate such models to a discrete space like Cardiogenesis (Section~\ref{experimentsreprogramming}); such adaptability to different state space types is a major advantage of \name.

\xhdr{Pareto vs. Constrained Optimization Evaluation Framework} 
Many of the safe RL methods discussed above frame the problem as a Constrained Markov Decision Process (CMDP), which is inherently a constrained optimization problem framing rather than a Pareto optimization framing like ours. In a constrained optimization framing, one objective (usually the avoidance objective in reach-avoid problems) is treated as a constraint that must be satisfied in order for the other objective (the reach objective) to even be considered when determining success; for instance: $\max_{\theta}\; J_{\text{reach}}(\theta) \quad\text{s.t.}\quad \mathbb{E}_{\tau \sim \pi_\theta(\cdot \mid g, B)} \big[\, M(\tau, B) \,\big] \le k$) 1) This implies an ordering/hierarchy of importance and priority between the two objectives. That is, in this formulation, there is always a way to determine which is the best method: first compare constraint satisfaction, if constraint is satisfied, then compare main objective. A Pareto-style evaluation framework is more general and precludes imposing such a priority hierarchy in evaluation and comparison. 2) Under a strict constraint optimization framework, one of the objectives (the constraint) can only be compared on binary granularity on the episode level (either the constraint, like avoiding a region, is satisfied in the episode or it is not), losing nuanced, granular comparison between methods on that objective. This is fitting for scenarios where avoidance is a \textit{strict}, hard constraint. However, in many cases, avoidance is not a strict/hard constraint (e.g. passing through a construction zone in driving navigation is allowed and not destructive, though still undesirable) and sometimes even unavoidable (e.g. Cardiogenesis scenario 3, Section~\ref{experimentsreprogramming}). In such cases, when avoid-regions can be or must be violated, how \textit{much} they are violated matters (e.g., how much of your route you spend in a construction zone affects how fast you reach your destination, how much time your cell spends in an undesirable gene expression state affects risk of tumorigenesis), which also relates to our motivation of using MNC as our evaluation metric of avoidance behavior (Appendix~\ref{sec:expsetup-gym-appendix}).

However, we recognize that there exist reach-avoid applications that do involve strict avoidance constraints and do fit under the priority hierarchy implied by a constrained optimization framing; in these cases, immediately terminating the episode upon violation and considering the run a failure can be a more informative evaluation framework. Therefore, for additional insight, we also run the evaluations for the \texttt{UMazeObstacle} and \texttt{AMazeObstacle} environments where trajectories are immediately terminated upon violating an avoid-region and considered a failure, and report the failure rate (FR). The results are presented in Table~\ref{tab:table14}. Note that, as usual, results for 3 or more avoid regions are zero-shot for ]name but retrained for baselines. As shown, for all \texttt{AMazeObstacle} settings and ID/slighlty OOD Umaze settings, \name maintains comparable or superior FR to the best baseline (AM-Lag), although AM-Lag starts to beat out \name on \texttt{UMazeObstacle} when the number of avoid-regions becomes more OOD for \name ($\geq$ 5) but is still ID for AM-Lag. Overall, trends/differences in FR are less drastic between the models than in MNC, so when FR values are similar (and especially when they’re high, e.g. $>$ 0.5), MNC on unterminated trajectories becomes more informative as it showcases the difference in the \textit{quality} of the trajectory in scenarios where violation is not fatal/is inevitable.


\section{Broader Impacts}\label{sec:broaderimpacts-appendix}

Foundational methods for learning reach-avoid policies, like \name, have strong potential in improving the performance of technologies in many application domains, such as general robotics, autonomous vehicles, and bioengineering. While we believe the advancement of these downstream domains can greatly benefit society, we do acknowledge that such technologies can also be used maliciously. \name and any derivatives should never be used to create autonomous agents with the harmful goals (e.g., self-navigating vehicles that maliciously target individuals) or to induce harmful biological states (e.g., programming cells to pathological cell states). \name is designed with the intent of enabling the development of \textit{safer} sequential decision making agents that can improve the convenience and health of individuals in our society.

\section{Limitations}\label{sec:limitations-appendix} 
The flexibility and zero-shot capabilities of \name are balanced out by the fact that it takes more computational resources to train upfront compared to the baselines models, having many more parameters as it is based on a GPT-2 architecture (see "Compute Resources" section under Appendix~\ref{sec:modeldetails-training-appendix}). However, given that \name can be trained to achieve a strong policy on 1 GPU in less than day in most cases, and the fact that its zero-shot generalizability reduces the amount of task-specific retraining required, this is not an unreasonable tradeoff. 

Additionally, as explored in our stress-testing experiments, there does seem to be an empirical limit to \name's zero-shot generalization to OOD avoid-region specifications (Section~\ref{experimentsreprogramming} Results 4, Appendix~\ref{sec:newexp-appendix-oodlimits}). While the extent of the OOD generalization is impressive, there may be scenarios in which a further extent of generalization is desirable.

We also acknowledge that there may be reach-avoid application domains that do not strictly require all of the Desirable Properties presented in Section~\ref{problemformulation}. For example, in many real-world contexts, we do have a lot of high-quality expert demonstrations to use for training (e.g., for autonomous driving). Therefore, the set of Desirable Properties we propose is more of an idealistic upper bound, such that a reach-avoid learning approach that satisfies all of these properties will very likely be appropriate for \textit{any} reach-avoid problem.

Finally, the scope of our work is to establish the prospect and viability of hindsight avoid-region relabeling (HAR) as a paradigm for reach-avoid learning, recognizing that there are multitudinous opportunities for the development of customized, improved HAR sampling strategies optimized for specific applications and state spaces. For example, while we demonstrated promising potential for HAR and \name to be applied to high-dimensional, pixel-based observation spaces in a proof-of-concept experiment (Section~\ref{sec:newexp-appendix-image}), we acknowledge that more in-depth, dedicated exploration of the nuanced challenges at this front (e.g., handling first-person pixel-based observation spaces) would be a meaningful complementary line of future research, which could potentially involve non-trivial HAR design optimizations specific to those challenges.

\section{Code}\label{sec:code-appendix}
The code for this project can  be accessed at the following GitHub repository: \url{https://github.com/mims-harvard/RADT}. Our repository is built on top of the repositories for MGPO~\citep{mgpo} and RbSL~\citep{cao2024offline}.

\clearpage
\section{Additional Tables and Figures}\label{sec:figures-appendix}

\begin{table}[!h]
  \centering
  \caption{Results for the success indicator token ablation study for \texttt{FetchReachObstacle-BoxWidth0.16} (avg. over 3 seeds).}\label{tab:table5}
  \resizebox{0.55\textwidth}{!}{%
  \begin{tabular}{*4c}
    \toprule
     & \textbf{+SI} & \textbf{-SI, full data} & \textbf{-SI, successes only} \\
    \hline \\
    \textbf{MNC} & $0.049 {\scriptscriptstyle \pm0.016}$ & $0.629 {\scriptscriptstyle \pm0.018} $ & $0.614 {\scriptscriptstyle \pm0.027}$ \\
    \textbf{SR} & $0.989 {\scriptscriptstyle \pm0.01}$ & $0.921 {\scriptscriptstyle \pm0.016} $ & $0.917 {\scriptscriptstyle \pm0.024}$ \\
    \bottomrule
  \end{tabular}}
\end{table}

\begin{table}[!h]
  \centering
  \caption{Results for the goal start/avoid start tokens ablation study (avg. over 3 seeds).}\label{tab:table6}
  \resizebox{0.85\textwidth}{!}{%
  \begin{tabular}{*5c}
    \toprule
    \multicolumn{1}{c}{\textbf{Environment}} & \multicolumn{2}{c}{$\text{\textbf{+AS,GS}}$}  & \multicolumn{2}{c}{\textbf{-AS,GS}} \\ \cmidrule(r){2-3} \cmidrule(l){4-5}
    \omit & \makecell{MNC} & \makecell{SR}  & \makecell{MNC} & \makecell{SR}  \\
    \hline \\
    FetchReachObstacle-BoxWidth0.16 & $0.049 {\scriptscriptstyle \pm0.016}$ & $0.989 {\scriptscriptstyle \pm0.01}$ & $0.098 {\scriptscriptstyle \pm0.014}$ & $0.892 {\scriptscriptstyle \pm0.055}$ \\
    UMazeObstacle-NumAvoid3 & $0.045 {\scriptscriptstyle \pm0.009}$ & $0.868 {\scriptscriptstyle \pm0.028}$ & $0.060 {\scriptscriptstyle \pm0.005}$ & $0.882 {\scriptscriptstyle \pm0.033}$ \\
    UMazeObstacle-NumAvoid5 & $0.087 {\scriptscriptstyle \pm0.005}$ & $0.853 {\scriptscriptstyle \pm0.015}$ & $0.102 {\scriptscriptstyle \pm0.004}$ & $0.865 {\scriptscriptstyle \pm0.043}$ \\
    UMazeObstacle-NumAvoid7 & $0.127 {\scriptscriptstyle \pm0.01}$ & $0.832 {\scriptscriptstyle \pm0.038}$ & $0.133 {\scriptscriptstyle \pm0.008}$ & $0.878 {\scriptscriptstyle \pm0.003}$ \\
    
    \bottomrule
  \end{tabular}}
\end{table}

\begin{table}[!h]
  \centering
  \caption{Results for the avoid-region token and size component ablation study in \texttt{FetchReachObstacle}.}\label{tab:table6.1}
  \resizebox{0.7\textwidth}{!}{%
  \begin{tabular}{*7c}
    \toprule
    \multicolumn{1}{c}{Box width} & \multicolumn{2}{c}{$\text{\textbf{+AR token}}$}  & \multicolumn{2}{c}{\textbf{-whole AR tokens}} & \multicolumn{2}{c}{\textbf{-AR size component}} \\ \cmidrule(r){2-3} \cmidrule(l){4-5} \cmidrule(l){6-7}
    \omit & \makecell{MNC} & \makecell{SR}  & \makecell{MNC} & \makecell{SR} & \makecell{MNC} & \makecell{SR}  \\
    \hline \\
    0.16 & $0.049$ & $0.989$ & $0.619$ & $0.9$ & $0.464$ & $0.98$ \\
    0.18 & $0.107$ & $0.978$ & $0.747$ & $1.0 $ & $0.513$ & $0.92$ \\
    0.2 & $0.164$ & $0.983$ & $0.853$ & $0.99$ & $0.655$ & $0.98$ \\
    0.22 & $0.247$ & $0.989$ & $0.91$ & $0.99$ & $0.72$ & $0.98$ \\
    0.24 & $0.409$ & $1.0$ & $0.943$ & $0.98$ & $0.83$ & $0.98$ \\
    
    \bottomrule
  \end{tabular}}
\end{table}

\begin{table}[!h]
  \centering
  \caption{Results for evaluating \name on extreme avoid-region numbers in  \texttt{UMazeObstacle} (avg. over 3 seeds).}\label{tab:table7}
  \resizebox{0.75\textwidth}{!}{%
  \begin{tabular}{*5c}
    \toprule
    \multicolumn{1}{c}{\textbf{\# of avoid-regions}} & \multicolumn{2}{c}{$\text{\textbf{\name}}^\#$}  & \multicolumn{2}{c}{\textbf{AM-Lag}} \\ \cmidrule(r){2-3} \cmidrule(l){4-5}
    \omit & \makecell{MNC (1e-2)} & \makecell{SR}  & \makecell{MNC (1e-2)} & \makecell{SR}  \\
    \hline \\
   10 & \cellcolor{blue!15}$\mathbf{16.267} {\scriptscriptstyle \pm1.002}$ & \cellcolor{blue!15}$\mathbf{0.923} {\scriptscriptstyle \pm0.029}$ & $\mathbf{16.2} {\scriptscriptstyle \pm0.5}$ & $0.82 {\scriptscriptstyle \pm0.036}$ \\
   15 & \cellcolor{blue!15}$23.667 {\scriptscriptstyle \pm0.513}$ & \cellcolor{blue!15}$\mathbf{0.92} {\scriptscriptstyle \pm0.005}$ & $\mathbf{21.6} {\scriptscriptstyle \pm0.656}$ & $0.75 {\scriptscriptstyle \pm0.015}$ \\
   20 & \cellcolor{blue!15}$29.333 {\scriptscriptstyle \pm1.762}$ & \cellcolor{blue!15}$\mathbf{0.912} {\scriptscriptstyle \pm0.01}$ & $\mathbf{24.397} {\scriptscriptstyle \pm1.114}$ & $0.79 {\scriptscriptstyle \pm0.026}$ \\
    \bottomrule
  \end{tabular}}
  {\footnotesize $^\# = \text{Model capable of zero-shot generalization.}$ 
  \colorbox{blue!15}{Results highlighted in blue are zero-shot results.}}
\end{table}

\begin{table}[!h]
  \centering
  \caption{\name results for different levels of expert-random data mixture, zero-shot on all box widths over 0.16 (avg. over 3 seeds).}\label{tab:table8}
  \resizebox{0.9\textwidth}{!}{%
  \begin{tabular}{*9c}
    \toprule
    \multicolumn{1}{c}{\textbf{Expert \%}} & \multicolumn{2}{c}{$\textbf{0}$}  & \multicolumn{2}{c}{\textbf{10}} & \multicolumn{2}{c}{\textbf{20}} & \multicolumn{2}{c}{\textbf{30}} \\
    \cmidrule(r){2-3} \cmidrule(l){4-5} \cmidrule(l){6-7} \cmidrule(l){8-9}
    \makecell{Box Width} & \makecell{MNC} & \makecell{SR}  & \makecell{MNC} & \makecell{SR}  & \makecell{MNC} & \makecell{SR}& \makecell{MNC} & \makecell{SR}  \\
    \hline \\
    0.16 & $0.049 {\scriptscriptstyle \pm0.016}$ & $0.989 {\scriptscriptstyle \pm0.010}$ & $0.053 {\scriptscriptstyle \pm0.004}$ & $0.972 {\scriptscriptstyle \pm0.025}$ & $0.055 {\scriptscriptstyle \pm0.012}$ & $0.967 {\scriptscriptstyle \pm0.029}$ & $0.053 {\scriptscriptstyle \pm0.017}$ &
    $0.95 {\scriptscriptstyle \pm0.017}$  \\ 
    0.18 & $0.107 {\scriptscriptstyle \pm0.018}$ & $0.978 {\scriptscriptstyle \pm0.009}$ & $0.096 {\scriptscriptstyle \pm0.003}$ & $0.921 {\scriptscriptstyle \pm0.008}$ & $0.105 {\scriptscriptstyle \pm0.024}$ & $0.959 {\scriptscriptstyle \pm0.009}$ & $0.103 {\scriptscriptstyle \pm0.019}$ & $0.95 {\scriptscriptstyle \pm0.033}$\\ 
    0.20 & $0.164 {\scriptscriptstyle \pm0.027}$ & $0.983 {\scriptscriptstyle \pm0.017}$ & $0.146 {\scriptscriptstyle \pm0.017}$ & $0.922 {\scriptscriptstyle \pm0.011}$ & $0.137 {\scriptscriptstyle \pm0.021}$ & $0.928 {\scriptscriptstyle \pm0.054}$ & $0.142 {\scriptscriptstyle \pm0.031}$ & $0.888 {\scriptscriptstyle \pm0.025}$ \\ 
    0.22 & $0.247 {\scriptscriptstyle \pm0.018}$ & $0.989 {\scriptscriptstyle \pm0.019}$ & $0.203 {\scriptscriptstyle \pm0.032}$ & $0.944 {\scriptscriptstyle \pm0.042}$ & $0.191 {\scriptscriptstyle \pm0.044}$ & $0.894 {\scriptscriptstyle \pm0.034}$ & $0.185 {\scriptscriptstyle \pm0.138}$ & $0.856 {\scriptscriptstyle \pm0.059}$ \\ 
    0.24 & $0.409 {\scriptscriptstyle \pm0.012}$ & $1.0 {\scriptscriptstyle \pm0.0}$ & $0.389 {\scriptscriptstyle \pm0.047}$ & $ 0.906 {\scriptscriptstyle \pm0.034}$ & $0.301 {\scriptscriptstyle \pm0.015}$ & $0.744 {\scriptscriptstyle \pm0.092}$ &  $0.317 {\scriptscriptstyle \pm0.047}$ & $0.794 {\scriptscriptstyle \pm0.041}$ \\ 
    \bottomrule
  \end{tabular}}
\end{table}

\begin{table}[!h]
  \centering
  \caption{AM-Lag (top) and RbSL (bottom) baseline results for different levels of expert-random data mixture, retrained on each box width.}\label{tab:table8.1}
  \resizebox{0.7\textwidth}{!}{%
  \begin{tabular}{*9c}
    \toprule
    \multicolumn{1}{c}{\textbf{Expert \%}} & \multicolumn{2}{c}{$\textbf{0}$}  & \multicolumn{2}{c}{\textbf{10}} & \multicolumn{2}{c}{\textbf{20}} & \multicolumn{2}{c}{\textbf{30}} \\
    \cmidrule(r){2-3} \cmidrule(l){4-5} \cmidrule(l){6-7} \cmidrule(l){8-9}
    \makecell{Box Width} & \makecell{MNC} & \makecell{SR}  & \makecell{MNC} & \makecell{SR}  & \makecell{MNC} & \makecell{SR}& \makecell{MNC} & \makecell{SR}  \\
    \hline \\
    0.16 & $0.011$ & $0.95$ & $0.013 $ & $0.949$ & $0.01$ & $0.932$ & $0.021$ &
    $0.941$  \\
    0.20 & $0.255$ & $0.955$ & $0.218$ & $0.945$ & $0.142$ & $0.941$ & $0.146$ & $0.935$ \\ 
    0.24 & $0.457$ & $0.964$ & $0.405$ & $ 0.926$ & $0.33$ & $0.879$ &  $0.323$ & $0.833$ \\ 
    \bottomrule
  \end{tabular}}
  \resizebox{0.7\textwidth}{!}{%
  \begin{tabular}{*9c}
    \toprule
    \multicolumn{1}{c}{\textbf{Expert \%}} & \multicolumn{2}{c}{$\textbf{0}$}  & \multicolumn{2}{c}{\textbf{10}} & \multicolumn{2}{c}{\textbf{20}} & \multicolumn{2}{c}{\textbf{30}} \\
    \cmidrule(r){2-3} \cmidrule(l){4-5} \cmidrule(l){6-7} \cmidrule(l){8-9}
    \makecell{Box Width} & \makecell{MNC} & \makecell{SR}  & \makecell{MNC} & \makecell{SR}  & \makecell{MNC} & \makecell{SR}& \makecell{MNC} & \makecell{SR}  \\
    \hline \\
    0.16 & $0.362$ & $1.0$ & $0.353$ & $1.0$ & $0.258$ & $0.983$ & $0.203$ &
    $0.933$  \\
    0.20 & $0.588$ & $0.995$ & $0.558$ & $0.994$ & $0.413$ & $0.972$ & $0.414$ & $0.906$ \\ 
    0.24 & $0.729$ & $1.0$ & $0.695$ & $ 0.982$ & $0.704$ & $0.976$ &  $0.651$ & $0.913$ \\ 
    \bottomrule
  \end{tabular}}
  
\end{table}

\begin{table}[!t]
  \centering
  \caption{Results for \texttt{UMazeObstacle} with alternative representation (avg. over 3 seeds).}\label{tab:table9}
  \resizebox{0.9\textwidth}{!}{
  \begin{tabular}{*7c}
    \toprule
    \multicolumn{1}{c}{\textbf{\# Avoid}} & \multicolumn{2}{c}{$\text{\textbf{\name}}^\#$}  & \multicolumn{2}{c}{\textbf{AM-Lag}} & \multicolumn{2}{c}{$\textbf{RADT-AltRep}^\#$} \\ 
    \cmidrule(r){2-3} \cmidrule(l){4-5} \cmidrule(l){6-7} 
    \omit & \makecell{MNC (1e-2)} & \makecell{SR}  & \makecell{MNC (1e-2)} & \makecell{SR}   & \makecell{MNC (1e-2)} & \makecell{SR} \\
    \hline
    3 & $4.455	{\scriptscriptstyle \pm0.895}$ & $0.868	{\scriptscriptstyle \pm0.028}$	& $4.47	{\scriptscriptstyle \pm0.94}$ & $0.645	{\scriptscriptstyle \pm0.01}$ 	& $4.587 {\scriptscriptstyle \pm0.805}$	& $0.91	{\scriptscriptstyle \pm0.035}$ \\
    5 & \cellcolor{blue!15} $8.75	{\scriptscriptstyle \pm0.531}$ & \cellcolor{blue!15} $0.853	{\scriptscriptstyle \pm0.015}$ &	$10.14	{\scriptscriptstyle \pm1.9}$ & $0.76	{\scriptscriptstyle \pm0.01}$	& \cellcolor{blue!15} $8.897	{\scriptscriptstyle \pm1.127}$ & \cellcolor{blue!15}	$0.883	{\scriptscriptstyle \pm0.038}$ \\
    7 & \cellcolor{blue!15}$12.7	{\scriptscriptstyle \pm1.0}$	&\cellcolor{blue!15} $0.832	{\scriptscriptstyle \pm0.038}$	& $12.72	{\scriptscriptstyle \pm0.702}$ &	$0.777	{\scriptscriptstyle \pm0.028}$ & \cellcolor{blue!15} $12.723	{\scriptscriptstyle \pm0.804}$ & \cellcolor{blue!15} $0.887	{\scriptscriptstyle \pm0.05}$ \\
    \bottomrule
   \end{tabular}}
  {\footnotesize $^\# = \text{Model capable of zero-shot generalization.}$ 
  \colorbox{blue!15}{Results highlighted in blue are zero-shot results.}}
\end{table}

\begin{table}[!h]
  \centering
  \caption{Results for the avoid-region relabeling ablation study on the \texttt{FetchReachObstacle} environment (avg. over 3 seeds).}\label{tab:table10}
  \resizebox{0.65\textwidth}{!}{
  \begin{tabular}{*5c}
    \toprule
    \multicolumn{1}{c}{\textbf{Box Width}} & \multicolumn{2}{c}{\textbf{+ avoid-region relabeling}}  & \multicolumn{2}{c}{\textbf{- avoid-region relabeling}} \\ 
    \cmidrule(r){2-3} \cmidrule(l){4-5}
    \omit & \makecell{MNC} & \makecell{SR}  & \makecell{MNC} & \makecell{SR}  \\
    \hline \\
   0.16 
   & $0.049 {\scriptscriptstyle \pm0.016}$ 
   & $0.989 {\scriptscriptstyle \pm0.010}$ 
   & $0.372 {\scriptscriptstyle \pm0.023}$ 
   & $0.983 {\scriptscriptstyle \pm0.016}$ \\
   0.18 
   & \cellcolor{blue!15}$0.107 {\scriptscriptstyle \pm0.018}$ 
   & \cellcolor{blue!15}$0.978 {\scriptscriptstyle \pm0.009}$ 
   & \cellcolor{blue!15}$0.516 {\scriptscriptstyle \pm0.06}$ 
   & \cellcolor{blue!15}$0.978 {\scriptscriptstyle \pm0.01}$ \\
   0.2 
   & \cellcolor{blue!15}$0.164 {\scriptscriptstyle \pm0.027}$
   & \cellcolor{blue!15}$0.983 {\scriptscriptstyle \pm0.017}$ 
   & \cellcolor{blue!15}$0.654 {\scriptscriptstyle \pm0.006}$ 
   & \cellcolor{blue!15}$0.972 {\scriptscriptstyle \pm0.025}$  \\
   0.22 
   & \cellcolor{blue!15}$0.247 {\scriptscriptstyle \pm0.018}$ 
   & \cellcolor{blue!15}$0.989 {\scriptscriptstyle \pm0.019}$ 
   & \cellcolor{blue!15}$0.667 {\scriptscriptstyle \pm0.07}$ 
   & \cellcolor{blue!15}$0.961 {\scriptscriptstyle \pm0.01}$ \\
   0.24 
   & \cellcolor{blue!15}$0.409 {\scriptscriptstyle \pm0.012}$ 
   & \cellcolor{blue!15}$1.0 {\scriptscriptstyle \pm0.0}$ 
   & \cellcolor{blue!15}$0.784 {\scriptscriptstyle \pm0.009}$ 
   & \cellcolor{blue!15}$0.983 {\scriptscriptstyle \pm0.017}$ \\
   \bottomrule 
   \end{tabular}}
  {\footnotesize 
  \colorbox{blue!15}{Results highlighted in blue are zero-shot results.}}
\end{table}

\begin{table}[!h]
  \centering
  \caption{Results for the \texttt{MotionPlanning} environment that involves simultaneous varying of avoid-region number, location, shape, and size.}\label{tab:table10.1}
  \resizebox{0.4\textwidth}{!}{%
  \begin{tabular}{*4c}
    \toprule
     & \textbf{\name} & \textbf{AM-Lag} & \textbf{RbSL} \\
     \hline \\
   MNC (1e-2)
   & $\mathbf{2.12}$ 
   & $3.81$ 
   & $5.14$  \\
   SR
   & $\mathbf{0.99}$ 
   & $\mathbf{1.0}$ 
   & $0.41$  \\
   \bottomrule
  \end{tabular}}
\end{table}

\begin{table}[!h]
  \centering
  \caption{Results for the \texttt{MazeObstacleImage-NumAvoid1} environment (avg. over 3 seeds).}\label{tab:table11}
  \resizebox{0.65\textwidth}{!}{%
  \begin{tabular}{*5c}
    \toprule
     & \textbf{\name} & \textbf{AM-Lag} & \textbf{RbSL} & \textbf{WGCSL} \\
     \hline \\
   MNC (1e-2)
   & $\mathbf{1.1} {\scriptscriptstyle \pm0.95}$ 
   & $1.6 {\scriptscriptstyle \pm1.23}$ 
   & $1.87 {\scriptscriptstyle \pm0.78}$ 
   & $2.55 {\scriptscriptstyle \pm1.44}$ \\
   SR
   & $\mathbf{0.617} {\scriptscriptstyle \pm0.029}$ 
   & $0.427 {\scriptscriptstyle \pm0.031}$ 
   & $0.557 {\scriptscriptstyle \pm0.121}$ 
   & $\mathbf{0.62} {\scriptscriptstyle \pm0.072}$ \\
   \bottomrule
  \end{tabular}}
\end{table}

\begin{table}[!h]
  \centering
  \caption{Results for discrete versions of baselines on \texttt{Cardiogenesis}.}\label{tab:table12}
  \resizebox{0.75\textwidth}{!}{%
  \begin{tabular}{*7c}
    \toprule
    \multicolumn{1}{c}{\textbf{Scenario + setup}} & \multicolumn{3}{c}{\textbf{discretized AM-Lag}}  & \multicolumn{3}{c}{\textbf{dicretized RbSL}} \\ \cmidrule(r){2-4} \cmidrule(l){5-7}
    \omit & \makecell{\% visitation} & \makecell{mean length}  & \makecell{SR} & \makecell{\% visitation} & \makecell{mean length}  & \makecell{SR} \\
    \hline \\
    Scenario 1 -avoid & $1.0$ & $29.4$ & $0.14$ & $1.0$ & $4.2$ & $1.0$\\
    Scenario 1 +avoid & $0.0$ & $31.1$ & $0.0$ & $0.0$  & $31.0$ & $0.0$\\
    Scenario 2 -avoid & $0.51$ & $13.17$ & $0.67$ & $0.32$  & $18.29$ & $0.59$ \\
    Scenario 2 +avoid & $0.33$ & $15.68$ & $0.69$ & $0.26$  & $4.53$ & $1.0$ \\
    Scenario 3 -avoid & $1.0$ & $27.06$ & $0.28$ & $1.0$  & $11.6$ & $0.98$ \\
    Scenario 3 +avoid & $1.0$ & $29.32$ & $0.12$ & $0.0$  & $26.6$ & $0.0$ \\
    \bottomrule
  \end{tabular}}
\end{table}

\begin{table}[!h]
  \centering
  \caption{Failure rates for the \texttt{UMazeObstacle} (left) and \texttt{AMazeObstacle} (right) environments.}\label{tab:table14}
  \resizebox{0.35\textwidth}{!}{%
  \begin{tabular}{*4c}
    \toprule
     \# Avoid & $\textbf{\name}^\#$ & \textbf{AM-Lag} & \textbf{RbSL} \\
     \hline \\
   1
   & $0.103$ 
   & $0.113$ 
   & $0.127$  \\
   3
   & $0.34$ 
   & $0.337$ 
   & $0.348$  \\
   5
   & \cellcolor{blue!15}$0.56$
   & $0.513$ 
   & $0.621$  \\
   7
   & \cellcolor{blue!15}$0.591$
   & $0.563$ 
   & $0.621$  \\
   \bottomrule
  \end{tabular}}
  \resizebox{0.35\textwidth}{!}{%
  \begin{tabular}{*4c}
    \toprule
     \# Avoid & $\textbf{\name}^\#$ & \textbf{AM-Lag} & \textbf{RbSL} \\
     \hline \\
   1
   & $0.104$ 
   & $0.13$ 
   & $0.12$  \\
   3
   & $0.152$ 
   & $0.16$ 
   & $0.29$  \\
   5
   & \cellcolor{blue!15}$0.312$ 
   & $0.3$ 
   & $0.33$  \\
   7
   & \cellcolor{blue!15}$0.453$ 
   & $0.46$ 
   & $0.45$  \\
   \bottomrule
  \end{tabular}}
  {\footnotesize $^\# = \text{Model capable of zero-shot generalization.}$ 
  \colorbox{blue!15}{Results highlighted in blue are zero-shot results.}}
\end{table}

\begin{figure}[!ht]
\begin{centering}
\includegraphics[width=\textwidth]{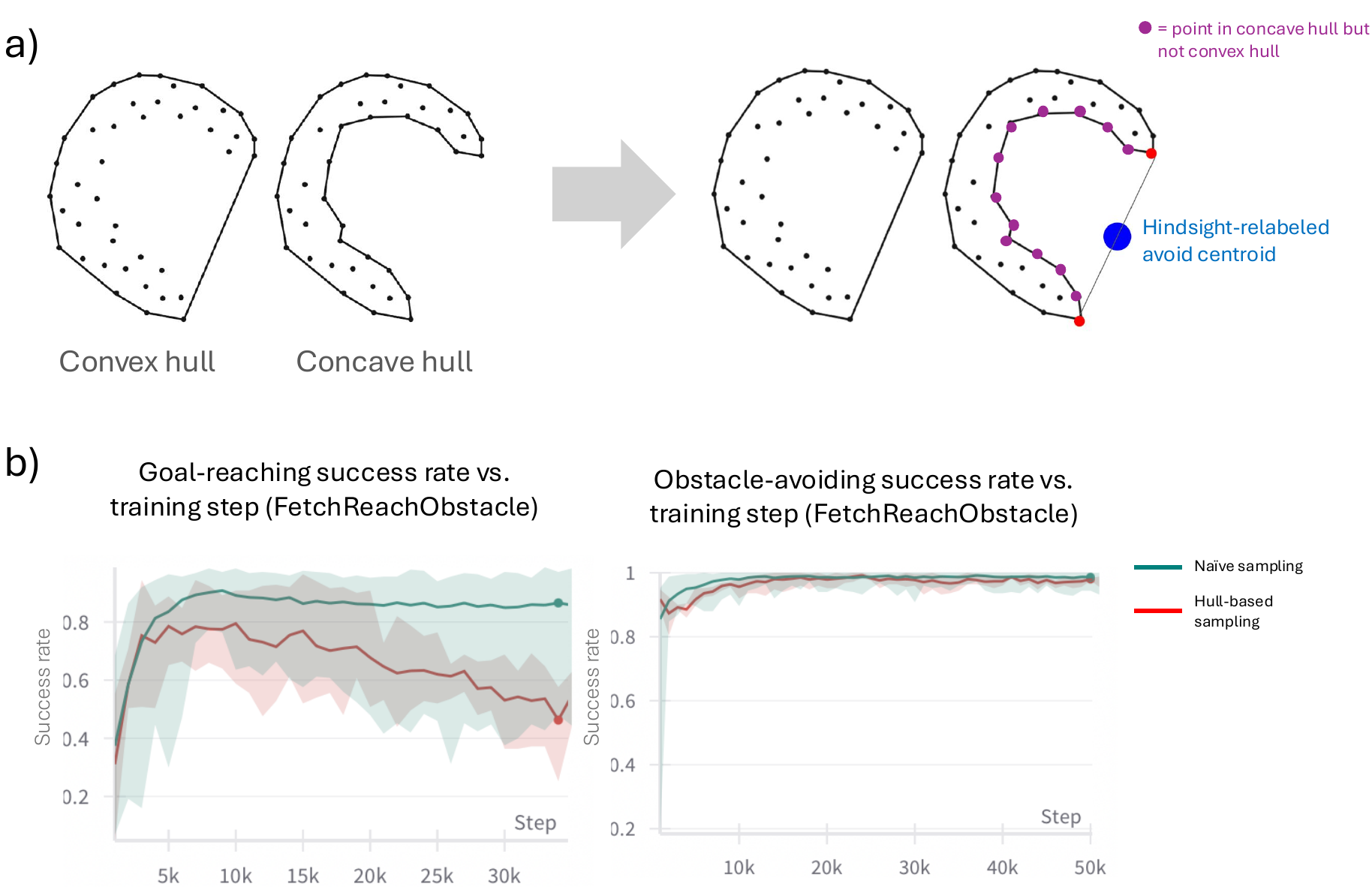}
\caption[]{a) 2D depiction of the contour-based centroid sampling strategy for avoiding region relabeling. The convex and concave hulls are calculated for the set of data points in the state space for a training trajectory. Points that are part of the concave hull but not the convex hull (depicted here in purple) comprise concave portions of the trajectory that can be viewed as "wrapping around" some unknown avoid centroid. To generate an avoid centroid that fits this intuitive interpretation in hindsight, we locate the two points bordering this concave region (depicted here in red) and sample a point in between them. The resulting trajectory of data points looks like it's "trying" to avoid the blue hindsight-relabeled avoid centroid by wrapping around it. Figure built upon an illustration from \cite{convexconcave}. b) Using the contour-based strategy for sampling avoids centroids results in a higher maximum SR for the \texttt{FetchReachObstacle} environment and notably mitigates/delays overfitting.} \label{fig:hull}
\end{centering}
\end{figure}

\begin{figure}[!t]
\begin{centering}
\includegraphics[width=\textwidth]{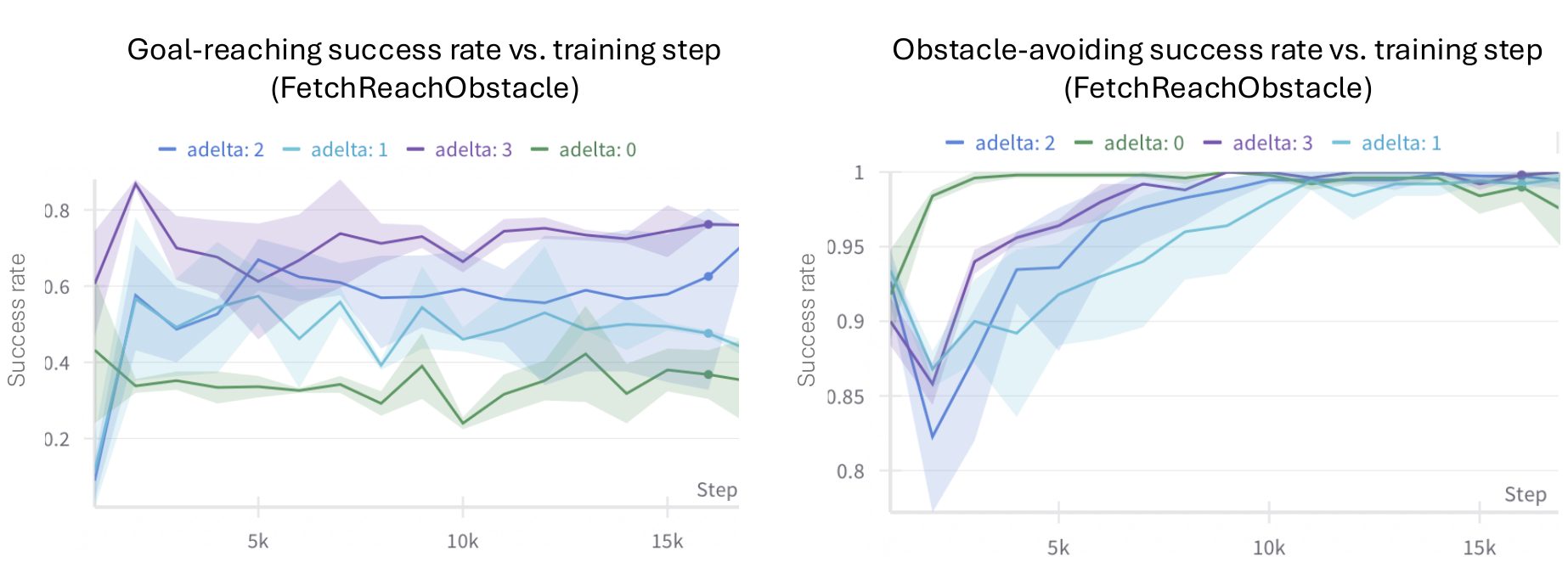}
\caption[]{The maximum goal-reaching success rate for \name trained on the FetchReachObstacle task improves with increasing the attention boosting bias \texttt{adelta} to the prompt tokens. The maximum obstacle-avoiding success rate, on the other hand, seems to be unaffected by the value of \texttt{adelta}.}\label{fig:adelta}
\end{centering}
\end{figure}

\begin{figure}[!t]
\begin{centering}
\includegraphics[width=\textwidth]{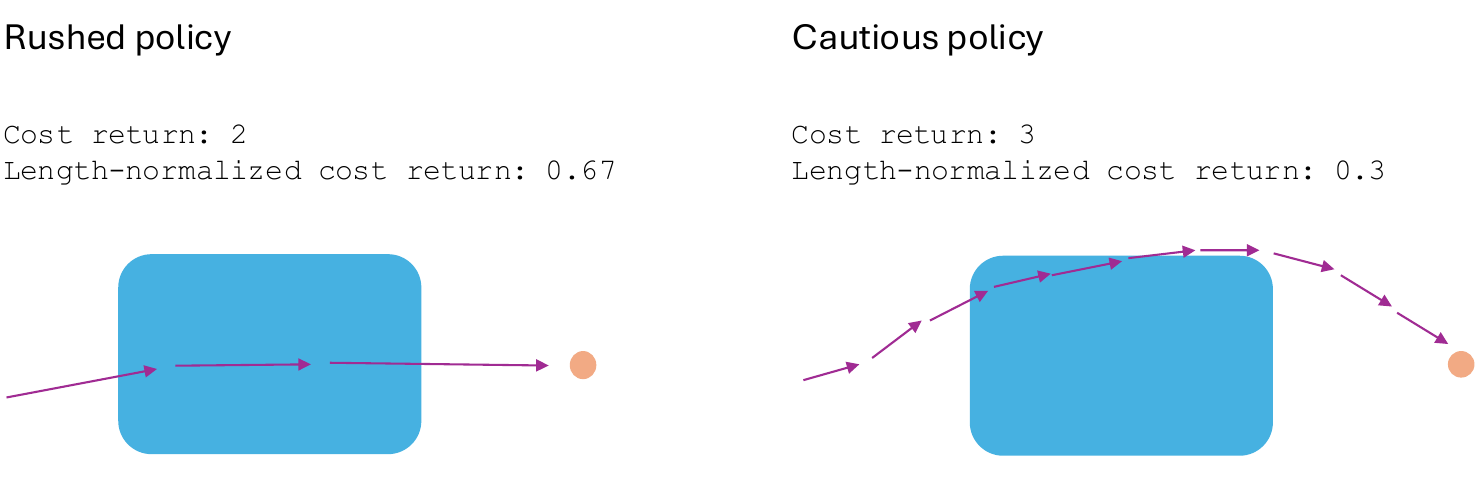}
\caption[]{A policy that rushes directly through the avoid-region to get to the goal as quickly as possible may earn a low absolute cost return but a high length-normalized cost return. On the other hand, a slower, more cautious policy that takes more timesteps to reach the goal but attempts to circumvent the avoid-region may accumulate a similar absolute cost return, but a lower length-normalized cost return.}\label{fig:cost}
\end{centering}
\end{figure}

\begin{figure}[!t]
\begin{centering}
\includegraphics[width=\textwidth]{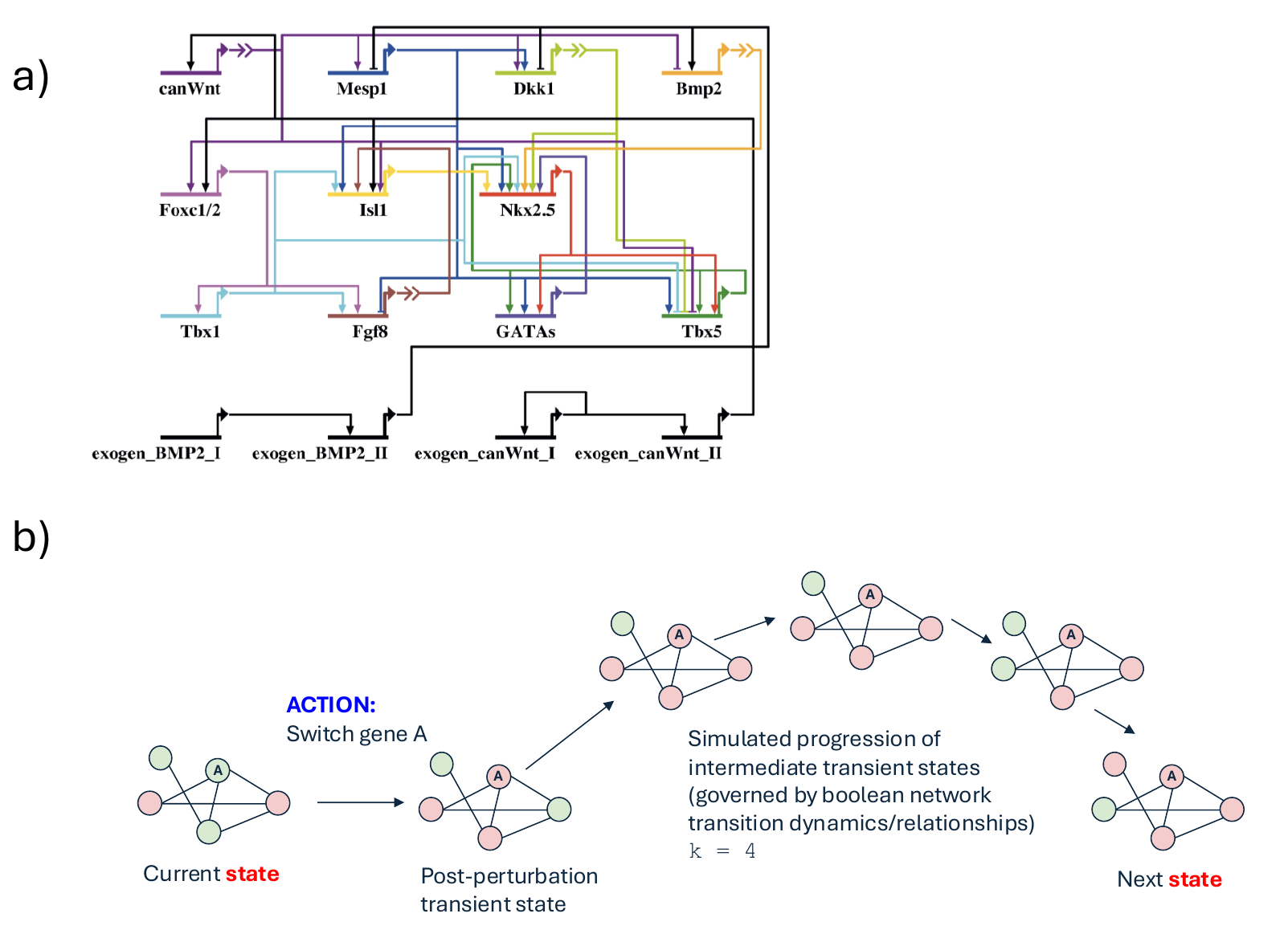}
\caption[]{a) A diagram representing the 15-gene boolean network model for mouse cardiogenesis. b) A depiction of one state-action-state transition simulated in the \texttt{Cardiogenesis} environment, where \texttt{k = 4}.}\label{fig:cardiomodel}
\end{centering}
\end{figure}

\begin{figure}[!t]
\begin{centering}
\includegraphics[width=\textwidth]{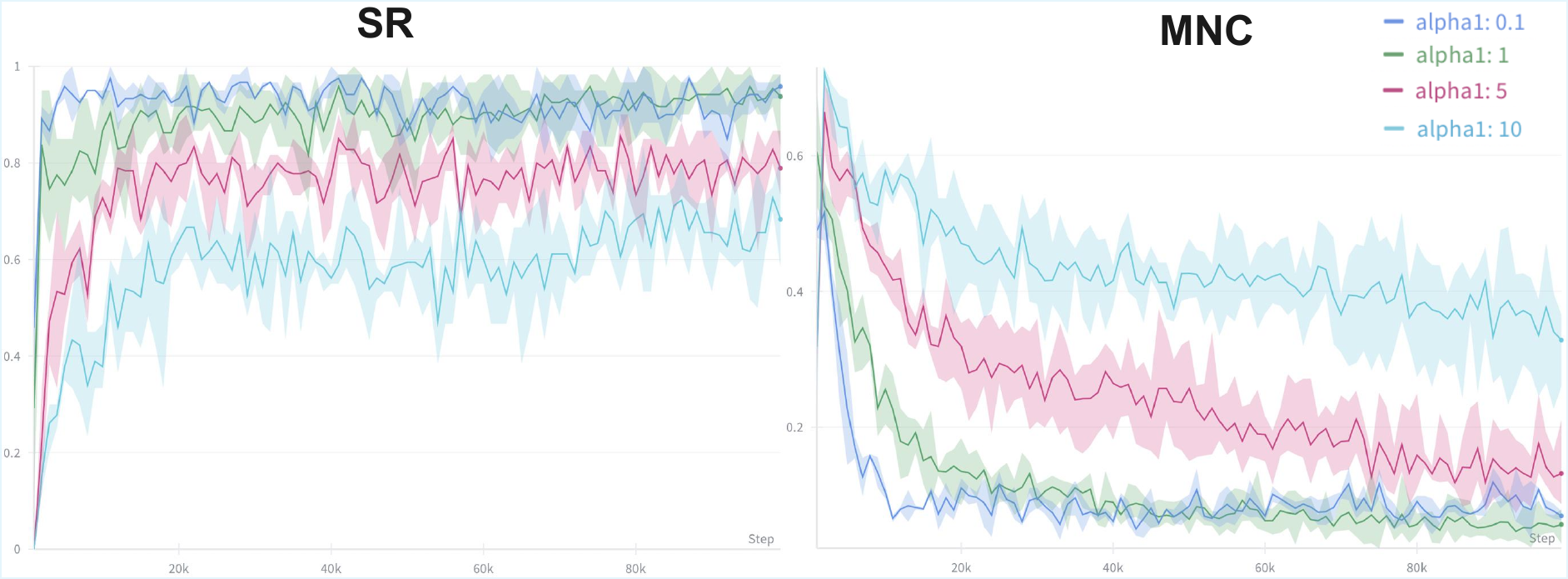}
\caption[]{The effect of different values for the loss function hyperparameter $\alpha$ (ranging from $\alpha = 0.1$ to $\alpha = 10$). High values of $\alpha$ (e.g. 5, 10)  de-prioritze the action loss too much and leads to poor goal-reaching (SR) and avoidance (MNC) behavior in the resulting policy. Small $\alpha$ values (e.g., 0.1, 1) lead to convergence on a policy with strong SR and MNC; however, when $\alpha$ is too small ($\alpha = 0.1$), the lowest MNC achieved is not as low as the lowest MNC achieved when $\alpha = 1$. We chose the value $\alpha = 1$. Training curves shown are averaged across 3 replicate runs per unique value of $\alpha$. }\label{fig:initialhpo}
\end{centering}
\end{figure}

\begin{figure}[!t]
\begin{centering}
\includegraphics[width=\textwidth]{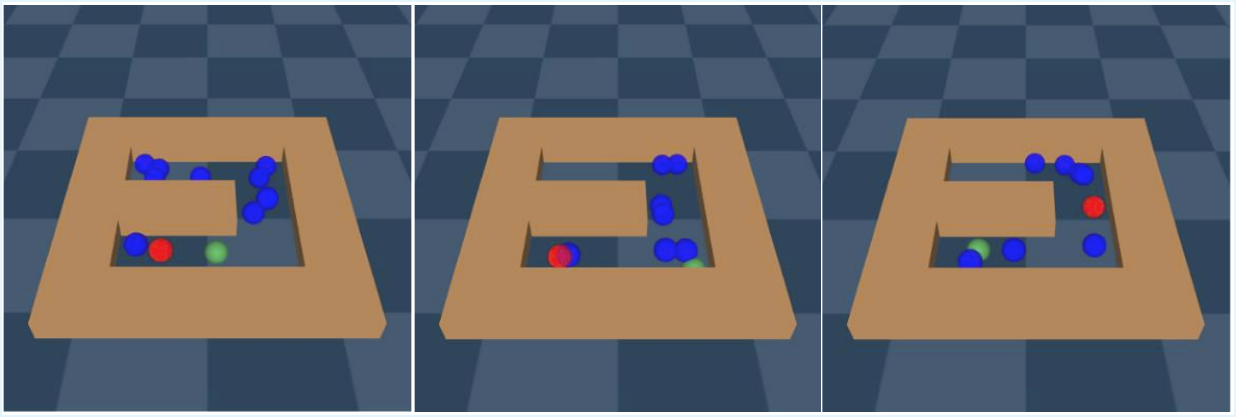}
\caption[]{When the number of avoid-regions is high (e.g. 7 and 10 avoid-regions depicted in the U-Maze examples here), there is a often quite a lot of overlapping avoid-regions that compose together to form larger avoid-regions of more complex shape that the agent must circumvent.}\label{fig:complex}
\end{centering}
\end{figure}

\begin{figure}[!t]
\begin{centering}
\includegraphics[width=\textwidth]{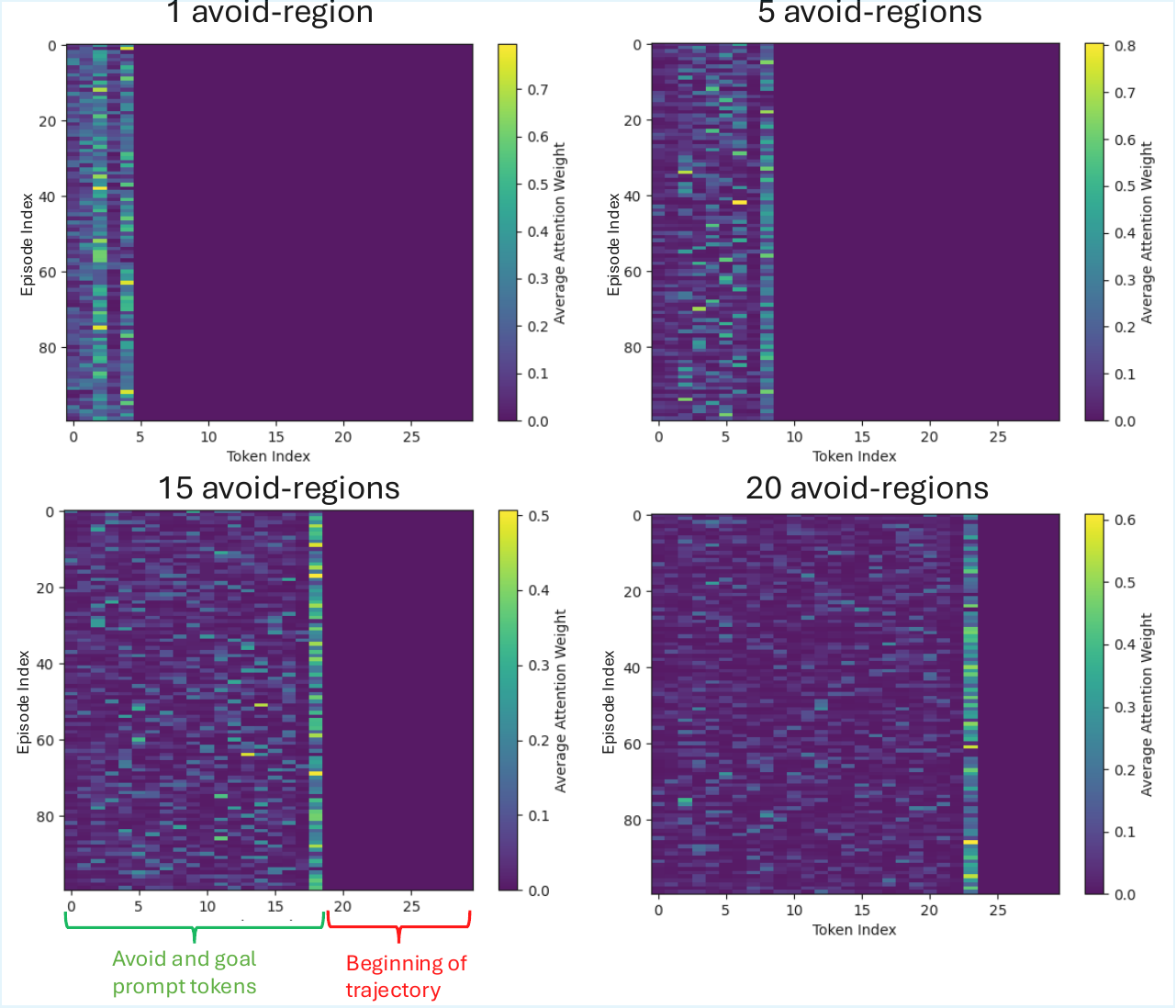}
\caption[]{Plots of average attention weights to the first 30 tokens of the input sequence when \name is outputting the last action of an evaluation episode (each episode is a row). The first 30 tokens include both the prompt (which include 1, 5, 15, or 20 avoid-region tokens) and the beginning of the trajectory history. \name pays little attention to the beginning of the trajectory history as it has learned a mostly Markovian policy. No noticeable attention sinks and debilitating attention uniformity is observed with increasing prompt length, but the attention to individual avoid-region tokens on average does decrease with increasing prompt length. Attention to the goal token remains strong regardless of prompt length.}\label{fig:fig11}
\end{centering}
\end{figure}

\begin{figure}[!t]
\begin{centering}
\includegraphics[width=\textwidth]{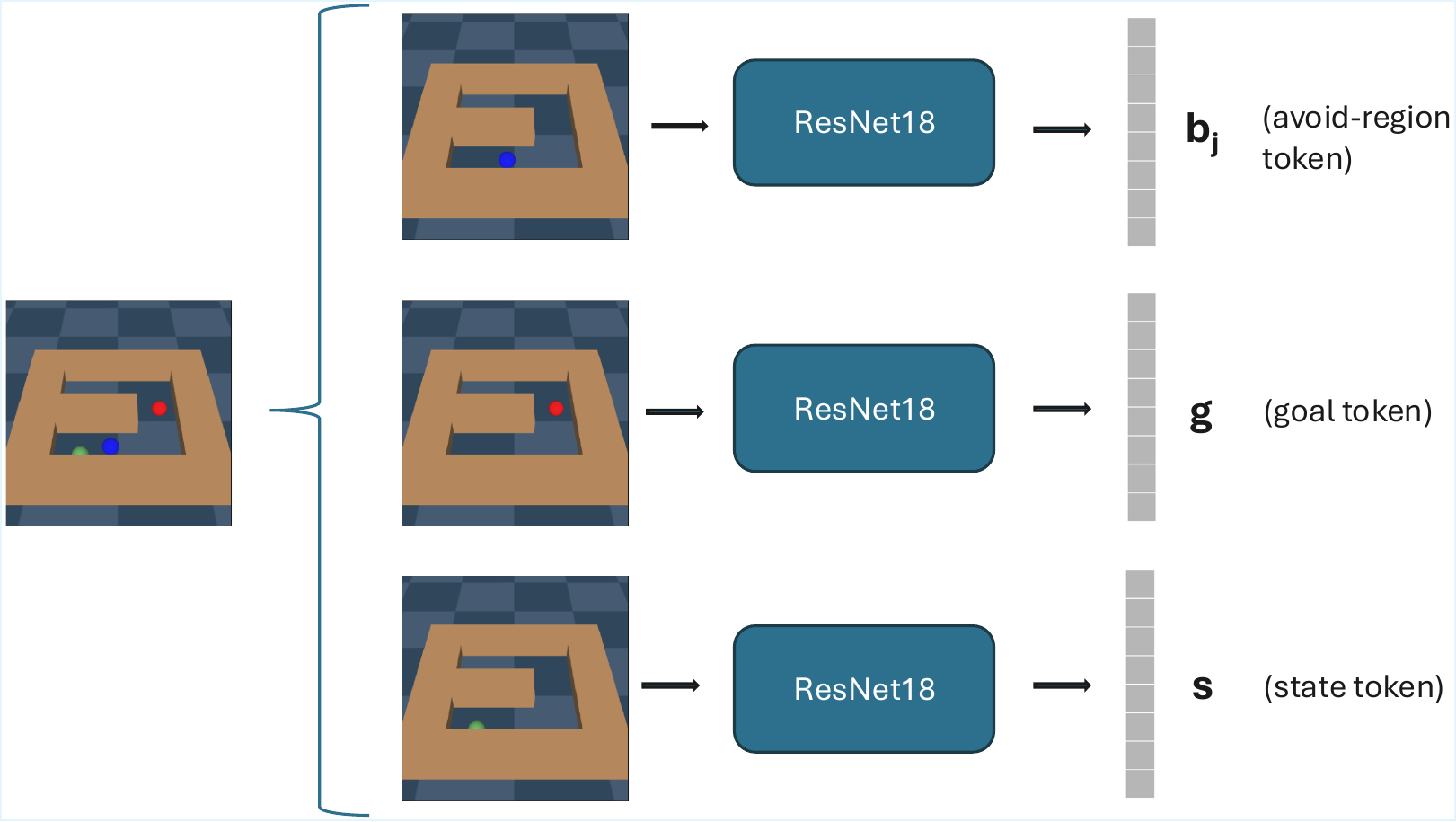}
\caption[]{To obtain the avoid-region, goal, and state tokens for the \texttt{UMazeObstacleImage} environment, we use 224x224 images capturing the positions of the avoid-region(s) (blue), goal state (red), and agent position (green) in the maze. The images are fed through a pretrained ResNet-18 model to obtain 1000-dimensional flattened vector representations $\mathbf{b}_j$, $\mathbf{g}$, and $\mathbf{s}$.}\label{fig:fig12}
\end{centering}
\end{figure}

\clearpage
\section{Additional References}\label{sec:references-appendix}
\section*{Illustration Citations (NIAID NIH Bioart)}

[The following images are included in our figures.]

NIAID Visual \& Medical Arts., (10/7/2024). Fibroblast.
NIAID NIH BIOART Source. bioart.niaid.nih.gov/bioart/
152, a.

NIAID Visual \& Medical Arts., (10/7/2024). Fibroblast.
NIAID NIH BIOART Source. bioart.niaid.nih.gov/bioart/
153, b.

NIAID Visual \& Medical Arts., (10/7/2024). Fibroblast.
NIAID NIH BIOART Source. bioart.niaid.nih.gov/bioart/
154, c.

NIAID Visual \& Medical Arts. (10/7/2024). Generic Im-
mune Cell. NIAID NIH BIOART Source. bioart.niaid.
nih.gov/bioart/173.

NIAID Visual \& Medical Arts. (10/7/2024). Human Male
Outline. NIAID NIH BIOART Source. bioart.niaid.nih.
gov/bioart/232.

NIAID Visual \& Medical Arts. (10/7/2024). Intermediate Progenitor Cell. NIAID NIH BIOART Source. bioart.niaid.nih.gov/bioart/258

\bibliographystyleAppendix{iclr2026_conference}
\bibliographyAppendix{references-appendix}


\newpage
\section*{NeurIPS Paper Checklist}

The checklist is designed to encourage best practices for responsible machine learning research, addressing issues of reproducibility, transparency, research ethics, and societal impact. Do not remove the checklist: {\bf The papers not including the checklist will be desk rejected.} The checklist should follow the references and follow the (optional) supplemental material.  The checklist does NOT count towards the page
limit. 

Please read the checklist guidelines carefully for information on how to answer these questions. For each question in the checklist:
\begin{itemize}
    \item You should answer \answerYes{}, \answerNo{}, or \answerNA{}.
    \item \answerNA{} means either that the question is Not Applicable for that particular paper or the relevant information is Not Available.
    \item Please provide a short (1--2 sentence) justification right after your answer (even for \answerNA). 
\end{itemize}

{\bf The checklist answers are an integral part of your paper submission.} They are visible to the reviewers, area chairs, senior area chairs, and ethics reviewers. You will also be asked to include it (after eventual revisions) with the final version of your paper, and its final version will be published with the paper.

The reviewers of your paper will be asked to use the checklist as one of the factors in their evaluation. While \answerYes{} is generally preferable to \answerNo{}, it is perfectly acceptable to answer \answerNo{} provided a proper justification is given (e.g., error bars are not reported because it would be too computationally expensive'' or ``we were unable to find the license for the dataset we used''). In general, answering \answerNo{} or \answerNA{} is not grounds for rejection. While the questions are phrased in a binary way, we acknowledge that the true answer is often more nuanced, so please just use your best judgment and write a justification to elaborate. All supporting evidence can appear either in the main paper or the supplemental material, provided in appendix. If you answer \answerYes{} to a question, in the justification please point to the section(s) where related material for the question can be found.

IMPORTANT, please:
\begin{itemize}
    \item {\bf Delete this instruction block, but keep the section heading ``NeurIPS Paper Checklist"},
    \item  {\bf Keep the checklist subsection headings, questions/answers and guidelines below.}
    \item {\bf Do not modify the questions and only use the provided macros for your answers}.
\end{itemize}


\begin{enumerate}

\item {\bf Claims}
    \item[] Question: Do the main claims made in the abstract and introduction accurately reflect the paper's contributions and scope?
    \item[] Answer: \answerYes{} 
    \item[] Justification: We explicitly listed 4 main contributions at the end of the Introduction. Contributions 1 and 2 were detailed in Section~\ref{method}, contributions 3 and 4 were demonstrated in Section~\ref{experiments}.

\item {\bf Limitations}
    \item[] Question: Does the paper discuss the limitations of the work performed by the authors?
    \item[] Answer: \answerYes{} 
    \item[] Justification: Yes, we include a section on Limitations in Appendix~\ref{sec:limitations-appendix}.

\item {\bf Theory assumptions and proofs}
    \item[] Question: For each theoretical result, does the paper provide the full set of assumptions and a complete (and correct) proof?
    \item[] Answer: \answerNA{} 
    \item[] Justification: Our paper has no rigorous theoretical results and does not claim any rigorous theoretical results.

    \item {\bf Experimental result reproducibility}
    \item[] Question: Does the paper fully disclose all the information needed to reproduce the main experimental results of the paper to the extent that it affects the main claims and/or conclusions of the paper (regardless of whether the code and data are provided or not)?
    \item[] Answer: \answerYes{} 
    \item[] Justification: Hyperparameters and other technical specifications (e.g. RL environment specifications) are detailed in the Appendix. A GitHub repository is provided in Appendix~\ref{sec:code-appendix}, with relevant commands to run main experiments in the README.

\item {\bf Open access to data and code}
    \item[] Question: Does the paper provide open access to the data and code, with sufficient instructions to faithfully reproduce the main experimental results, as described in supplemental material?
    \item[] Answer: \answerYes{} 
    \item[] Justification: Hyperparameters and other technical specifications (e.g. RL environment specifications) are detailed in the Appendix. A GitHub repository is provided in Appendix~\ref{sec:code-appendix}, with relevant commands to run main experiments in the README.

\item {\bf Experimental setting/details}
    \item[] Question: Does the paper specify all the training and test details (e.g., data splits, hyperparameters, how they were chosen, type of optimizer) necessary to understand the results?
    \item[] Answer: \answerYes{} 
    \item[] Justification: Hyperparameters and other technical specifications (e.g. data prep, RL environment specifications) are detailed in the Appendix.

\item {\bf Experiment statistical significance}
    \item[] Question: Does the paper report error bars suitably and correctly defined or other appropriate information about the statistical significance of the experiments?
    \item[] Answer: \answerYes{}{} 
    \item[] Justification: We train each model on each benchmarking task 3 times, and include the standard deviation both in the table and as represented by error bars in the figures. This is standard practice for RL papers. No notion of statistical significance is defined.

\item {\bf Experiments compute resources}
    \item[] Question: For each experiment, does the paper provide sufficient information on the computer resources (type of compute workers, memory, time of execution) needed to reproduce the experiments?
    \item[] Answer: \answerYes{} 
    \item[] Justification: Compute resources requirements are listed in the Appendix.
    
\item {\bf Code of ethics}
    \item[] Question: Does the research conducted in the paper conform, in every respect, with the NeurIPS Code of Ethics \url{https://neurips.cc/public/EthicsGuidelines}?
    \item[] Answer: \answerYes{}{} 
    \item[] Justification: Our data contains no information that can be used to harm individuals, as they are all produced from simulations. Our study does not involve human subjects. Our work does not seem to be at high risk for the potential concerns described in "Societal Impact and Potential Harmful Consequences". 

\item {\bf Broader impacts}
    \item[] Question: Does the paper discuss both potential positive societal impacts and negative societal impacts of the work performed?
    \item[] Answer: \answerYes{} 
    \item[] Justification: In the Appendix, we include a section on Broader Impacts and discuss how our work provides foundational developments that can be potentially misused in downstream applications, including the potential misuse of autonomous vehicles/robots and bio-engineered/genetically engineered cells.
    
\item {\bf Safeguards}
    \item[] Question: Does the paper describe safeguards that have been put in place for responsible release of data or models that have a high risk for misuse (e.g., pre-trained language models, image generators, or scraped datasets)?
    \item[] Answer: \answerNA{} 
    \item[] Justification: All our data are generated from simulations and have no known risk of misuse. Our models are also trained to be run on simulated environments and have no clear risk of potential misuse.

\item {\bf Licenses for existing assets}
    \item[] Question: Are the creators or original owners of assets (e.g., code, data, models), used in the paper, properly credited and are the license and terms of use explicitly mentioned and properly respected?
    \item[] Answer: \answerYes{} 
    \item[] Justification: All original papers for baseline models are cited. All creators of benchmarking environments are cited. We built our repo off of MGPO and RbSL, and credited them in the README and Appendix.

\item {\bf New assets}
    \item[] Question: Are new assets introduced in the paper well documented and is the documentation provided alongside the assets?
    \item[] Answer:  \answerYes{} 
    \item[] Justification: We document our code in the provided anonymized repo.

\item {\bf Crowdsourcing and research with human subjects}
    \item[] Question: For crowdsourcing experiments and research with human subjects, does the paper include the full text of instructions given to participants and screenshots, if applicable, as well as details about compensation (if any)? 
    \item[] Answer: \answerNA{} 
    \item[] Justification: We do not have human subjects involved in this study.

\item {\bf Institutional review board (IRB) approvals or equivalent for research with human subjects}
    \item[] Question: Does the paper describe potential risks incurred by study participants, whether such risks were disclosed to the subjects, and whether Institutional Review Board (IRB) approvals (or an equivalent approval/review based on the requirements of your country or institution) were obtained?
    \item[] Answer: \answerNA{} 
    \item[] Justification:  We do not have human subjects involved in this study.

\item {\bf Declaration of LLM usage}
    \item[] Question: Does the paper describe the usage of LLMs if it is an important, original, or non-standard component of the core methods in this research? Note that if the LLM is used only for writing, editing, or formatting purposes and does \emph{not} impact the core methodology, scientific rigor, or originality of the research, declaration is not required.
    \item[] Answer: \answerNA{} 
    \item[] Justification: No LLMs were used for core method development.

\end{enumerate}

\end{document}